\documentclass{ecai} 

\usepackage{latexsym}
\usepackage{amssymb}
\usepackage{amsmath}
\usepackage{amsthm}
\usepackage{booktabs}
\usepackage{enumitem}
\usepackage{graphicx}
\usepackage{color}

\usepackage[switch]{lineno}
\usepackage{pdflscape}
\usepackage{booktabs}
\usepackage{longtable}
\usepackage{rotating}
\usepackage{multirow}

\DeclareMathOperator*{\argmax}{argmax}
\DeclareMathOperator*{\argmin}{argmin}

\usepackage{algorithm}
\usepackage[noend]{algpseudocode}
\usepackage{multirow}
\usepackage{subcaption}

\newcommand{\BibTeX}{B\kern-.05em{\sc i\kern-.025em b}\kern-.08em\TeX}

\begin{document}


\begin{frontmatter}


\paperid{2316} 


\title{Real-time goal recognition using approximations in Euclidean space}


\author[B]{\fnms{Douglas}~\snm{Antunes Tesch}\orcid{0000-0003-3037-919X}\thanks{Corresponding Author. Email: douglas.tesch@edu.pucrs.br.}
}
\author[A]{\fnms{Leonardo}~\snm{Amado}\orcid{0000-0001-6119-4601}
}
\author[A, B]{\fnms{Felipe}~\snm{Meneguzzi}\orcid{0000-0003-3549-6168}} 

\address[A]{University of Aberdeen}
\address[B]{Pontifical Catholic University of Rio Grande do Sul}


\begin{abstract}
While recent work on online goal recognition efficiently infers goals under low observability, comparatively less work focuses on online goal recognition that works in both discrete and continuous domains.
Online goal recognition approaches often rely on repeated calls to the planner at each new observation, incurring high computational costs. 
Recognizing goals online in continuous space quickly and reliably is critical for any trajectory planning problem since the real physical world is fast-moving, e.g. robot applications. 
We develop an efficient method for goal recognition that relies either on a single call to the planner for each possible goal in discrete domains or a simplified motion model that reduces the computational burden in continuous ones. 
The resulting approach performs the online component of recognition orders of magnitude faster than the current state of the art, making it the first online method effectively usable for robotics applications that require sub-second recognition.
\end{abstract}

\end{frontmatter}


\section{Introduction}

Goal recognition focuses on predicting an agent's behavior and determining its goal through observing the agent's actions~\cite{sukthankar2014plan,Mirsky:2021dp,Meneguzzi2021}. 
The ability to anticipate an agent's behavior is critical for autonomous agents working cooperatively and competitively.
In cooperative settings, effective goal recognition allows agents to obviate explicit communication to coordinate their joint plans. 
By contrast, effective goal recognition in non-cooperative settings allows agents to anticipate an opponent's moves and counter them in a timely fashion. 
In robot football, for instance, this is critical in anticipating the opponent's trajectory and developing a winning counter-strategy. 
Similarly, for cooperation within a match, team members can choose plans that ease recognition of their goals to others in the same team, minimizing explicit communication~\cite{Zhang2017}.

The current state-of-the-art in online goal recognition for continuous and discrete domains stems from approaches from Vered~\cite{vered2017heuristic,vered2018towards}.
These methods use an off-the-shelf planner to dynamically compute the probabilities of goal hypotheses as new observations arrive. 
While these methods use a heuristic to minimize the number of calls to the planner during the online phase, it still needs multiple calls to a full-fledged planner. 
These calls, however few, can be expensive, making this approach unsuitable for fast recognition under certain conditions. Recent research on goal recognition substantially improves efficiency for both domains~\cite{masters2017cost,ramonAIJ,pereira2017landmark}. 
However, these approaches formulate the problem in a discrete space using a planning formalism (typically STRIPS) or rely on a discretization of continuous state space~\cite{kaminka2018plan} or action space~\cite{fitzpatrick2021behaviour}.
By contrast, robotics applications where the agent's state includes specific configurations of a robot's degrees of freedom (position, translations, and angular rotation) cannot be trivially discretized.
Most approaches for online goal recognition in continuous domains focus on the path-planning problem, disregarding dynamics characteristics of the agents, so the plan consists only of a geometric collision-free path~\cite{masters2017cost,kaminka2018plan}. 
However, trajectory planning for collision-free paths requiring all dynamics and constraints of the agent is much more complex. 
Computing the probability of a single observation and hypothesis can take many minutes. 
Thus, such approaches are unsuitable for robots recognizing the goals of other robots while in motion since relying on calls to a motion planner incurs an unacceptable computational cost~\cite{jain2021optimal}.

We address these problems through a novel formulation for online goal recognition that works in continuous (trajectory planning problem) and discrete (STRIPS) domains.
Our contribution is fourfold.
First, we develop an online inference method to compute the probability distribution of the goal hypotheses based on the work of Ramírez~\cite{ramirez2010probabilistic} and Masters~\cite{masters2017cost} that obviates the need for calls to a planner during run-time.
We base our inference method on the Euclidean distance between a pre-computed sub-optimal trajectory and observations of the actual agent.
Second, we employ a predefined approximation of the agent's motion model for motion applications that obviates the need for costly computations of sub-optimal paths. 
The two contributions described above are mainly responsible for reducing the computational burden by orders of magnitude.
Third, we adapt our continuous inference to recognize goals in discrete domains using Euclidean distance as measure, overcoming a limitation of previous approaches for goal recognition.
Fourth, we extend our inference process so that it generates multiple paths for the same goal to account for different approximately similar alternative paths towards the same goal.    
We show empirically that this improves the general quality of the recognition process.   

\section{Online Goal Recognition}
\label{sec:online_goal_recognition}

We adapt notation from Vered~\cite{vered2017heuristic}, and expand it with the notion of time critical to real applications, i.e., robotics. 
An online goal recognition (GR) problem is a quintuple $R:= \langle W,I,G,M,O \rangle$, where $W: \mathbb{R}^{n}$ is an $n$-dimensional continuous state space, e.g., position, angles, velocity, acceleration, time, etc.\footnote{For our experiments in Continuous Domains we use $n=10$.} 
By convention, we always use time as the last dimension and omit time in examples in which time is irrelevant (e.g., initial states). 
For improved readability, we denote a particular state sampled at a discrete-time $t_k$ as $w[t_k] \in \mathbb{R}^{n}$.  
$I \in W$ is the initial state of the agent in the zero-time step. 
$G$ is a set of partial states where, depending on the goal recognition problem, the goal state may be missing different types of information, which represent goal hypotheses with size $\mathcal{N}$, where $g_n \in G$ is a particular goal state in the set. 
$M$ is a set comprising all possible trajectories.
Finally, $O \subseteq M$ is a discrete set of observations representing snapshots through the real trajectory.
As an online problem, the size of the observation set increases as the agent receives new information, where $o[t_k] \in O$ is an observation sampled at time $t_k$.

A trajectory is a sequence of $T$ discrete timed states as $m_I^{g_n} = \left[ w[0], w[1], \ldots, w[T-1] \right]$, i.e., it is the set of states $m_I^{g_n} \subset W$ ordered by the last component of each $w \in m_I^{g_n}$. 
$M_I^{g_n} \subseteq M$ is the set of all possible trajectories starting in the initial state $I$ and finishing in $g_n$; $m_I^{g_n} \in M_I^{g_n}$ is one particular trajectory to a goal $g_n$; 
$R$ is an offline problem when the final discrete time step $t_k$ is known; otherwise, $R$ is an online problem.
A solution to the online GR problem $R$ is a hidden goal $g_n$ reachable through a trajectory $m_I^{g_n}$, and which maximizes the conditional probability $P(m_I^{g_n} \mid O)$ of a trajectory given the current observation set $O$ defined in Eq.~\ref{eq:arg}, where ${m_I^{g_n}}^R$ is the trajectory that best explains the observations. 
\begin{equation} \label{eq:arg}
    {m_I^{g_n}}^R = \argmax_{m_I^{g_n} \in M} P(m_I^{g_n} \mid O)
\end{equation} 
This formulation is similar to Ramírez~\cite{ramirez2009plan}---the main difference is that we search for a trajectory $m_I^{g_n}$ instead of a plan. 
Instead of computing the trajectory's probability as proportional to the cost differential between an optimal trajectory and the observed one, we compute an $m_I^{g_n}$ that matches the observations and maximizes $P(m_I^{g_n} \mid O)$. 
This formulation makes three key assumptions: i) agents pursue a single goal that does not change over time; ii) agents always prefer the cheapest cost trajectories; and iii) all goal hypotheses are mutex, i.e., all are different and the agent pursues exactly one. 
\begin{align} \label{eq:P}
    P(m_I^{g_n} \mid O) =& \rho P(O \mid m_I^{g_n})P(m_I^{g_n}) \\
             =& \rho P(O \mid m_I^{g_n})P(m_I^{g_n} \mid g_n)P(g_n) \nonumber
\end{align}
Thus, we compute the conditional probability in Eq.~\ref{eq:P}, where $P(g_n)$ is a uniform distribution indicating the probability that the robot is pursuing the goal $g_n$; $\rho$ is a normalization to avoid computing $P(O)$. 
To solve Eq.~\ref{eq:P}, we need to compute $P(m_I^{g_n} \mid g_n)$ and $P(O \mid m_I^{g_n})$. 
We can compute $P(m_I^{g_n} \mid g_n)$ by synthesizing an optimal trajectory hypothesis ${m_I^{g_n}}^*$ that disregards the observations and aims for the final goal $g_n \in G$. 
In our case, we assume that if there are multiple optimal paths $m_I^{g_n*}$, then $P({m_I^{g_n*}} \mid g_n)$ is uniform (i.e. the agent picks any optimal trajectory at random). 
Otherwise, the probability $P({m_I^{g_n*}} \mid g_n)$ is always one.
Computing $P(O \mid m_I^{g_n})$ is a challenging task since we do not have a Probability Density Function (PDF) for the observations. 
A common approach for this situation is approximating the probability distribution for a parametric function considering an assumption about their characteristics; for example, consider an assumption that the probability $P(O \mid m_I^{g_n})$ increases as the trajectory gets closer to the observations \cite{hogg2019introduction}.
Thus, Eq.~\ref{eq:P_prop} approximates the conditional probability of $P(O \mid m_I^{g_n}$, where $dist(\cdot,\cdot)$ is a spatial distance calculation, e.g., Euclidean distance; $N \in \mathbb{N}$ is the actual number of observations at this moment; 
${m_I^{g_n}}^*[t_k]$ and $o[t_k]$ are the state in the optimal trajectory and the observation at the discrete step-time $t_k$. 
$\mathcal{T}$ is a set containing the discrete timestamp $t_k$ of all observations, i.e., we can sample states with a synchronized timestamp. 
Thus, Eq.~\ref{eq:P_prop} computes the conditional probability not only under full observability but also under partial observability, using the timestamp set $\mathcal{T}$ to match observations with their respectively optimal trajectory state.
The semantics of the timestamps depends on whether the domain is continuous or discrete. 
In continuous domains, we use a local clock to attach timestamps to each observation as it arrives. 
Note that continuous domains inherently imply partial observability, since there is an infinite number of possible states between each observation.  
By contrast, in discrete domains (STRIPS), goal recognition methods do not attach timestamps to observations, and assume a total order between observations. 
Here, we consider partial observability to be equivalent to the problem of online recognition, that is observations are consecutive and sequentially sampled, with the missing observations corresponding to the suffix of the full plan.

The conditional probability $P(O \mid {m_I^{g_n}}^*)$ must increase as the observations $O$ get closer to an optimal trajectory ${m_I^{g_n}}^*$. 
Thus, the value of $P(O \mid {m_I^{g_n}}^*)$ represents the probability that the observations belong to trajectory ${m_I^{g_n}}^*$ and Eq.~\ref{eq:P_prop} decays exponentially as the average error increases among all terms in the observations set $O$ and the optimal state trajectory in the same instant of discrete-time $t_k$.
\begin{align}\label{eq:P_prop}
    P(O \mid {m_I^{g_n}}^*):= 1 - \mathrm{e}^{\left(-1/\dfrac{1}{N}{\displaystyle \sum_{t_k \in \mathcal{T}} dist(o[t_k],{m_I^{g_n}}^*[t_k])}\right)},& \\
    n \in [1, \ldots, \cal{N}]& \nonumber
\end{align}
We can redefine the conditional probability of Eq.~\ref{eq:P} as Eq.~\ref{eq:P_modifi}, and the most likely goal hypothesis $g_n$ is that with the highest $P({m_I^{g_n}}^* \mid O)$ value.
An offline stage can compute optimal trajectories for each goal hypothesis, and at each new observation, we compare the observations with such trajectories to recognize the most likely goal hypothesis.
We do this by averaging the distances between an optimal trajectory and the observation samples using only the Eqs.~\ref{eq:P_prop}-\ref{eq:P_modifi}, with no online calls to a planner. 
Therefore, planner calls in online goal recognition depend only on the number of goal hypotheses in the set $G$, which differs from previous work that depends on the number of observations $O$ and goals $G$ in the sets.

We note that while Masters~\cite{masters2017cost} shows the last observation is enough to compute the conditional probability for path-planning problems, we average over the entire sequence of observations to improve the method's robustness to outliers and noisy observations.  
\begin{align} \label{eq:P_modifi}
    P({m_I^{g_n}}^* \mid O) = \rho P(O \mid {m_I^{g_n}}^*)P({m_I^{g_n}}^* \mid g_n)P(g_n),& \\
    n \in [1, \ldots, \mathcal{N}]& \nonumber
\end{align}
Computing the inference step in goal recognition by comparing states, instead of computing a cost function, can lead to problems in continuous domains.
Specifically, there is an infinite number of approximately optimal trajectories ${m_I^{g_n}}^*$ an agent can use to navigate between the initial state $I$ and the goal state $g_n$ while avoiding obstacles.
In some obstacle configurations, it is even possible that the agent has multiple optimal paths for reaching a goal. 
To deal with this problem, we based our process of goal inference on an average among a collection of solutions from the same problem.  
For this, we compute $k$ solution trajectories ${m_I^{g_n}}^*$ for problems comprised the common initial state and each goal hypotheses. 
Eqs.~\ref{eq:P_prop_multi} and \ref{eq:P_modifi_multi} summarize the averaging process above; ${\mathcal{M}_{g_n}[j]}^*$ defines a matrix containing all solutions trajectories ${m_I^{g_n}}^*$ where $j \in [1,\ldots,k]$.
\begin{align}\label{eq:P_prop_multi} \nonumber
    &P(O \mid {\mathcal{M}_{g_n}}^*):= \\ 
    &\frac{1}{k}\sum_{j=1}^{k}{1 - \mathrm{e}^{\left(-1/\dfrac{1}{N}{\displaystyle \sum_{t_k \in \mathcal{T}} dist(o[t_k],{\mathcal{M}_{g_n}}^*[j][t_k])}\right)}}, \\
    &\qquad \qquad \qquad \qquad \qquad \qquad \qquad \quad \,\, n \in [1, \ldots, \cal{N}] \nonumber
\end{align}
\begin{align} \label{eq:P_modifi_multi}
    P({\mathcal{M}_{g_n}}^* \mid O) = \rho P(O \mid {\mathcal{M}_{g_n}}^*)P({\mathcal{M}_{g_n}}^* \mid g_n)P(g_n),& \\
    n \in [1, \ldots, \mathcal{N}]& \nonumber
\end{align}

\section{Trajectory Planning Approximation}
\label{sec:continuous_domains}

Consider a common problem of robot navigation with obstacle avoidance and dynamic restrictions defined in Cartesian X and Y axes, where the continuous state domain is defined as $W = \left[x, y, \dot{x}, \dot{y}, t_k \right]$, i.e., positions, velocity and the discrete-time $t_k$ in both axes.
Our online approach to goal recognition needs an optimal trajectory ${m_I^{g_n}}^*[t_k] = [x, y]$ to compute probabilities for the goal hypotheses.
However, generating an optimal collision-free trajectory in continuous Cartesian space is a problem of trajectory planning that has a high computational cost~\cite{choset2005principles,jain2021optimal,pivtoraiko2009differentially}, which we mitigate by approximating the optimal trajectory ${m_I^{g_n}}^*$ and directly applying it to Eq.~\ref{eq:P_modifi}.
This problem becomes more challenging still if the environment is a dynamical system, and we want to compute an optimal trajectory~\cite{schwarting2018planning}.
In this paper, a dynamical system is an environment whose behavior can be described by sequential ordered differential equations~\cite{niku2020introduction}.
Recall that a trajectory $m_I^{g_n}$ is a sequence of states that describes the agent's movement within such a dynamical system. 
To navigate such system, the agent needs a policy that applies a correct control input that drives the states to the desired goal~\cite{niku2020introduction}.\footnote{In dynamical systems, a set of goal states is called a reference.} 
Computing a trajectory $m_I^{g_n}$ in any dynamical model requires motion planning to generate such control inputs over time.
We avoid running a motion planner by approximating ${m_I^{g_n}}^*$ through a polynomial model, which we compute using a method of trajectory generation from robotics~\cite{craig2005introduction}, which computes a trajectory using fewer motion parameters, consequently reducing the dimensionality of the optimization problem.

The motion parameters are the agent's desired dynamics characteristics or state at a specific time used to compute a full continuous trajectory between two points~\cite{lynch2017modern}. 
These motion parameters depend on the type of trajectory to be computed, e.g., linear, trapezoidal, and polynomial. 
In this paper, we use a polynomial trajectory that takes the desired time duration and position, velocity, and acceleration as motion parameters in the initial and final states of a trajectory.
Algorithm~\ref{alg:continuous_inference} describes the general framework of using a vector representation and a trajectory approximation in a continuous goal recognition process, which we detail in the following subsections.

\subsection{Polynomial Trajectory}

In an obstacle-free environment, we only need a single trajectory to describe the movement of an agent between two points. 
However, in an environment with obstacles, in most cases, a single low-polynomial trajectory cannot reach the goal without violating some restrictions. 
A common approach to generating such trajectories while keeping the polynomial degree low is to compute separate sub-trajectories that, when sequenced, form a complete trajectory between the initial and desired final state~\cite{lynch2017modern}.
To compute each sub-trajectory, we use the concept of a via point, which is a vector that includes some motion parameters, as shown in Eq.~\ref{eq:term_viapoints}, where $i \in \left[ 1,\, 2,\, \ldots,\, q \right]$; $q \in \mathbb{N}^*$ is the number of via points in the trajectory; the terms in the vector $v_i$ are the position, velocity, acceleration in the respectively Cartesian axes per via point $i$, and $td_i$ is the time duration of a single sub-trajectory between via points $i$ and $i+1$.
Two via points have all the parameters to compute a single trajectory. 
We define a sequence of via points to generate a complete trajectory, starting with the desired initial state $I$ and ending with the final goal state $g$, which we define in Eq.~\ref{eq:viapoints}. 
\begin{gather} \label{eq:term_viapoints}   
    v_i = \begin{bmatrix}
        x_i & y_i & \dot{x}_i & \dot{y}_i & \ddot{x}_i & \ddot{y}_i & td_i
    \end{bmatrix}^T \\ \label{eq:viapoints}
    V|_I^{g_n} = \begin{bmatrix}
        v_1 & v_2 & \ldots & v_q
    \end{bmatrix}
\end{gather}
Next, we compute a trajectory between each via point of Eq.~\ref{eq:viapoints} using a model that approximates the agent dynamics. 
We chose a fifth-degree polynomial for two key reasons. 
First, this type of trajectory can handle the agent's dynamic constraints, such as position, velocity, and acceleration. 
Second, certain fifth-degree polynomials are the most complex polynomials without discontinuities that have computationally cheap analytical solutions. 
Eqs.~\ref{X_i}-\ref{ddotY_i}~\cite[Ch. 7]{craig2005introduction} define the resulting model, where $x|_i^{i+1}(t)$ and $y|_i^{i+1}(t)$ are the positions, $\dot{x}|_i^{i+1}(t)$ and $\dot{y}|_i^{i+1}(t)$ are the velocities, and $\ddot{x}|_i^{i+1}(t)$ and $\ddot{y}|_i^{i+1}(t)$ are the accelerations in instant of time $t$ in X and Y Cartesian axes; $a_{fi}$ and $b_{fi}$ are unknown coefficients where $f \in [1,2, \ldots, 5]$. 
$x|_i^{i+1}(t)$ refers to the trajectory on the X axis (and respectively $y|_i^{i+1}(t)$ on the Y axis) with initial in $v_i$ and final in $v_{i+1}$ via points.
\begin{gather}\label{X_i}
    \hspace{-1em} x|_i^{i+1}(t) = a_{5i}t^5 + a_{4i}t^4 + a_{3i}t^3 + a_{2i}t^2 +   a_{1i}t + a_{0i} \\ \label{Y_i} 
    y|_i^{i+1}(t) = b_{5i}t^5 + b_{4i}t^4 + b_{3i}t^3 + b_{2i}t^2 + b_{1i}t + b_{0i} \\ \label{dotX_i}
    \dot{x}|_i^{i+1}(t) = 5a_{5i}t^4 + 4a_{4i}t^3 + 3a_{3i}t^2 + 2a_{2i}t +   a_{1i} \\ \label{dotY_i} 
    \dot{y}|_i^{i+1}(t) = 5b_{5i}t^4 + 4b_{4i}t^3 + 3b_{3i}t^2 + 2b_{2i}t +   b_{1i} \\
    \label{ddotX_i}
    \ddot{x}|_i^{i+1}(t) = 20a_{5i}t^3 + 12a_{4i}t^2 + 6a_{3i}t + 2a_{2i} \\ \label{ddotY_i} 
    \ddot{y}|_i^{i+1}(t) = 20b_{5i}t^3 + 12b_{4i}t^2 + 6b_{3i}t + 2b_{2i} 
\end{gather}

The coefficients $a_{wi}$ and $b_{wi}$ in the model trajectory of Eqs.~\ref{X_i}-\ref{ddotY_i} can be computed analytically by Eqs.~\ref{eq:ai5}-\ref{eq:ai0}, where $x_{i}$ and $x_{i+1}$ are the position; $\dot{x}_{i}$ and $\dot{x}_{i+1}$ are the velocity; $\ddot{x}_{i}$ and $\ddot{x}_{i+1}$ are the acceleration in via points $v_i$ and $v_{i+1}$, respectively.
For space, we omit the computation of the coefficients $b_{wi}$ of Eqs.~\ref{Y_i}, \ref{dotY_i}, and \ref{ddotY_i}, as they are exactly as above, but changing the work axes to Y.
\begin{gather}
    \nonumber
    \label{eq:ai5} a_{5i} = \frac{1}{2{td_i}^5} \big[ {td}_i\left[(\ddot{x}_{i+1} - \ddot{x}_{i})td_i - 6(\dot{x}_{i+1} + \dot{x}_{i})\right], \\ 
    + 12(x_{i+1} - x_{i}) \big], \\ 
    \nonumber
    a_{4i} = \frac{1}{2{td_i}^4} \big[ td_i\left(16\dot{x}_{i} + 14\dot{x}_{i+1} + (3\ddot{x}_{i} - 2\ddot{x}_{i+1})td_i\right) \\
    + 30(x_{i} - x_{i+1}) \big], \\ \nonumber
    a_{3i} = \frac{1}{2{td_i}^3} \big[ td_i\left((\ddot{x}_{i+1} - 3 \ddot{x}_{i})td_i - 8\dot{x}_{i+1} - 12\dot{x}_{i} \right) \\
    + 20(x_{i+1} - x_{i})\big], \\ \label{eq:ai0}
    a_{2i} = \frac{\ddot{x}_{i}}{2}, \,\,\, a_{1i} = \dot{x}_{i}, \,\,\,  a_{0i} = x_{i}.
\end{gather}
After computing each trajectory of Eqs.~\ref{X_i} and \ref{Y_i} using the via points sequence of Eq.~\ref{eq:viapoints}, we can concatenate all of them to synthesize a full path between initial and goal states as shown in Eqs.~\ref{eq:qx} and \ref{eq:qy}, where $\otimes$ is the concatenation operator; $X|_i^{i+1}$ and $Y|_i^{i+1}$ are the vectors from Eqs.~\ref{X_i} and \ref{Y_i}, respectively. 
Thus, we derive the approximate trajectory by sequencing the vectors as Eq.~\ref{eq:m_g_optimal}.
\begin{gather}\label{eq:qx}
    X|_1^q = X|_1^2 \otimes X|_2^3 \otimes \ldots \otimes X|_{q-1}^q \\ \label{eq:qy}
    Y|_1^q = Y|_1^2 \otimes Y|_2^3 \otimes \ldots \otimes Y|_{q-1}^q \\
    \hat{m}_I^{g_n} = \begin{bmatrix} \label{eq:m_g_optimal}
        X|_1^q & Y|_1^q
    \end{bmatrix}
\end{gather}

\subsection{Finding via point parameters}

\begin{algorithm}[tb]
\caption{Continuous Online Goal Recognition in Vector Representation}\label{alg:continuous_inference}
\begin{algorithmic}[1]\small
\Function{{\nobreak}ContinuousVecInference}{$W, I, G, V_{max}, k$}
    \For{$\textbf{all} \, {g_n} \in \mathcal{G}$} \Comment{Precompute approximate trajectories}
        \label{alg:continuous_inference:loop1start}
        \State $\mathcal{M}_{g_n} \gets \emptyset$ \Comment{Vector to save all solution trajectories}
        \State $\mathcal{V} \gets$ generate $k$ position parameters $RRT^*(W, I, g_n)$ \label{alg:continuous_rrt}
        \For{$\textbf{each}~\text{trajectory} \, V \vert_{I}^{g_n} \in \mathcal{V}$ indexed by $i$}
            \State $q \gets \vert V \vert_{I}^{g_n} \vert$
            \State $u \gets$ $\Call{Optimization}{V \vert_{I}^{g_n}, W, V_{max}}$ \label{alg:line_mu}
            \For{$\textbf{all} \, {j} \in 0,\ldots, q-1$} \label{alg:continuous_line_for}
                \State $V \vert_{I}^{g_n}[j]$ $\gets$ insert remaining parameters $u[j]$ \label{alg:continuous_line_V}
            \EndFor
            \State $\hat{m}_{I}^{g_n} \gets \Call{computePolyTrajectory}{V \vert_{I}^{g_n}}$ \label{alg:continuous_m_i^g}
            \State $\mathcal{M}_{g_n}[i] \gets \hat{m}_{I}^{g_n}$ \label{alg:continuous_M}
        \EndFor
    \EndFor\label{alg:continuous_inference:loop1end}
    \While{$New \, o_k \in O$ is available}  \Comment{Online recognition} \label{alg:continuous_line_while}
        \For{$\textbf{all} \, g_n \in \mathcal{G}$}
            \State $P(\mathcal{M}_{g_n} \mid O) \gets\Call{ComputePr}{O, \mathcal{M}_{g_n}}$\label{alg:continuous_line_compute}
        \EndFor
    \EndWhile\label{alg:continuous_line_while:end}
    \State \Return $\argmax_{g_n} P(\mathcal{M}_{g_n} \mid O)$
\EndFunction
\end{algorithmic}
\end{algorithm}

Algorithm~\ref{alg:continuous_inference} brings together the computations necessary to compose full trajectories from via points.
While we present the algorithm as a single function, in practice, the loop in Lines~\ref{alg:continuous_inference:loop1start}--\ref{alg:continuous_inference:loop1end} takes place offline, whereas only the inference of Lines~\ref{alg:continuous_line_while}--\ref{alg:continuous_line_while:end} takes place at recognition time. 
The algorithm starts by generating $k$ different sequences of via point position parameters for all goals in Lines~\ref{alg:continuous_inference:loop1start}--\ref{alg:continuous_rrt}.
To compose the trajectory using the via points from Eq.~\ref{eq:viapoints}, we need to find all motion parameters of each via point in the sequence $V|_I^{g_n}$, i.e., position, velocity, acceleration, and time duration. 
The position can be obtained through an optimal geometric planner such as Rapidly-exploring Random Trees (RRT$^{*}$)~\cite{karaman2011sampling}, Batch Informed Trees (BIT$^{*}$)~\cite{gammell2020batch}, and Sparse Roadmap Spanner (SPARS)~\cite{dobson2014sparse}, for example.
Given the initial position $I$ and the goal position $g_n$, a geometric planner can produce a sequence of via point position parameters, even in an environment with obstacles.
As our approach uses multiple solutions in the inference process, Line~\ref{alg:continuous_rrt} in Algorithm~\ref{alg:continuous_inference} generates $k$ different sequences of via point position parameters for the same goal hypothesis and store them in a vector $\mathcal{V}$.

Line~\ref{alg:line_mu} in Algorithm~\ref{alg:continuous_inference} computes the remaining via point parameters of $V|_I^{g_n}$ by an RL-based optimization defined by Eqs.~\ref{eq:policy_improvement1}-\ref{eq:policy_improvement2} that enforces the agent's dynamic constraints~\cite{bertsekas2019reinforcement}, where $h(s_i,\, u_i,\, s_{i+1}) = td_i$ is the cost function; $V_{max}$ is a maximum velocity vector for the trajectory; 
$s_i$ and $u_i$ are states and actions set defined by the Eqs.~\ref{eq:states_infinite}-\ref{eq:actions_infinite}, where $s_i,\, u_i \in \mathbb{R}^{4}$. 
Here, with one set of states $s_i$, actions $u_i$, and position via points in $v_i$ and $v_{i+1}$, it is possible to compute a single trajectory of $X|_i^{i+1}$ and $Y|_i^{i+1}$ from Eqs.~\ref{eq:qx}-\ref{eq:qy}. 

\begin{equation}\label{eq:policy_improvement1}
        u^* = \argmin_{u \in U} \left( J_u(s_{i}) \right),
\end{equation}
\begin{equation} \label{eq:costfunction_viapoint}
        J_{\mu}(s_i) = h(s_i, u_i, s_{i+1}) + J_{\mu}(s_{i+1}) \,\,\ i = 1,\, 2, \ldots, \, q-1.
\end{equation}
\begin{equation}\label{eq:policy_improvement2}
        subj. \,\, to:  ||\left[ \dot{x}|_i^{i+1}(t) \,\, \dot{y}|_i^{i+1}(t) \right]|| \leq V_{max},  \,\, \forall t \in (0, \,\, td_i] 
\end{equation}
\begin{equation} \label{eq:states_infinite}
    s_i =  
    \begin{bmatrix}
        \dot{x}_i & \dot{y}_i & \ddot{x}_{i} & \ddot{y}_{i} 
    \end{bmatrix}^T,
\end{equation}
\begin{equation}
    u_i = 
    \begin{bmatrix} \label{eq:actions_infinite}
        \dot{x}_{i+1} & \dot{y}_{i+1} & \ddot{x}_{i+1} & \ddot{y}_{i+1} & td_i
    \end{bmatrix}^T
\end{equation}

We optimize to find high velocities while penalizing violation of its maximal constraint and to minimize the trajectories' overall time duration $td_i$. 
The optimization process from Eqs.~\ref{eq:policy_improvement1}-\ref{eq:policy_improvement2} is often done incrementally.
This requirement is due to the continuous states from the formulation.  
The resulting iterative optimization process finds the velocities, accelerations, and time duration terms of Eq.~\ref{eq:viapoints} for each $v_i$.
We can use them to compose each via point vector $v_i$ of Eq.~\ref{eq:term_viapoints} and finally build the $V \vert_{I}^{g_n}$ matrix from Eq.~\ref{eq:viapoints} that contains all via points necessary to compute a complete trajectory. 
Lines~\ref{alg:continuous_line_for} and \ref{alg:continuous_line_V} of Algorithm~\ref{alg:continuous_inference} corresponds to this process. 
Using the $V \vert_{I}^{g_n}$ matrix, Algorithm~\ref{alg:continuous_inference} computes each trajectory sequence of Eqs.~\ref{eq:qx}-\ref{eq:qy} and composes the full trajectory $\hat{m}_{I}^{g_n}$ of Eq.~\ref{eq:m_g_optimal} as function \textsc{ComputePolyTrajectory} in Line~\ref{alg:continuous_m_i^g}. 
Vector $\mathcal{M}_{g_n}$ stores the full trajectory $\hat{m}_{I}^{g_n}$ in Line~\ref{alg:continuous_M}.
Lines~\ref{alg:continuous_line_while}-\ref{alg:continuous_line_compute} constitute the goal inference step at run-time with each new observation using the \textsc{ComputePr} function, which implements Eqs.~\ref{eq:P_prop_multi}-\ref{eq:P_modifi_multi} for each goal hypothesis, i.e., it computes the average conditional probability among all trajectories for each hypothesis.

\section{Experiments in Continuous Domains} \label{sec:experiments}
We use a simple but realistic simulation to compare our method with the state-of-the-art in online goal recognition for continuous domains.
Our experiments use 29 benchmark scenarios from Moving-AI~\cite{sturtevant2012benchmarks} based on Starcraft maps\footnote{\texttt{http://movingai.com/benchmarks/sc1/index.html}}. 
All scenarios comprise a 512x512 pixel map and the group of points in axes X and Y used as goal hypotheses. 
Further details are available in the online supplement~\cite{tesch2023online}.
We conducted the experiments using a 2.2GHz six-core Intel i7 CPU with 24GB RAM, running Ubuntu 22.04.

Each scenario consists of a map and a number of points we can use as goal and initial states to define goal recognition problems.
Scenario maps are $10m \times 10m$ Cartesian $X$ and $Y$ spaces with walls acting as obstacles.  
Each scenario contains eight randomly spread spatial points $p_n \in \mathbb{R}^3$ where $n \in [1,\, \ldots,\, 8]$, each of which containing $x$ and $y$ coordinates plus an orientation in radians. 
We select random points until we have eight that comply with two constraints to avoid trivial goal recognition settings: they must be at least $23cm$ away from any wall and $2m$ away from every other point.

\begin{figure*}[tb]
    \centering
    \begin{subfigure}[t]{0.26\textwidth}
        \centering
        \includegraphics[width=\linewidth]{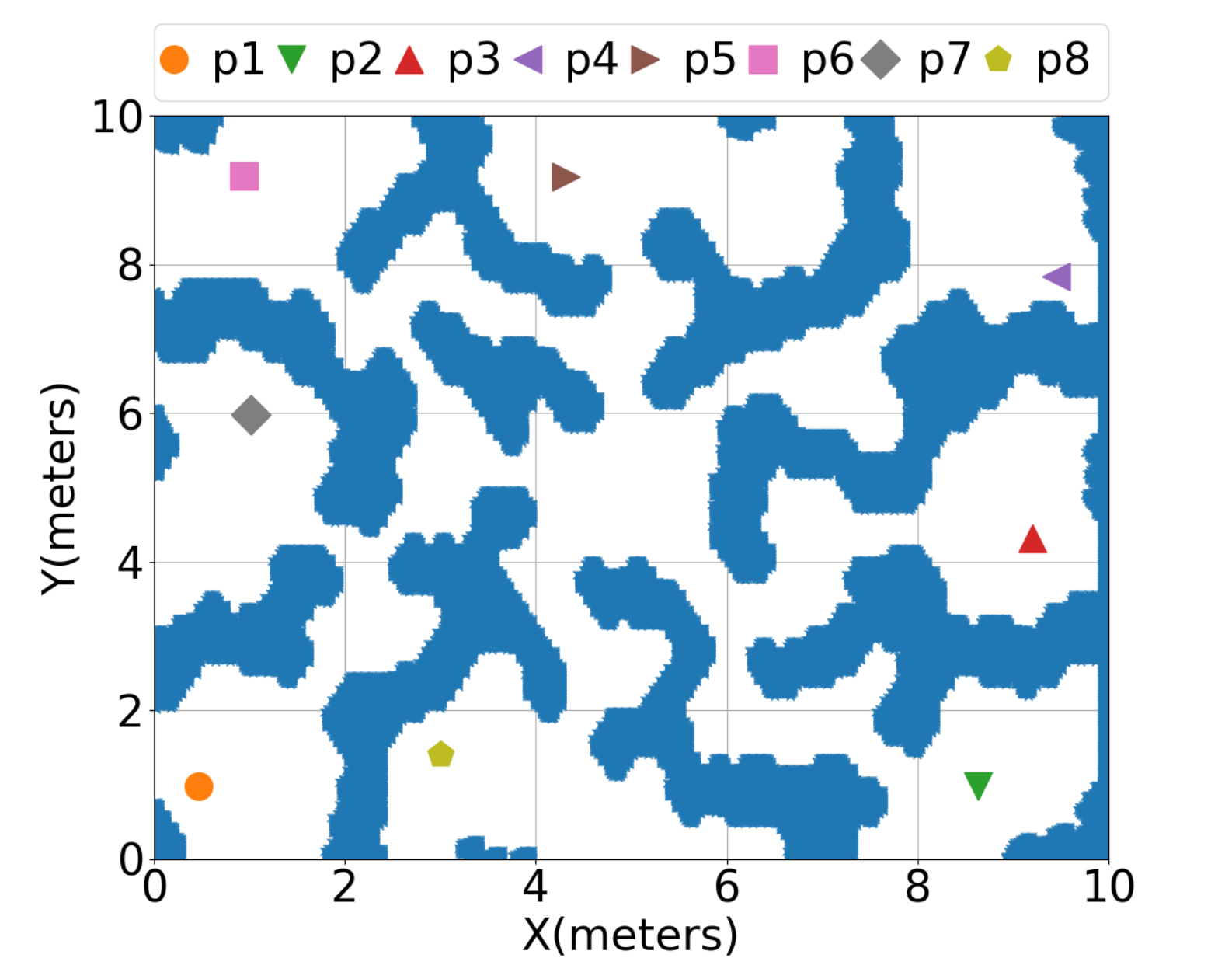}    
        \caption{Starcraft's BigGameHunters map. Marks represent potential positions.
        }
        \label{fig:biggamehunter}
    \end{subfigure}
    \centering
    \begin{subfigure}[t]{0.24\textwidth}
    \includegraphics[width=.9\linewidth]{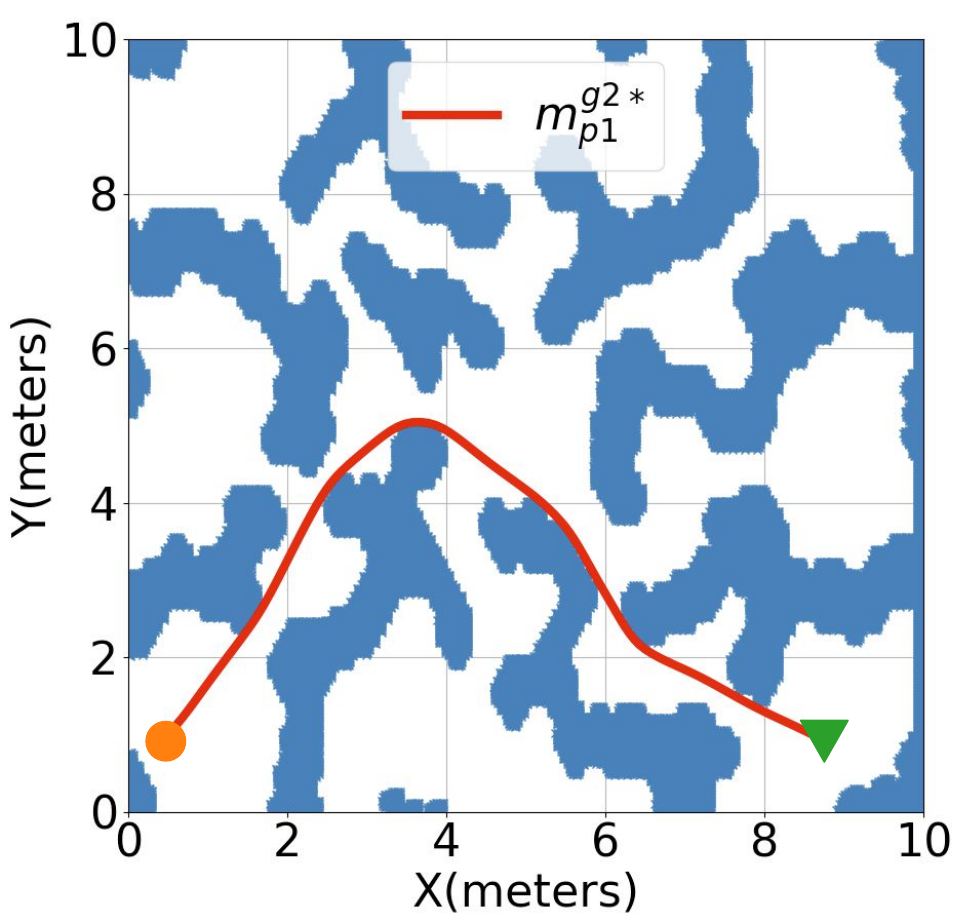}
    \caption{Robot optimal trajectory obtained thought optimization from initial point $p_1$ to goal point $g_2$.}
    \label{fig:ref-traj}
    \end{subfigure}
    \hfill
    \begin{subfigure}[t]{0.24\textwidth}
    \centering
    \includegraphics[width=0.9\linewidth]{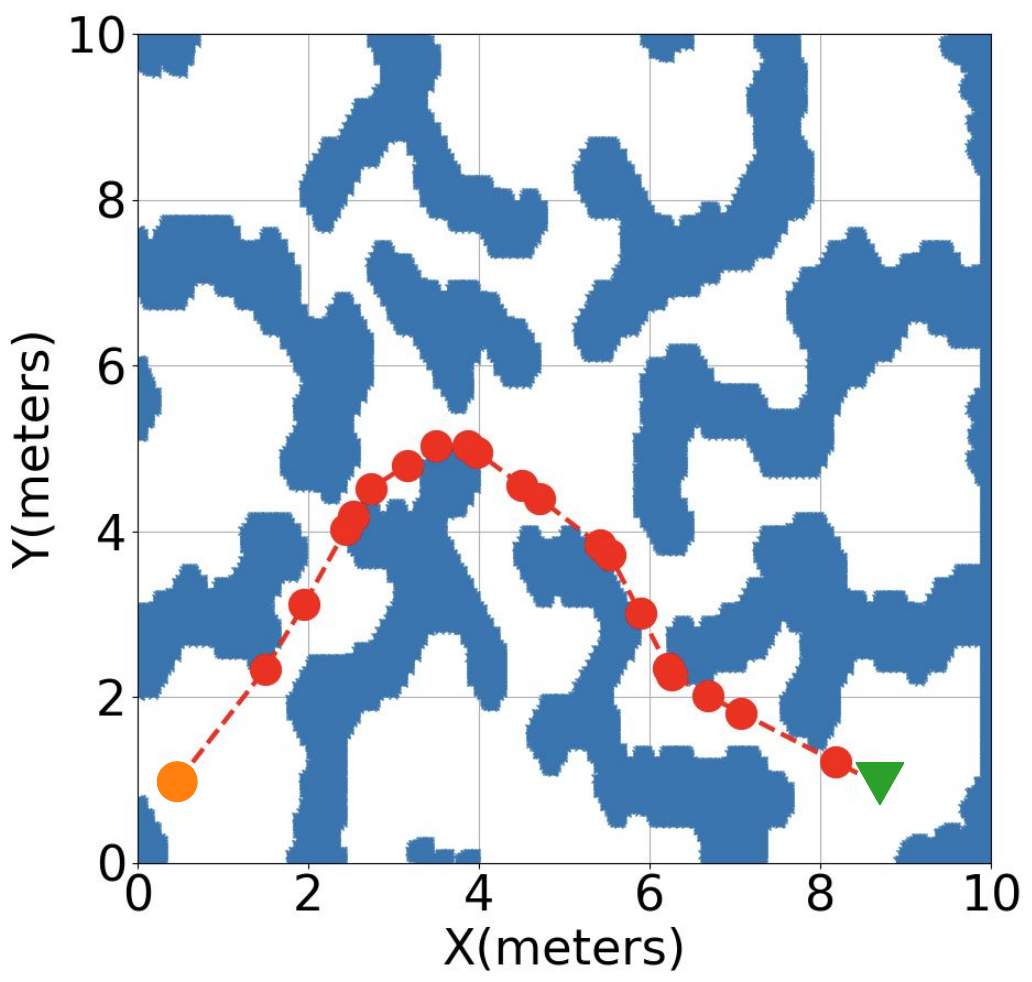}
    \caption{Example of output from ${RRT}^*$ considering initial and goal states as $p_1$ and $g_2$.}
    \label{fig:viapoints}
    \end{subfigure}
    \hfill
    \begin{subfigure}[t]{0.24\textwidth}
    \centering
    \includegraphics[width=.9\linewidth]{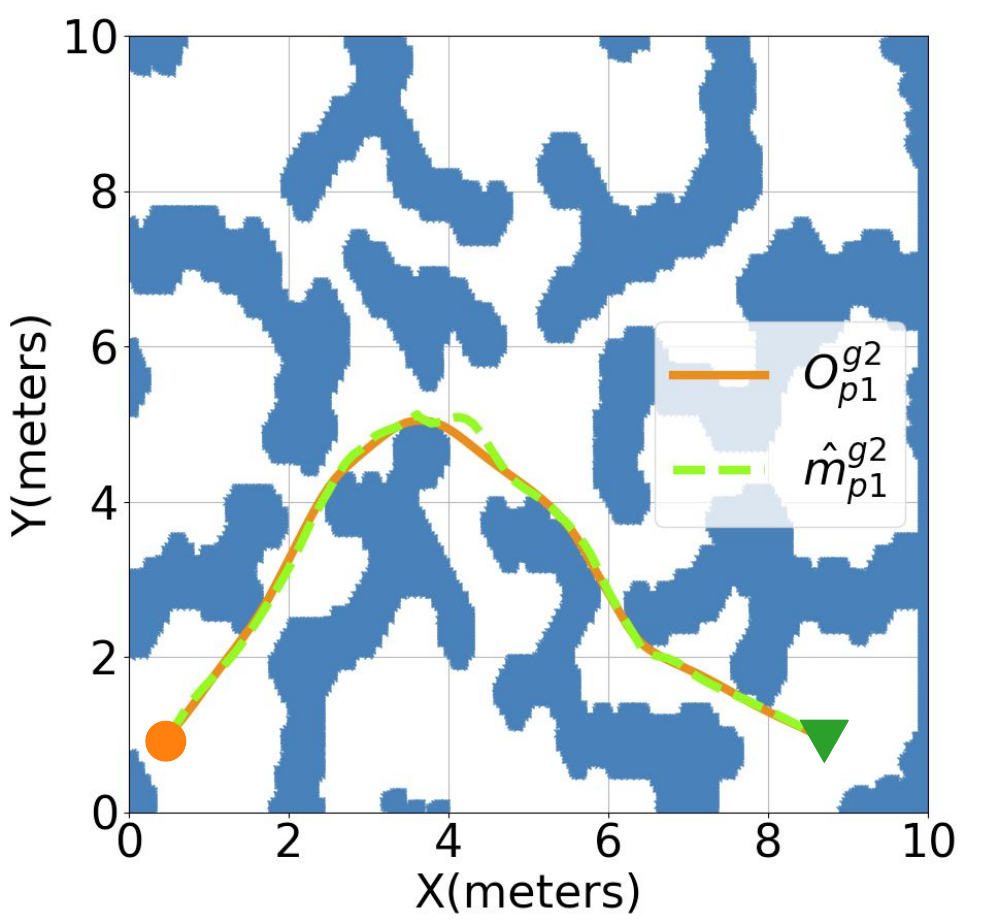}
    \caption{Comparison between the approximated trajectory $\hat{m}_{p_1}^{g_2}$ and the observation $O_{p_1}^{g_2}$.}
    \label{fig:obser}
    \end{subfigure}
    \vspace*{1em}
    \caption{Example of trajectories generated in different stages of the experimental scenario.}
    \vspace*{1em}
\end{figure*}

We use one scenario with eight points deliberately distributed around the map in Figure~\ref{fig:biggamehunter} as a working example. 
White represents traversable space, and marked colored points represent potential initial and goal position points.
The experiment samples observations from a simulation of a common robot with a two-wheeled motor and a unidirectional wheel defined by Eq.~\ref{eq:dynamicrobo}, where $\alpha(t)$ is the velocity control and $\omega(t)$ is the angular control rate. 
$x_r(t)$ and $y_r(t)$ are the positions in Cartesian axes and $\theta_r$ is the orientation in radians. 
We use a sampling period of $0.1$ seconds and disregard dynamics such as wheel friction, motor dynamics, and elastic deformations.
\begin{gather}\nonumber
    \dot{x}_r(t) = \alpha (t)cos(\theta(t)), \\\label{eq:dynamicrobo}
    \dot{y}_r(t) = \alpha (t) sin(\theta(t)), \\ \nonumber
    \dot{\theta}_r(t) =  \omega(t).
\end{gather} 
We generate recognition problems in each map using all combinations of the points in the scenario, with the remaining points being goal hypotheses. 
Thus, we have one problem where the ground truth is a trajectory from $p_1$ to $p_2$ (whose $x,y$ coordinates we call $g_2$), with $p_2$ to $p_8$ being goal hypotheses, another one with $p_8$ to $p_7$ (\textit{q.v.} $g_7$) with $p_1$ to $p_7$ as goal hypotheses, and so on. 
This yields 56 goal recognition problems per map using the Cartesian positions $x$ and $y$ of each point $p_n$ from Figure~\ref{fig:biggamehunter}. 

We assume agents optimize total motion time following the dynamical robot model of Eq.~\ref{eq:dynamicrobo} defined by Eqs.~\ref{eq:argminerro}-\ref{eq:mpc_contrans}, where $\mathit{tf}$ is the total simulation time; $\omega_{lim}$ is the maximal angular velocity; $g_n$ is a goal in Cartesian position $x, y$; 
$wall(x_r(t), y_r(t))$ is a function that measures the Euclidean distance from the robot position to its nearest wall obstacle at time $t$; $wall_{lim}$ is the minimum separation between the robot and an obstacle.
In our experiments $\omega_{lim} = 3$ rad/s, $g_n$ is sampled from $p_n$ with free angular orientation, and $wall_{lim}$ is $0.01$ meters for all experiments.
We represent the complete observation from the initial state $I$ to the goal state $g_n$ as $O_{I}^{g_n}$.
Figure~\ref{fig:ref-traj} exemplifies the robot's optimal trajectory obtained through optimization in the simulated environment. 
In this example, the robot is pursuing the goal point $g_2$ from the initial point $p_1$ and the recognition process has full observability so that $O_{I}^{g_n} = {m_{p_1}^{g_2}}^*$.
\begin{gather} \label{eq:argminerro}
    {\alpha}^*, {\omega}^* = \argmin_{\alpha,\,\,\omega} \mathit{tf} \\
    subj.  \,\, to: \,\,\,  \begin{bmatrix}
        x_r(tf) & y_r(tf)
    \end{bmatrix} = g_n  \\
    \lvert \alpha(t) \rvert \leq V_{max}, \\
         \lvert \omega(t) \rvert  \leq \omega_{lim}, \\ \label{eq:mpc_contrans}
        wall(x_r(t), y_r(t)) \geq wall_{lim} \,\, \forall \,\, t \in [0,\,\, tf]
\end{gather}
\subsection{Computing the Via Point Parameters}

We use the Open Motion Planning Library (OMPL) \cite{OMPL} with the ${RRT}^*$ geometric planning algorithm to compute all the position parameters of Eq.~\ref{eq:viapoints}. 
This off-the-shelf planner provides a cost-minimal sequence of via point positions from an initial position to a goal.
Calls to ${RRT}^*$ have a 5-second time limit and use distance as the cost function so that the via points are part of one shortest path to the goal.
Figure~\ref{fig:viapoints} shows an example of an output provided for the ${RRT}^*$ planner, where the initial and goal states are $p_1$ and $g_2$, respectively. 
Circles are the via points positions from ${RRT}^*$, and the dashed line connects them in sequence

Next, we need to find the velocity parameters of each via point. 
We implement the optimization from Eqs.~\ref{eq:policy_improvement1}-\ref{eq:policy_improvement2} in SciPy in its default configuration. 
The optimization settings are: maximum velocity vector of $V_{max} = 1$; random initial actions $u$; all acceleration terms in the via points being zero. 

\begin{figure}[tb]
    \centering
    \includegraphics[width=.9\linewidth]{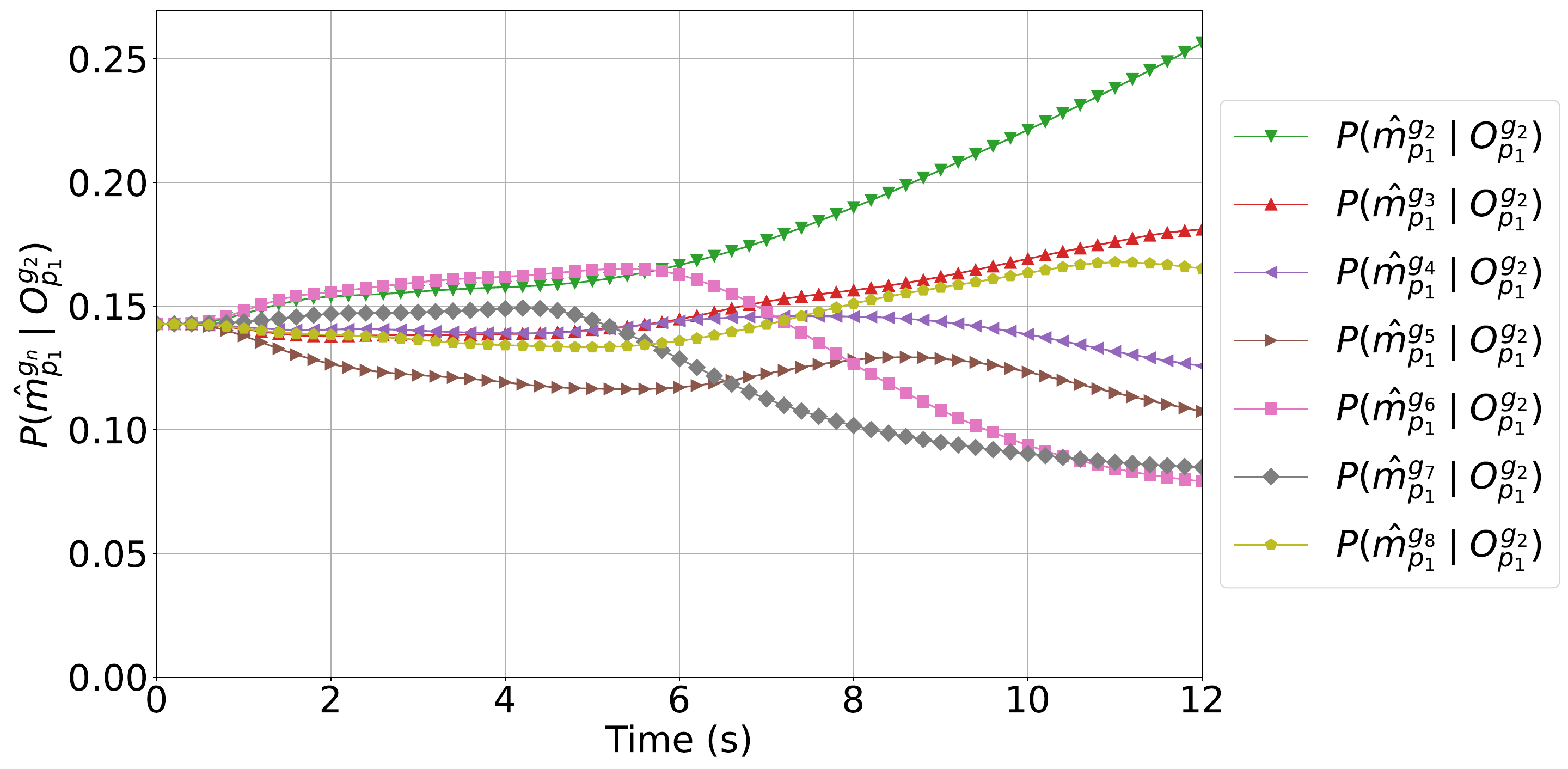}
    \caption{Conditional probabilities $P(\hat{m}_{p_1}^{g_n} \mid  O_{p_1}^{g_2})$ for all goals in problems starting at $p_1$ and goal point at $g_2$.}
    \label{fig:P(O|g)}
    \vspace*{1em}
\end{figure}

We use the via points to compute the approximate trajectory $\hat{m}_I^g$ from Eq.~\ref{eq:m_g_optimal}.
Figure~\ref{fig:obser} shows the trajectory difference between an estimated trajectory $\hat{m}_{p_1}^{g_2}$ and its respectively observation $O_{p_1}^{g_2}$.
The final stage of our method is to compute the conditional probability of Eq.~\ref{eq:P_modifi} using the estimate trajectories $\hat{m}_{I}^{g_n}$ instead of the optimal for each goal at each new observation, allowing us to infer the most likely goal hypotheses.
Figure~\ref{fig:P(O|g)} illustrates, in our working example, the conditional probability values $P(\hat{m}_{p_1}^{g_n} \mid  O_{p_1}^{g_2})$ of Eq.~\ref{eq:P_modifi} for all goal points in the set over time.

\subsection{Results}

Table~\ref{tab:planner_compare} compares our method (Vector) with the Mirroring of Kaminka~\cite{kaminka2018plan} (Mirroring) and the Recompute plus Prune (R+P) of Vered~\cite{vered2017heuristic}, in all goal recognition problems for all 28 scenarios. 
Mirroring requires $(|O|+1)|G|$ planner calls, whereas R+P requires between $|G|$ and $(|O|+1)|G|$ by heuristically deciding when to use the planner. 
The table shows average values over all scenarios and reports the positive predictive value (PPV) of the correct goal in percentage, the accuracy (ACC) of the predictions, the spread of goals in the output (SPR), the number of planner calls (PC) required to infer each goal, and the runtime of each method.
For each metric, we report their mean throughout the experiments and their standard deviation in parentheses. 
We separate the algorithm's online and offline parts to highlight the key advantage of our method. 
Table~\ref{tab:planner_compare} also shows the results of our method changing the number of solution trajectories $k$ used in inference. 
The supplementary material breaks down these results for each scenario.

\begin{table}[tb]
\centering
\fontsize{6.5}{7.8}\selectfont
\setlength{\tabcolsep}{1pt}
\begin{tabular}{c|c|c|c|c|c|cc}
\hline
 & PPV  & ACC & SPR & PC & Online & Offline  \\ 
 & (\%) &  (\%)   &     &    & Time(s)& Time(s) \\
 \hline
\multicolumn{1}{c|}{Mirr.} & $41.1 (9.3)$  & $85.2 (2.3)$ & $1.1 (0)$ & $49 (0)$ & $1.0e{4} (5.7e{3})$ & $2.4e{3} (1.6e{3})$ \\
\multicolumn{1}{c|}{R+P} & $44.3 (9.0)$  & $85.9 (2.2)$ & $1.1 (0.04)$ & $29.5 (2.7)$ & $4.8e{3} (3.0e{3})$& $2.3e{3} (1.5e{3})$ \\
\multicolumn{1}{c|}{Vec k=1} & $41.6 (7.8)$ & $85.4 (1.9)$ & $1.0 (0)$ & $7.0 (0)$ & $8.9\text{e-}2 (1.0\text{e-}{2})$ & $1.0e{2} (40.2)$ \\
\multicolumn{1}{c|}{Vec k=5} & $47.7 (9.2)$ & $86.9 (2.3)$ & $1.0 (0)$ & $7.0 (0)$ & $3.8\text{e-}2 (1.3\text{e-}3)$ & $1.7e{2} (53.6)$ \\
\multicolumn{1}{c|}{Vec k=10} & $48.5 (9.3)$ & $87.1 (2.3)$ & $1.0 (0)$ & $7.0 (0)$ & $4.1\text{e-}2 (1.4\text{e-}3)$ & $2.1e{2} (40.3)$ \\
\multicolumn{1}{c|}{Vec k=15} & $49.7 (8.6)$ & $87.4 (2.1)$ & $1.0 (0)$ & $7.0 (0)$ & $3.7\text{e-}2 (8.2\text{e-}4)$ & $2.2e{2} (41.2)$ \\
\multicolumn{1}{c|}{Vec k=20} & $49.8 (9.3)$ & $87.4 (2.3)$ & $1.0 (0)$ & $7.0 (0)$ & $3.8\text{e-}2 (8.0\text{e-}4)$ & $2.4e{2} (45.8)$ \\ \hline
\end{tabular}
\caption{Comparison among online goal recognition methods for continuous domains.}
\label{tab:planner_compare}
\end{table}

To compare online recognition performance over time, we divide each of the 56 observable trajectories (all combinations of initial states and goals) into six points equally spaced in time, named test observations. 
Thus, each problem has six sampled points used as observations ($\{ o_1, \dots, o_6\}$, omitting the final observation that indicates the goal), which we can use to measure online recognition accuracy (convergence to the correct goal) over time. 
To illustrate this, we go back to our working example with points deliberately distributed in the scenario from Figure~\ref{fig:biggamehunter}. 
Figure~\ref{fig:6obser} shows the six sample observations (red dots) over a complete trajectory (dashed line) with initial and goal states as $p_1$ and $g_2$. 
Figure~\ref{fig:margem_erro} shows the average goal recognition PPV (and its margin of error with a confidence level of 95$\%$) at each of the six sample observations for each method.
Results indicate that our method has better positive predictive value and accuracy within the standard deviation for all fractions of observations.

To better illustrate a case where the motion problem has more than one trajectory as a solution, we contrive a scenario where many combinations of initial states and goals will lead to two optimal, completely distinct trajectories. 
We compare our method Vector with Mirroring and R+P to show how each method deals with this challenging recognition problem. 
While the results for these problems are present in the aggregated results in this section, we provide further detail about them in the supplementary material~\cite{tesch2023online}. 

In conclusion, our approach using $k=1$ is competitive with the state of the art, with a marginal advantage in accuracy and positive predictive value within the standard deviation. 
Importantly, our approach offers a substantial speed-up.
While the offline computations have almost an order of magnitude speed up, the online computations improve by six orders of magnitude. 
Besides the improvements for the basic case of our algorithm, the addition of multiple solutions (i.e., increasing $k>1$) in the inference process brings an increase in accuracy and positive predictive value. 
The results show that with $k=5$, we widen the performance gap; increasing the $k$ value further away only brings a minor increase in performance, and we can see a stabilization around $k=15$.

\section{Discrete Domains}
\label{sec:discrete_domains}

Our approach so far works exclusively in continuous Euclidean state spaces represented as numeric vectors. 
While applying it directly to discrete domains is not trivial, we now show how converting the discrete state into a vectorial continuous state space representation allows us to apply this inference in such domains. 
STRIPS-style domains are the largest goal recognition datasets openly available~\cite{pereira2017landmark}.

We consider a discrete domain to be a STRIPS-style PDDL (Planning Domain Definition Language) description with the same semantics of Fikes et al.~\cite{fikes1971strips} as $\mathcal{D}$, with typed objects set $Z$ and a set of typed predicates $\mathcal{R}$. 
A classical planning problem can be described as $\mathcal{P} = \langle \mathcal{F}, \mathcal{A}, \mathcal{I}, \mathcal{G} \rangle$, where: $\mathcal{F}$ is the set of facts (instantiated predicates from Z); $\mathcal{A}$ is the set actions with objects from $Z$; $\mathcal{I} \in 2^\mathcal{F}$ is the initial state; and $\mathcal{G} \subseteq 2^\mathcal{F}$ is the goal state. 
A discrete goal recognition problem is a tuple $\mathcal{L} = \langle \mathcal{D}, \mathcal{O}, \mathcal{I}, \mathcal{G}\rangle$, where: $\mathcal{D}$ is the domain defined above; $\mathcal{O} \subseteq \mathcal{F}$ is a set of observations. 
A trajectory in a discrete domain is a sequence of ordered states, while a plan $\pi$ is a sequence of ordered actions. 
Plans here match exactly their definition from classical planning~\cite{GhallabNauTraverso2004}.

While not having the same problem of infinite possible approximately optimal plans (for non-zero action costs), discrete domains often have multiple optimal plans to achieve a goal from a given initial state. 
Therefore, we apply the same approach used in continuous domain inference in the discrete domain, i.e., we consider multiple solutions with the same goal hypotheses in our inference process. 
Our approach for discrete domains takes inspiration from~\citet{SohrabiRiabovUdrea2016} to use a Top-k planner to compute multiple plans. 
Top-k planners compute up to $k$ plans that are a solution to a problem $P$, if $P$ has more than $k$ possible solutions~\cite{SpeckMattmuellerNebel2020}.

Algorithm~\ref{alg:inference} summarizes our approach in discrete domains.
As with Algorithm~\ref{alg:continuous_inference}, we condense both the offline and online part in the same pseudocode, while in practice these are computed separately. 
Lines~\ref{alg:inference:loop1start}--\ref{alg:inference:loop1end} comprise the offline part of the algorithm, whereas Lines~\ref{alg:line:online_recognition}--\ref{alg:line:online_recognition:end} are the online part. 
The offline part of the algorithm starts by using the Top-k planner to generate $k$ plans $\pi_{g_n}$, for a goal hypothesis $g_n$ and store them into vector $\Pi$ (Line~\ref{alg:line:planner_call}). 
The algorithm then rolls out each solution plan $\pi_{g_n}$ from $\Pi$, generating a trajectory $m_I^{g_n}$ which it stores in vector $\mathcal{M}_{g_n}$. 
Lines~\ref{alg:line:rollout}--\ref{alg:line:M} represent this process. 
In the online inference, at each new observation, the algorithm computes the average conditional probability among all trajectories for each goal hypothesis following the Bayesian formulation of goal recognition. 
Like in the continuous domain, function \textsc{ComputePr} (Line~\ref{alg2:line:compute}) implements Eqs.~\ref{eq:P_prop_multi}-\ref{eq:P_modifi_multi} for each goal hypothesis. 

The difference from the continuous version of our approach is in the computation of the Euclidean distance of Eq.~\ref{eq:P_prop_multi}. 
To compute the distance between two sets of predicates (the observation and the computed state of a trajectory), we use the function of Eq.~\ref{eq:euclidiana_strips}. 
This computes the Euclidean distance between the two states directly from the STRIPS format of these states. Note that the expression within the modulo operator is the Symmetric difference.

\begin{algorithm}[tb]
\caption{Discrete Online Goal Recognition in Vector Representation}\label{alg:inference}
\begin{algorithmic}[1]\small
\Require $\mathcal{P} = \langle \mathcal{F}, \mathcal{A}, \mathcal{I}, \mathcal{G} \rangle$, $k$
\Function{onlineVectorInference}{$\mathcal{P}, k$}
    \For{$\textbf{all} \, {g_n} \in \mathcal{G}$} 
    \label{alg:inference:loop1start}
        \Comment{Precompute optimal plans}
        \State $\mathcal{M}_{g_n} \gets \emptyset$ \Comment{Vector to save all solution trajectories}
        \State $\Pi \gets \Call{Planner}{\mathcal{P}, g_n, k}$ \label{alg:line:planner_call} \Comment{Generate $k$ plans}
        \For{\textbf{each} plan $\pi_{g_n} \in \Pi$ indexed by $i$}
            \State $m_I^{g_n} \gets \Call{rollout}{\mathcal{P}, \pi_{g_n}}$ \label{alg:line:rollout}
            \State $\mathcal{M}_{g_n}[i] \gets m_I^{g_n}$ \label{alg:line:M}
        \EndFor
    \EndFor\label{alg:inference:loop1end}
    \While{$New \, o_k \in \mathcal{O}$ is available} \label{alg:line:online_recognition}\Comment{Online recognition}
        \For{$\textbf{all} \, g_n \in \mathcal{G}$}
            \State $P(\mathcal{M}_{g_n} \mid O)\gets\Call{ComputePr}{\mathcal{O}, \mathcal{M}_{g_n}}$\label{alg2:line:compute}
        \EndFor
    \EndWhile
    \label{alg:line:online_recognitionEnd}
    \State \Return $\argmax_{g_n} P(\mathcal{M}_{g_n} \mid O)$
\EndFunction\label{alg:line:online_recognition:end}
\end{algorithmic}
\end{algorithm}

\begin{equation} \label{eq:euclidiana_strips}
    dist = \sqrt{\mid (o -  m_I^{g_n}) \cup (m_I^{g_n} -  o))\mid}
\end{equation}

\section{Experiments in Discrete Domains}

\begin{figure*}[tb]
    \centering
    \begin{subfigure}[t]{0.24\textwidth}
    \includegraphics[trim={1cm 0cm 0.5cm 0.5cm},clip,width=.9\linewidth]{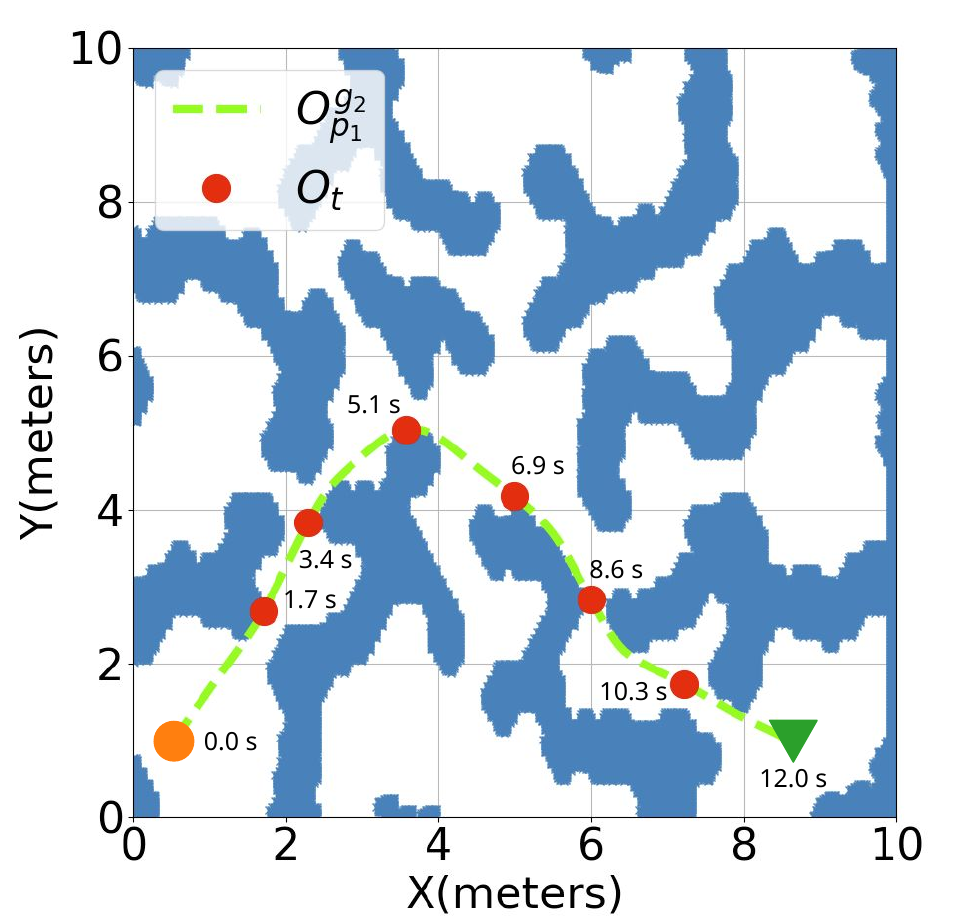}
    \caption{Trajectory example for $O_{p_1}^{g_2}$ with sampled observations.}
    \label{fig:6obser}
    \end{subfigure}
    \hfill
    \begin{subfigure}[t]{0.37\textwidth}
    \centering
    \includegraphics[trim={0cm 0cm 0cm 0cm},clip,width=1\linewidth]{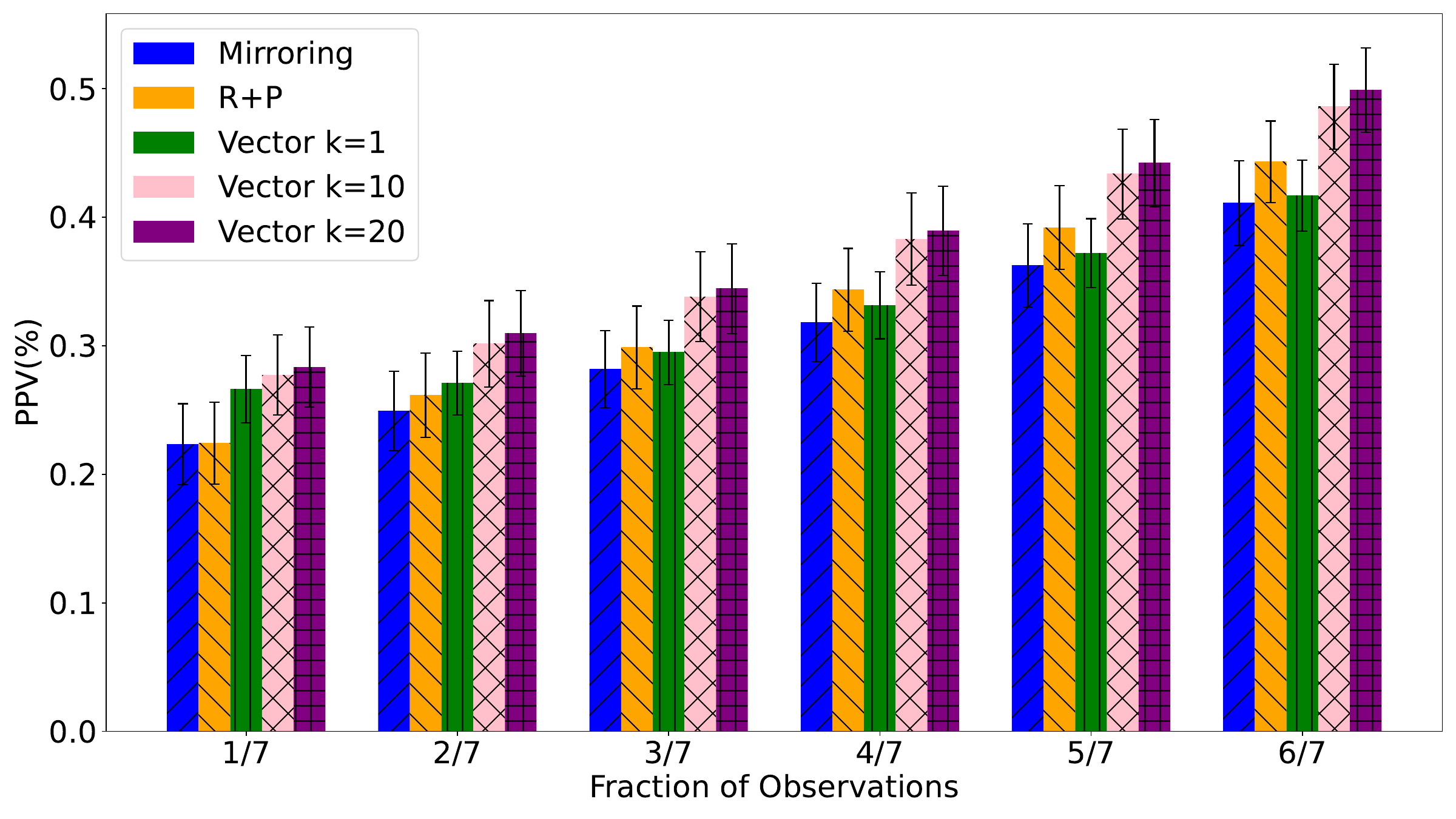}
    \caption{PPV percent and margin of error comparison from continuous domains over observations.}
    \label{fig:margem_erro}
    \end{subfigure}
    \hfill
    \begin{subfigure}[t]{0.37\textwidth}
    \centering
    \includegraphics[trim={0cm 0cm 0cm 0cm},clip,width=1.0\linewidth]{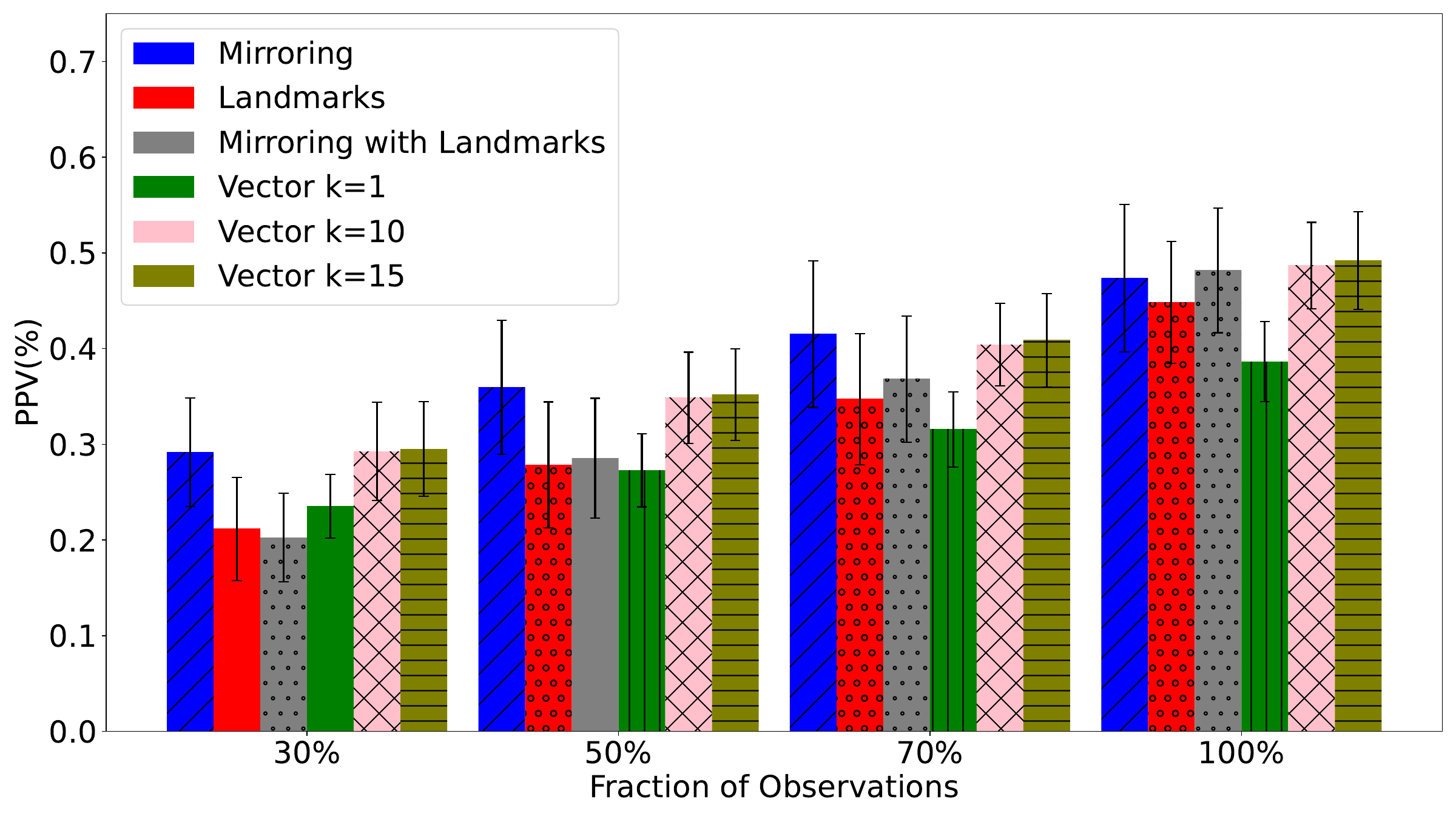}
    \caption{PPV percent and margin of error comparison from discrete domains over observations.}
    \label{fig:erro_discreto}
    \end{subfigure}
    \vspace*{1em}
    \caption{Comparison between on-line goal recognition methods.}
    \vspace*{1em}
\end{figure*}

We evaluate our inference method empirically against three online goal recognition methods. 
The first \textit{Mirroring} by Kaminka~\cite{kaminka2018plan} (Mirr.).
The second and third are \textit{Goal Recognition with Landmarks} (Land.) and \textit{Goal Mirroring with Landmarks} both by Vered~\cite{vered2018towards} (GM+L), where the last one is the current state-of-the-art.
Our experiments use an openly available goal and plan recognition dataset~\cite{pereira2017landmark}, which contains thousands of recognition problems comprising large and non-trivial planning problems in the STRIPS fragment of PDDL (with optimal and sub-optimal plans as observations), including domains and problems from datasets from ~\citeauthor{ramirez2009plan}~\citeyear{ramirez2009plan}. 
Domains include realistic applications (e.g., DWR, ROVERS, LOGISTICS), and hard artificial domains (e.g., SOKOBAN).

\begin{table}[tb]
\fontsize{6}{7.8}\selectfont
\setlength{\tabcolsep}{1pt}
\centering
\begin{tabular}{c|c|c|c|c|c|c}
\hline
 & PPV  & ACC & SPR & PC & Online & Offline  \\ 
 & (\%) & (\%)   &     &    & Time(s)& Time(s) \\
 \hline
Mirr. & $47.3 (13.6)$ & $81.8 (8.1)$ & $2.5 (1.2)$ & $124.8 (57.4)$ & $80.4 (117.6)$ & $4.6 (5.3)$ \\
Land. & $44.8 (11.2)$ & $82.7 (5.2)$ & $1.6 (0.51)$ & $0.0 (0)$ & $5.7\text{e-}1 (6.1\text{e-}1)$ & $1.3\text{e-}1 (2.6\text{e-}1)$ \\
GM+L      & $48.2 (11.5)$ & $84.7 (5.0)$ & $1.3 (0.47)$ & $29.9 (7.9)$ & $12.5 (12.5)$ & $4.8 (5.8)$ \\
Vec k=1   & $38.6 (7.4)$ & $76.3 (7.6)$ & $2.4 (0.52)$ & $8.2 (3.9)$ & $4.9\text{e-}3 (8.4\text{e-}3)$ & $6.8 (6.9)$ \\
Vec k=5    & $45.9 (7.6)$ & $81.7 (6.7)$ & $1.9 (0.44)$ & $8.2 (3.9)$ & $2.1\text{e-}2 (4.1\text{e-}2)$ & $14.5 (21.9)$ \\
Vec k=10   & $48.7 (7.9)$ & $83.7 (5.9)$ & $1.7 (0.45)$ & $8.2 (3.9)$ & $4.4\text{e-}2 (9.2\text{e-}2)$ & $23.5 (41.1)$ \\
Vec k=15   & $49.2 (8.9)$ & $84.2 (5.7)$ & $1.7 (0.44)$ & $8.2 (3.9)$ & $6.4\text{e-}2 (1.3\text{e-}1)$ & $33.66 (63.7)$ \\ \hline
\end{tabular}
\vspace*{1em}
\caption{Comparison among online goal recognition methods for discrete domains.} 
\label{tab:comparison_discrete}
\normalsize
\vspace*{1em}
\end{table}

Table~\ref{tab:comparison_discrete} shows the result of our empirical evaluation of these methods against our online goal recognition formulation for discrete domains. 
All results are averages over all problems in every experimented domain, and report positive predictive value (PPV); overall accuracy (ACC) for each experiment; spread (SPR) is the size of the hypothesis solution set chosen by recognition method and (PC) is the number of planner calls during the whole recognition process. 
The online supplement~\cite{tesch2023online} breaks the results down for each domain.
Table~\ref{tab:comparison_discrete} also shows results when our method uses $k$ solution plans during inference for $k \in \left[1,5,10,15 \right]$. 
These values correspond respectively to $3.0\%$, $15.2\%$, $30.4\%$, and $45.6\%$ of the total number of optimal solution plans on average over all problems.
Our implementation uses the SymK\footnote{\texttt{https://github.com/speckdavid/symk}} planner with the forward-search option. 
SymK is a state-of-the-art optimal Top-k planner based on symbolic search that extends Fast Downward~\cite{SpeckMattmuellerNebel2020}.

The results show that our approach can be competitive with the state-of-the-art while being much faster in inference (online) time, trading off in offline processing time. 
Likewise, the results in the continuous domain show that with $k=5$, our performance improves substantially; whereas increasing the $k$ value further only brings a minor increase in performance, and we can see a stabilization around $k=10$. 
The parameterization of $k$ does not consider whether all $k$ plans are indeed optimal. Indeed, increasing $k$ might not necessarily improve performance when the number of optimal plans for a problem is fewer than $k$. 
Additional experiments show that filtering the $k$ plans for optimality improves PPV by $4.5\%$. 

These results confirm that our method can be easily applied to discrete domains while retaining its strengths. 
Figure~\ref{fig:erro_discreto} compares results among the methods at four points during observation (and its margin of error with a confidence level of 95$\%$), at $30\%$, $50\%$, $70\%$, and $100\%$ of their respective full observation. 
All results shown in Figure~\ref{fig:erro_discreto} are averages over the problems of each domain, \emph{e.g.}, partial observation problems ($30\%$, $50\%$, and $70\%$) have an average of 12 problems for each domain, and the full observation has an average length of 12 actions.   
Results indicate that our approach is faster across the board in the online phase, using substantially fewer calls to the planner than all other approaches. 
Importantly, our approach provides superior accuracy and PPV, with a marginally higher spread.

\section{Related Work}
\label{sec:related_work}

Recent work in goal recognition follow the general approaches developed by Kaminka~\cite{kaminka2018plan} and Masters~\cite{masters2017cost}, which focuses on comparing the cost functions between an optimal trajectory and a trajectory following the observations. 
Fitzpatrick~\cite{fitzpatrick2021behaviour} has a different approach applied in a scenario without obstacles; the inference measures the error between a computed trajectory and the observations using Euclidean distance. 
Ignoring obstacles, however, is a major limitation for realistic (and widespread) ground navigation scenarios.  
Obstacle avoidance, however, creates additional challenges in goal recognition, for instance multiple sub-solutions, necessitating a goal recognizer to reason about different possible solutions available to an agent. 
This paper addresses such issues by reasoning over top-k plans.  

\section{Conclusion}
\label{sec:conclusion}

We introduce an approach for online goal recognition suitable in continuous and discrete domains. 
Our approach is suitable for recognizing agent motion in continuous environments with obstacles. 
If we know the observed agent's potential initial states and a set of possible goals a priori, we can execute the costliest computations in an offline stage. 
This allows computation of the inference process through a simple equation in milliseconds at each new observation, providing the probability distribution over the goals.
We develop a mechanism for discrete domains to convert STRIPS-style problems into vectors amenable to our function approximation.

Empirical evaluation shows our method is six orders of magnitude faster than the state-of-the-art in the online stage and four times faster than the state-of-the-art in the preprocessing stage for the continuous case. 
This advantage in execution time is due to our method using fewer planner calls ($|G|$), replacing most of the original planner calls with an approximate motion model. 
While our offline preprocessing time is higher than the fastest methods for discrete domains, it is the only approach with a similar runtime that works for both types of domains.
The major drawback of using an approximation is trading speed for accuracy. 
However, any method that takes minutes to perform online goal recognition in a moving robot is impractical.  
Thus, this paper sets us up to a new class of goal recognition methods suitable for applications in continuous domains where recognition must happen in milliseconds.

\section*{Acknowledgements}

This study was financed in part by the Coordena\c{c}\~ao de Aperfei\c{c}oamento de Pessoal de N\'ivel Superior Brasil (CAPES) - Finance Code 001.

\bibliography{ECAI24}

\section{Continuous Domain Results Breakdown}
\begin{figure*}[h]
    \centering
    
    \begin{subfigure}{0.45\textwidth}
        \includegraphics[width=\linewidth]{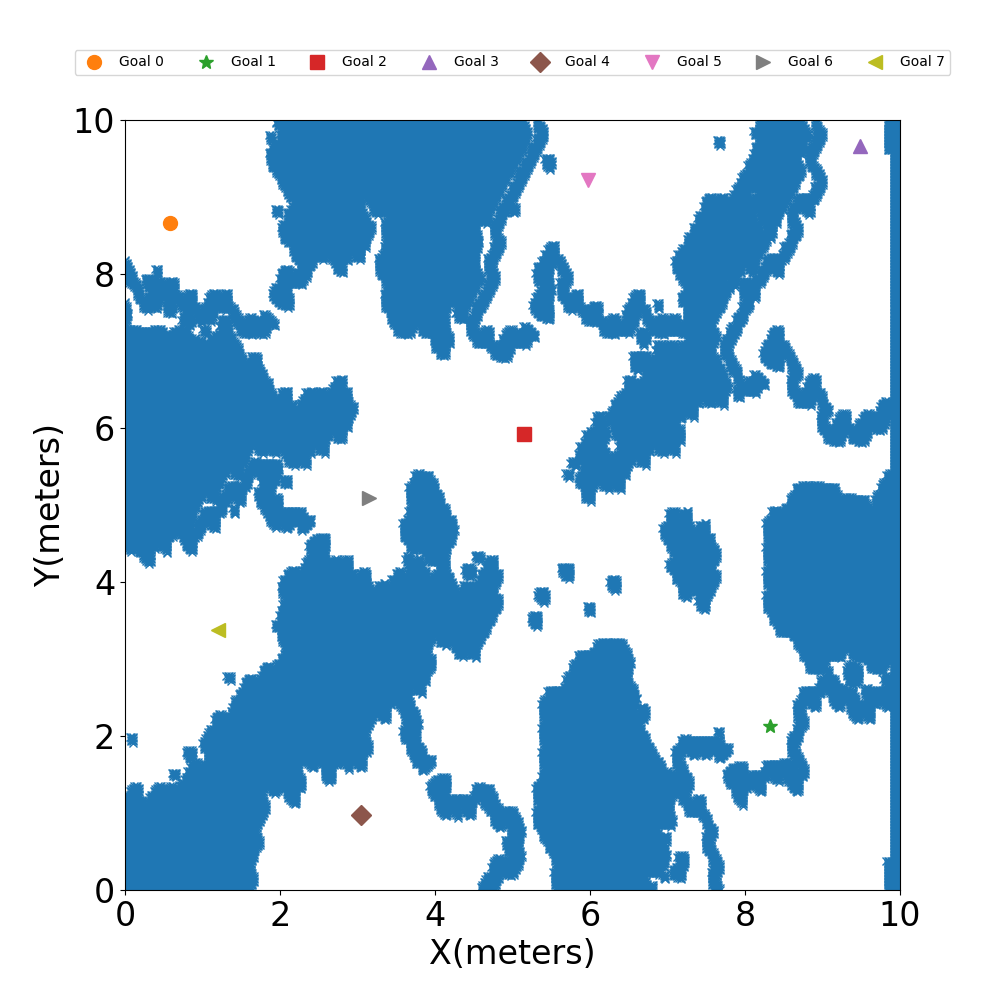}
        \caption{Aftershock scenario.}
        \label{fig:Aftershock}
    \end{subfigure}
    \hfill
    \begin{subfigure}{0.45\textwidth}
        \includegraphics[width=\linewidth]{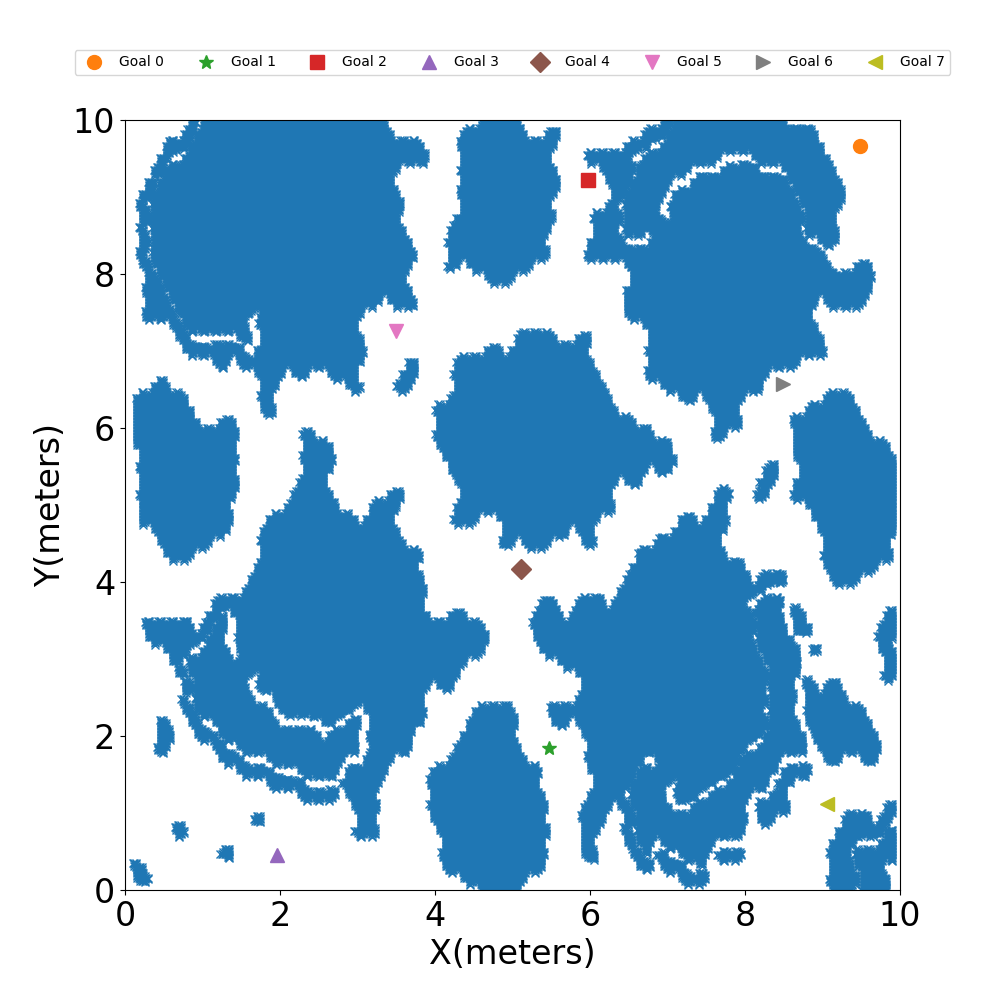}
        \caption{Archipelago scenario.}
    \end{subfigure}
    
    \vspace{\baselineskip}
    
    \begin{subfigure}{0.45\textwidth}
        \includegraphics[width=\linewidth]{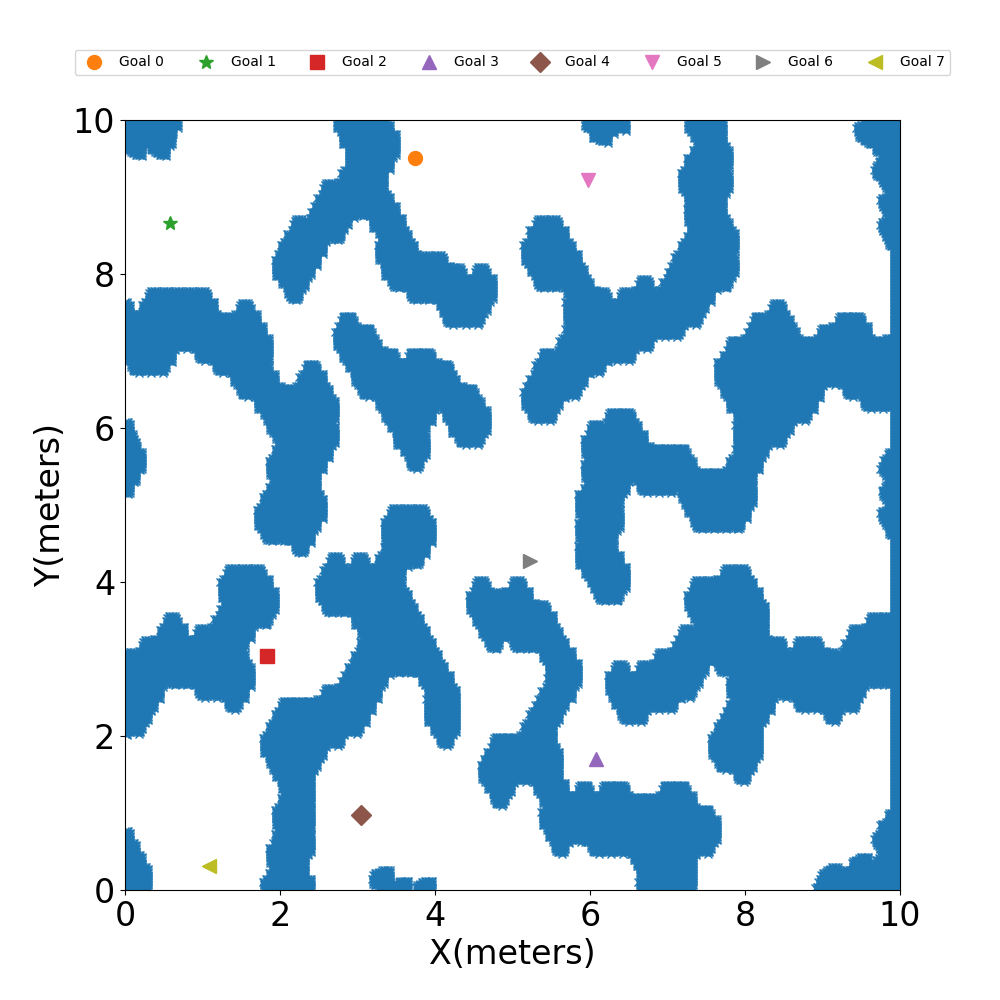}
        \caption{BigGameHunters scenario.}
    \end{subfigure}
    \hfill
    \begin{subfigure}{0.45\textwidth}
        \includegraphics[width=\linewidth]{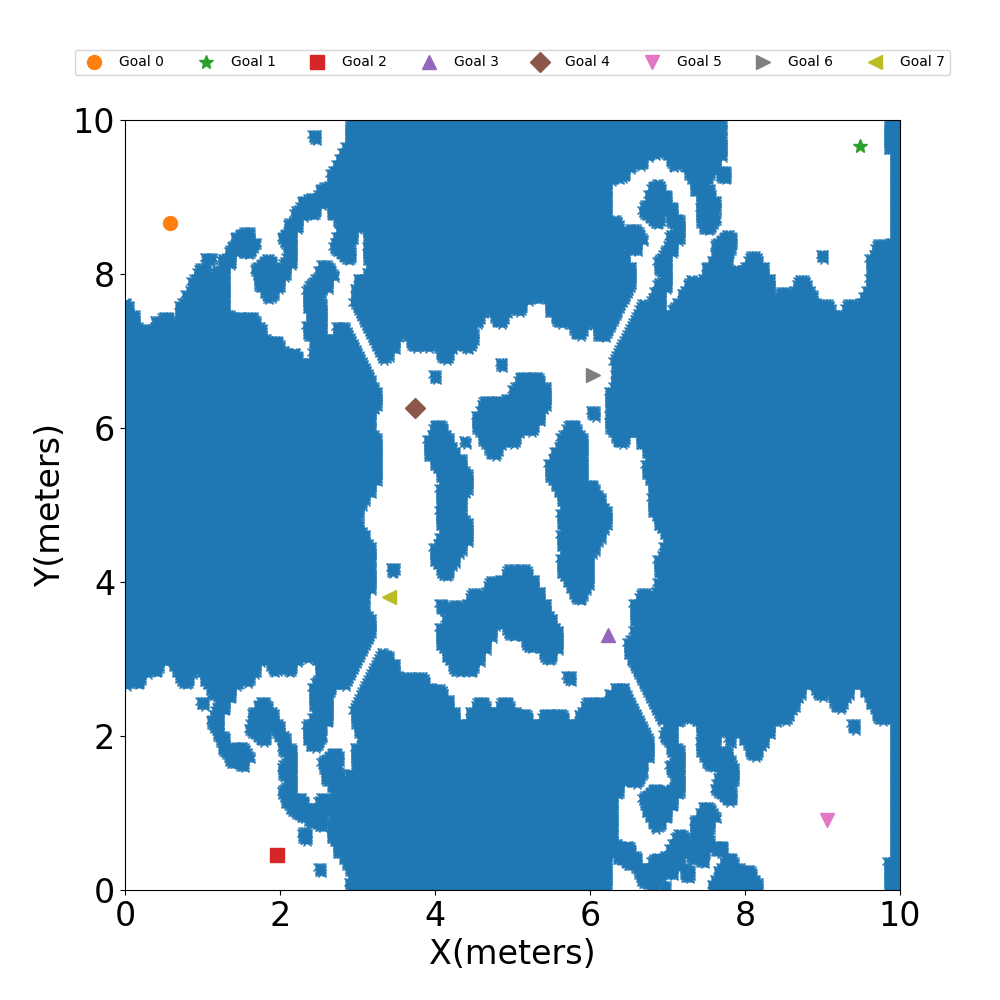}
        \caption{Brushfire scenario.}
    \end{subfigure}
    \vspace*{1em}
    \caption{Aftershock, Archipelago, BigGameHunters, and Brushfire scenarios used in the experiments.}
    \label{fig:after}
\end{figure*}

\begin{figure*}[h]
    \centering
    
    \begin{subfigure}{0.45\textwidth}
        \includegraphics[width=\linewidth]{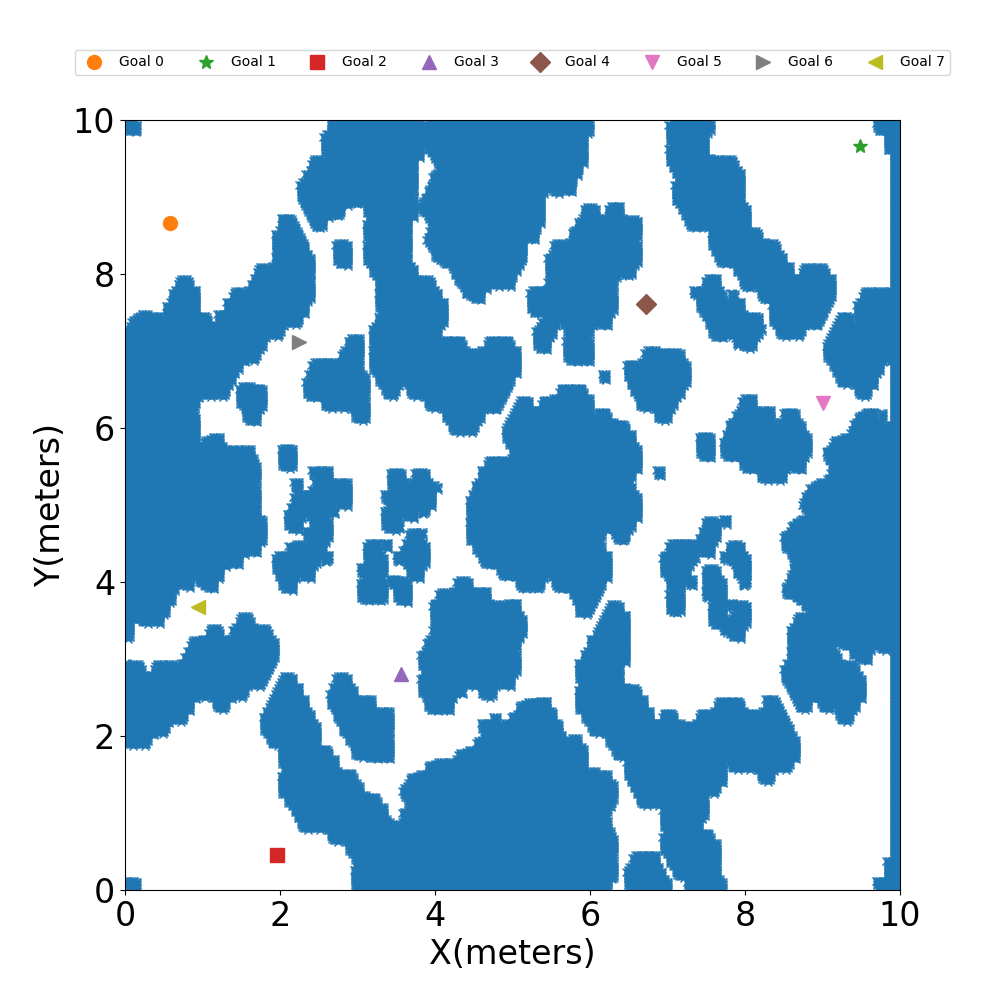}
        \caption{WarpGates scenario.}
    \end{subfigure}
    \hfill
    \begin{subfigure}{0.45\textwidth}
        \includegraphics[width=\linewidth]{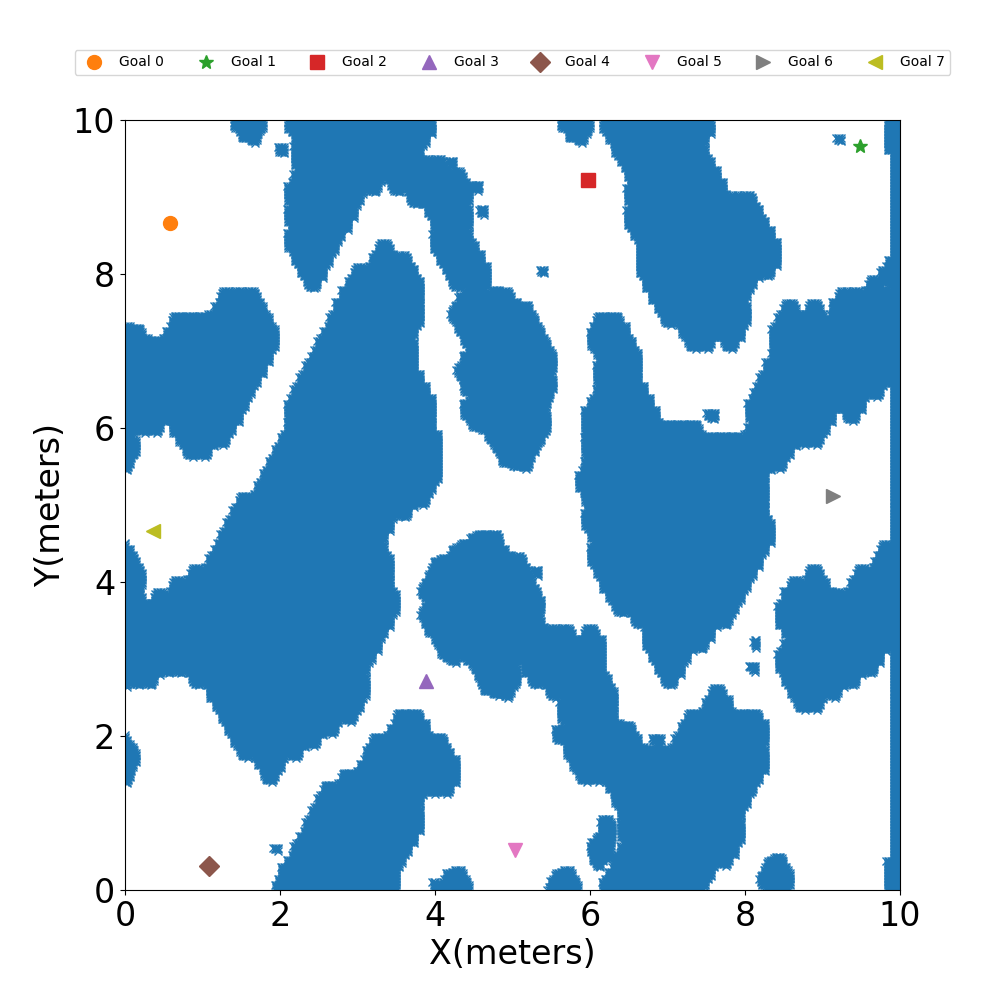}
        \caption{WatersEdge scenario.}
    \end{subfigure}
    
    \vspace{\baselineskip}
    
    \begin{subfigure}{0.45\textwidth}
        \includegraphics[width=\linewidth]{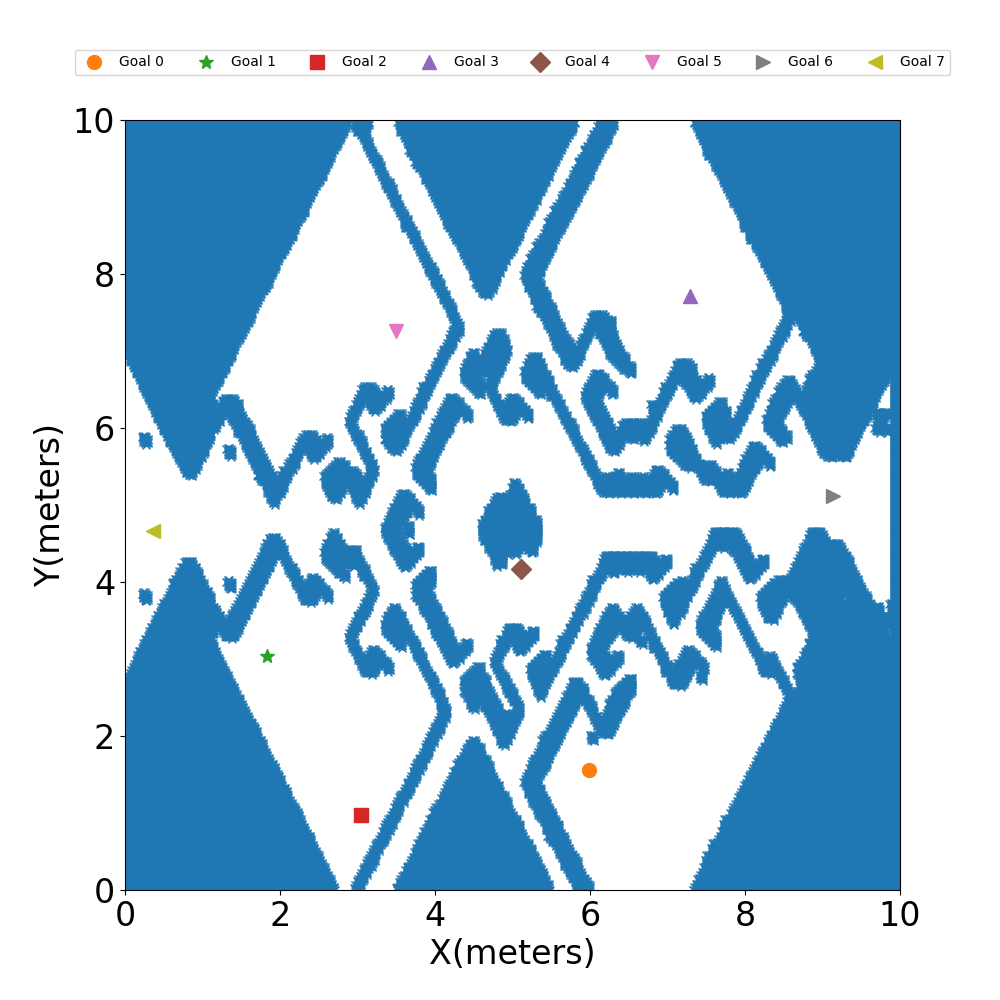}
        \caption{WaterWaypointJunctionsEdge scenario.}
    \end{subfigure}
    \hfill
    \begin{subfigure}{0.45\textwidth}
        \includegraphics[width=\linewidth]{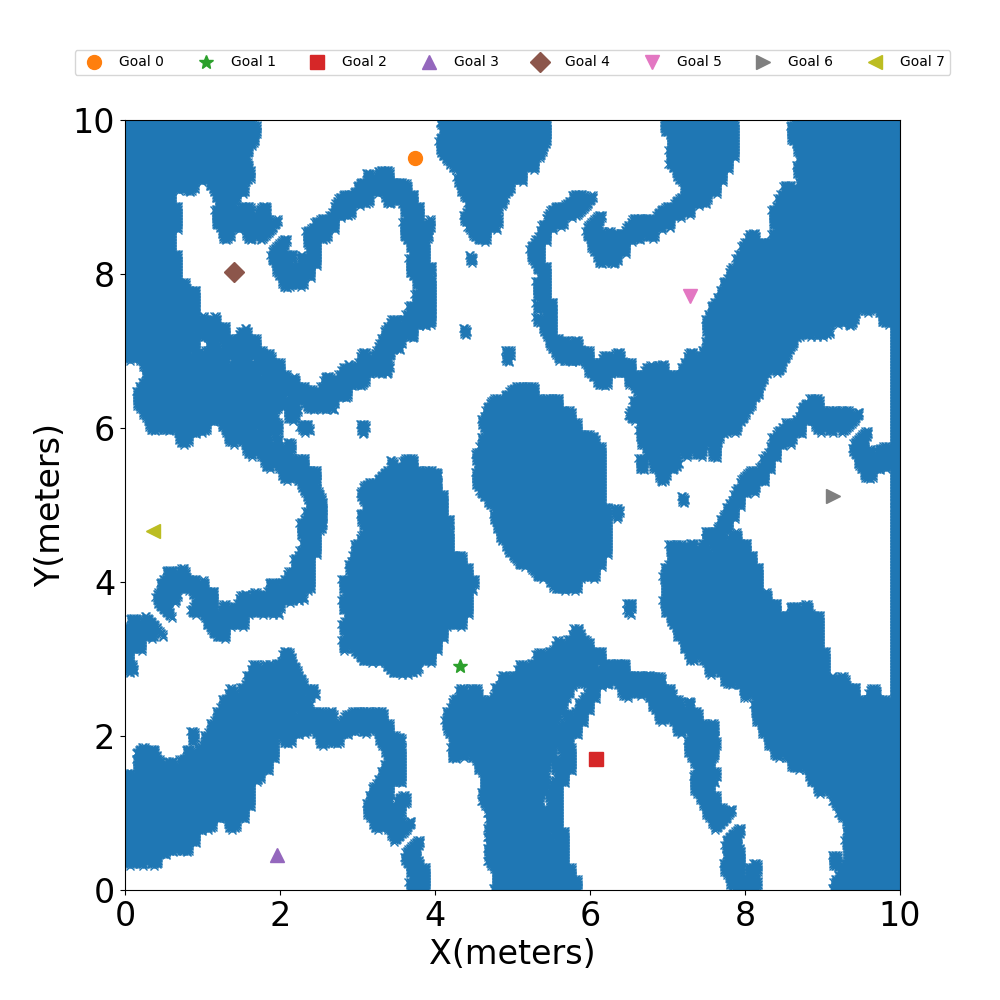}
        \caption{WinterConquest scenario.}
    \end{subfigure}
    \vspace*{1em}
    \caption{WarpGates, WatersEdge, WaypointJunction, and WinterConquest scenarios used in the experiments.}
    \label{fig:winter}
\end{figure*}

This appendix complements the information in Table 1 from the main text, which reports the overall results from the evaluation of the continuous domains over 28 benchmark scenarios from Moving-AI~\cite{sturtevant2012benchmarks}.
For each scenario, we sampled eight spatial points randomly throughout the scenario. 
The sampling points process follows two rules: at least $23cm$ from any wall and $2m$ away from every other point.
Figures~\ref{fig:after} to \ref{fig:winter} show each scenario with the sampled points used in the experiments. Tables~\ref{table:continuous_each}-~\ref{table:vector_k20} presents the results of each scenario individually, for Mirroring, R+P, and Vector inference for different $k$ values. Table \ref{table:vacancy} presents the percentage of vacancy for each scenario map, which means the amount of grid cells pixels are free of obstacles.   

Tables~\ref{table:continuous_each} and~\ref{table:vector_k20} show that our approach has a generally superior online time performance compared to the conventional methods. This increase in performance is due to our approach not using optimal trajectories in the inference process. We offered an alternative approximation solution aimed at speeding the whole process. The difference in using an approximated trajectory as inference can be seen in the results table and interpreted in the scenarios figures. The most challenging goal recognition problem in the scenarios studied happens when we have a specific combination of obstacles and hypothesis points where the trajectories computed in the inference process are very close or, in the worst case, equal. When this kind of problem happens, minor errors in the trajectory computation could be the difference between inferring a wrong hypothesis and a correct one.


\subsection{Multiple solution alternative}
 
\begin{figure}[h]
    \centering
    \includegraphics[width=\linewidth]{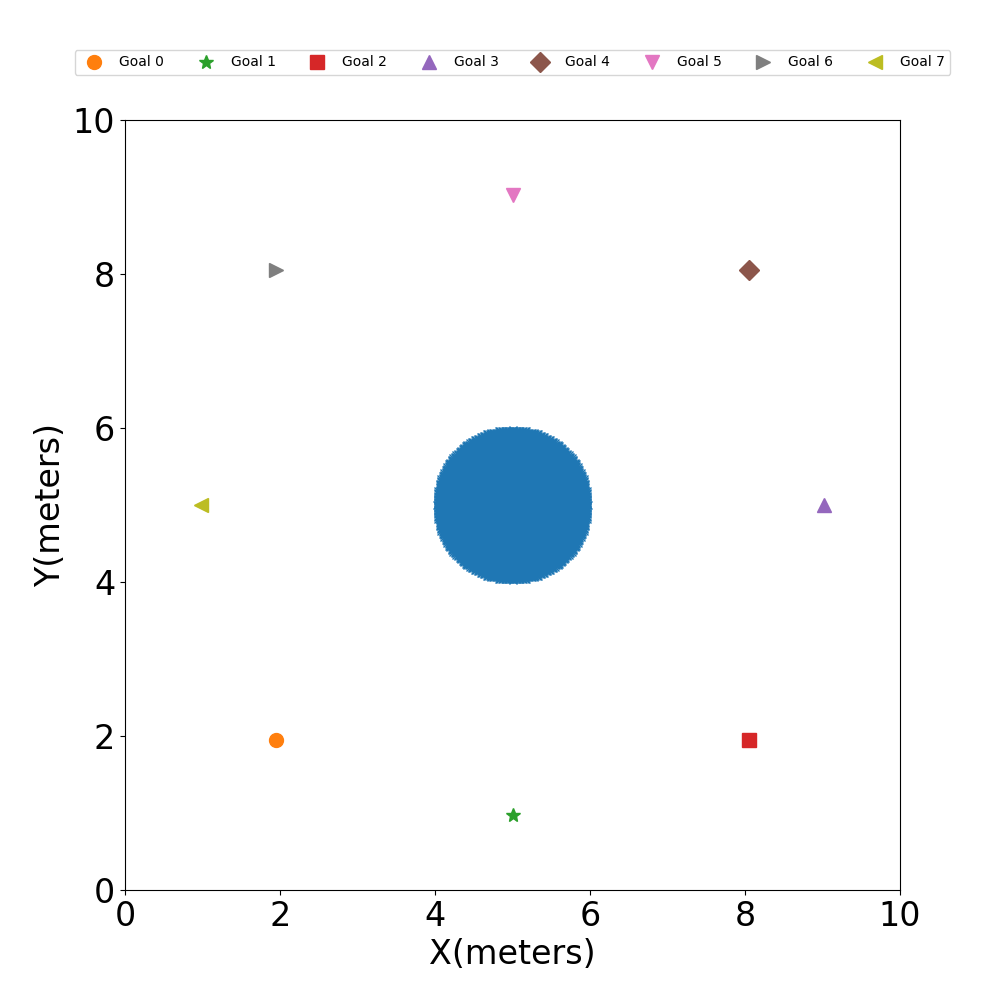}
    \caption{Circle scenario used in the experiments.}
    \label{fig:circle}
\end{figure}

To convey a scenario where computing only one optimal path may not be enough to recognize the intended goal correctly, we created a simple scenario to illustrate this problem and show the performance of the current methods of online goal recognition. Figure~\ref{fig:circle} shows the " Circle " working scenario.

As seen in Figure~\ref{fig:circle} each point has one symmetric combination with two minimum-cost solutions to the problem. For example, goal 1 could travel to the left side of the circle or to the right one to reach point 3; both have the exact cost. Tables~\ref{table:continuous_each} and~\ref{table:vector_k20} shows the results for the methods addressed in the paper and together with the results of our approach for the scenario Circle. The tables show that in this specific scenario, our method performs poorly compared to other conventional methods with $k=1$. These results indicate that our method has a disadvantage in identifying goals with more than one trajectory as a solution with a minimum cost. However, for results with $k$ greater than 5 we got an increase of about $20\%$ in the PPV performance at the cost of an increase of nearly five times on the online processing and an increase of 1.42 times on the offline processing. 

\section{Discrete Domain Results Breakdown}
This appendix complements the information in Table 2 of the paper, which reports the overall results from evaluating the discrete domains over the openly available goal and plan recognition dataset~\cite{pereira2017landmark}. 
Table 2 of the paper compares the discrete methods of goal recognition, where the results are present considering the average among all domains used in the experiment. 
Here, in Tables~\ref{table:discrete_mirroring_landmarks}-~\ref{table:discrete_vector_k15}, we detail the results for each domain individually. 
This table includes the following information: $\vert \mathcal{O} \vert$ is the observation set size, which represents the whole plan that achieves the hidden goal, i.e., having observed 100\% of the actions; $\vert \mathcal{G} \vert$ is the goal hypothesis set size; positive predictive value (PPV) in percentage; accuracy (ACC) for each experiment in percentage; spread (SPR) is the size of the hypothesis solution set chosen by recognition method; We separate the algorithm's online and offline parts (units in seconds) to highlight the key advantage of our method. Planner Call (PC) is the number of planner calls during recognition.

The landmarks-based method is the fastest approach in the offline stage because it does not use a planner in its inference. The disadvantage is an inferior PPV (and not handling continuous domains). 
The results indicate that our approach has the best performance in PPV and ACC metrics according to the number of solutions $k$. Furthermore, it has the fastest online inference process.
Indeed, increasing $k$ might not necessarily improve performance when the number of optimal plans for a problem is fewer than $k$. Table \ref{table:discrete_vector_k50} show experiments results with $k=50$ and filtering plans for optimality, the results show a general PPV improvement of $4.5\%$.

\begin{table*}[]
\centering
\setlength{\tabcolsep}{5pt}
\begin{tabular}{ccccccc|cccccc}
\cline{2-13}
 & \multicolumn{6}{c|}{Mirroring} & \multicolumn{6}{c}{R+P} \\ \hline
\multicolumn{1}{c|}{Scenario} & PPV(\%) & ACC(\%) & SPR & PC & \begin{tabular}[c]{@{}c@{}}Online\\ Time(s)\end{tabular} & \begin{tabular}[c]{@{}c@{}}Offline\\ Time(s)\end{tabular} & PPV(\%) & ACC(\%) & SPR & PC & \begin{tabular}[c]{@{}c@{}}Online\\ Time(s)\end{tabular} & \begin{tabular}[c]{@{}c@{}}Offline\\ Time(s)\end{tabular} \\ \hline
\multicolumn{1}{c|}{Aftershock} & 46.2 & 86.6 & 1.0 & 49.0 & 7.68E+3 & 2.26E+3 & 49.5 & 87.4 & 1.0 & 30.7 & 3.87E+3 & 2.26E+3 \\
\multicolumn{1}{c|}{Archipelago} & 40.3 & 85.0 & 1.0 & 49.0 & 6.14E+3 & 1.44E+3 & 48.0 & 87.0 & 1.0 & 29.4 & 2.88E+3 & 1.44E+3 \\
\multicolumn{1}{c|}{BigGameHunters} & 41.3 & 85.2 & 1.0 & 49.0 & 4.82E+3 & 1.68E+3 & 40.4 & 84.9 & 1.0 & 30.9 & 2.62E+3 & 1.68E+3 \\
\multicolumn{1}{c|}{Brushfire} & 38.3 & 84.6 & 1.0 & 49.0 & 9.51E+3 & 1.35E+3 & 40.1 & 85.0 & 1.0 & 31.1 & 5.05E+3 & 1.35E+3 \\
\multicolumn{1}{c|}{Caldera} & 51.3 & 87.8 & 1.0 & 49.0 & 3.82E+3 & 7.27E+2 & 58.7 & 89.6 & 1.0 & 27.3 & 1.56E+3 & 7.27E+2 \\
\multicolumn{1}{c|}{Circle} & 77.3 & 94.4 & 1.0 & 49.0 & 3.94E+3 & 8.41E+2 & 80.1 & 94.5 & 1.2 & 17.6 & 1.24E+3 & 8.47E+2 \\
\multicolumn{1}{c|}{CrescentMoon} & 43.5 & 85.8 & 1.0 & 49.0 & 6.60E+3 & 1.51E+3 & 44.0 & 85.9 & 1.1 & 28.3 & 2.83E+3 & 1.51E+3 \\
\multicolumn{1}{c|}{Desolation} & 37.0 & 84.1 & 1.0 & 49.0 & 6.57E+3 & 1.81E+3 & 41.4 & 85.3 & 1.0 & 31.4 & 3.02E+3 & 1.81E+3 \\
\multicolumn{1}{c|}{EbonLakes} & 34.4 & 83.6 & 1.0 & 49.0 & 1.18E+4 & 1.86E+3 & 36.7 & 84.0 & 1.0 & 29.2 & 5.56E+3 & 1.86E+3 \\
\multicolumn{1}{c|}{Entanglement} & 44.5 & 86.1 & 1.0 & 49.0 & 4.85E+3 & 7.44E+2 & 44.7 & 86.2 & 1.0 & 28.7 & 2.15E+3 & 7.44E+2 \\
\multicolumn{1}{c|}{Eruption} & 41.9 & 85.5 & 1.0 & 49.0 & 6.62E+3 & 2.26E+3 & 44.3 & 86.1 & 1.0 & 30.9 & 2.98E+3 & 2.26E+3 \\
\multicolumn{1}{c|}{HotZone} & 39.8 & 84.8 & 1.0 & 49.0 & 7.03E+3 & 1.74E+3 & 43.2 & 85.7 & 1.0 & 30.2 & 3.22E+3 & 1.74E+3 \\
\multicolumn{1}{c|}{Isolation} & 45.6 & 86.4 & 1.0 & 49.0 & 9.23E+3 & 1.57E+3 & 50.0 & 87.4 & 1.0 & 28.8 & 4.11E+3 & 1.57E+3 \\
\multicolumn{1}{c|}{Legacy} & 26.2 & 81.5 & 1.0 & 49.0 & 7.79E+3 & 1.56E+3 & 39.3 & 84.8 & 1.0 & 28.7 & 3.36E+3 & 1.07E+3 \\
\multicolumn{1}{c|}{OrbitalGully} & 38.3 & 84.6 & 1.0 & 49.0 & 1.60E+4 & 2.24E+3 & 45.6 & 86.3 & 1.0 & 28.0 & 6.28E+3 & 2.24E+3 \\
\multicolumn{1}{c|}{Predators} & 40.9 & 85.2 & 1.0 & 49.0 & 5.02E+3 & 9.59E+2 & 40.1 & 84.9 & 1.0 & 28.0 & 2.02E+3 & 9.59E+2 \\
\multicolumn{1}{c|}{Ramparts} & 27.9 & 81.9 & 1.0 & 49.0 & 2.51E+4 & 6.62E+3 & 34.4 & 83.6 & 1.0 & 33.6 & 1.23E+4 & 5.62E+3 \\
\multicolumn{1}{c|}{RedCanyons} & 42.2 & 85.6 & 1.0 & 49.0 & 7.06E+3 & 1.81E+3 & 42.9 & 85.7 & 1.0 & 31.0 & 3.51E+3 & 1.81E+3 \\
\multicolumn{1}{c|}{Rosewood} & 43.8 & 85.9 & 1.0 & 49.0 & 9.17E+3 & 2.31E+3 & 45.5 & 86.2 & 1.0 & 29.7 & 4.37E+3 & 2.31E+3 \\
\multicolumn{1}{c|}{Sanctuary} & 34.6 & 83.7 & 1.0 & 49.0 & 1.65E+4 & 3.49E+3 & 40.5 & 85.1 & 1.0 & 29.5 & 7.66E+3 & 3.49E+3 \\
\multicolumn{1}{c|}{ShroudPlatform} & 44.0 & 86.0 & 1.0 & 49.0 & 2.22E+4 & 6.50E+3 & 41.6 & 85.0 & 1.1 & 31.5 & 1.18E+4 & 6.08E+3 \\
\multicolumn{1}{c|}{SpaceAtoll} & 29.0 & 82.2 & 1.0 & 49.0 & 1.86E+4 & 4.22E+3 & 33.0 & 83.1 & 1.0 & 28.6 & 8.16E+3 & 4.22E+3 \\
\multicolumn{1}{c|}{SpaceDebris} & 38.4 & 84.6 & 1.0 & 49.0 & 9.48E+3 & 1.79E+3 & 38.3 & 84.6 & 1.0 & 30.6 & 4.96E+3 & 1.79E+3 \\
\multicolumn{1}{c|}{SteppingStones} & 54.0 & 88.5 & 1.0 & 49.0 & 2.80E+3 & 6.89E+2 & 54.4 & 88.6 & 1.0 & 28.7 & 1.35E+3 & 6.89E+2 \\
\multicolumn{1}{c|}{Triskelion} & 30.6 & 82.6 & 1.0 & 49.0 & 1.90E+4 & 5.04E+3 & 34.5 & 83.6 & 1.0 & 32.6 & 1.12E+4 & 5.04E+3 \\
\multicolumn{1}{c|}{WarpGates} & 38.4 & 84.6 & 1.0 & 49.0 & 9.89E+3 & 1.90E+3 & 38.3 & 84.3 & 1.0 & 29.9 & 4.37E+3 & 1.90E+3 \\
\multicolumn{1}{c|}{WatersEdge} & 40.4 & 85.1 & 1.0 & 49.0 & 1.20E+4 & 4.37E+3 & 41.2 & 84.9 & 1.0 & 29.8 & 6.42E+3 & 4.37E+3 \\
\multicolumn{1}{c|}{WaypointJunction} & 39.0 & 84.7 & 1.0 & 49.0 & 8.53E+3 & 1.71E+3 & 44.2 & 86.0 & 1.0 & 31.9 & 4.04E+3 & 1.71E+3 \\
\multicolumn{1}{c|}{WinterConquest} & 43.2 & 85.8 & 1.0 & 49.0 & 1.34E+4 & 3.45E+3 & 49.9 & 87.0 & 1.0 & 29.7 & 6.76E+3 & 3.39E+3 \\ \hline
\multicolumn{1}{c|}{Average} & 41.1 & 85.2 & 1.0 & 49.0 & 1.0E+4 & 2.3E+3 & 44.3 & 85.9 & 1.0 & 29.5 & 4.8E+3 & 2.3E+3 \\ \hline
\end{tabular}
\caption{Comparison among continuous online goal recognition methods for each scenario. Mirroring and R+P methods results.}
\label{table:continuous_each}
\end{table*}

\begin{table*}[]
\centering
\setlength{\tabcolsep}{5pt}
\begin{tabular}{ccccccc|cccccc}
\cline{2-13}
 & \multicolumn{6}{c|}{Vector k=1} & \multicolumn{6}{c}{Vector k=5} \\ \hline
\multicolumn{1}{c|}{Scenario} & PPV(\%) & ACC(\%) & SPR & PC & \multicolumn{1}{c}{\begin{tabular}[c]{@{}c@{}}Online\\ Time(s)\end{tabular}} & \multicolumn{1}{c|}{\begin{tabular}[c]{@{}c@{}}Offline\\ Time(s)\end{tabular}} & PPV(\%) & ACC(\%) & SPR & PC & \multicolumn{1}{c}{\begin{tabular}[c]{@{}c@{}}Online\\ Time(s)\end{tabular}} & \multicolumn{1}{c}{\begin{tabular}[c]{@{}c@{}}Offline\\ Time(s)\end{tabular}} \\ \hline
\multicolumn{1}{c|}{Aftershock} & 43.7 & 85.9 & 1.0 & 7.0 & 9.2E-02 & 7.4E+01 & 51.2 & 87.8 & 1.0 & 7.0 & 3.8E-02 & 2.1E+02 \\
\multicolumn{1}{c|}{Archipelago} & 41.4 & 85.3 & 1.0 & 7.0 & 9.2E-02 & 6.1E+01 & 44.3 & 86.1 & 1.0 & 7.0 & 3.8E-02 & 1.9E+02 \\
\multicolumn{1}{c|}{BigGameHunters} & 45.9 & 86.4 & 1.0 & 7.0 & 9.2E-02 & 5.8E+01 & 43.5 & 85.9 & 1.0 & 7.0 & 3.9E-02 & 1.9E+02 \\
\multicolumn{1}{c|}{Brushfire} & 40.8 & 85.2 & 1.0 & 7.0 & 9.3E-02 & 1.4E+02 & 52.1 & 88.0 & 1.0 & 7.0 & 4.3E-02 & 3.1E+02 \\
\multicolumn{1}{c|}{Caldera} & 47.4 & 86.9 & 1.0 & 7.0 & 9.1E-02 & 5.9E+01 & 56.8 & 89.2 & 1.0 & 7.0 & 3.9E-02 & 1.8E+02 \\
\multicolumn{1}{c|}{Circle} & 66.3 & 91.6 & 1.0 & 7.0 & 4.4E-02 & 1.5E+02 & 74.7 & 93.6 & 1.0 & 7.0 & 4.3E-02 & 1.6E+02 \\
\multicolumn{1}{c|}{CrescentMoon} & 40.8 & 85.2 & 1.0 & 7.0 & 9.0E-02 & 5.9E+01 & 41.7 & 85.4 & 1.0 & 7.0 & 3.9E-02 & 1.2E+02 \\
\multicolumn{1}{c|}{Desolation} & 39.9 & 85.0 & 1.0 & 7.0 & 9.1E-02 & 9.6E+01 & 51.2 & 87.8 & 1.0 & 7.0 & 4.0E-02 & 1.7E+02 \\
\multicolumn{1}{c|}{EbonLakes} & 43.2 & 85.8 & 1.0 & 7.0 & 9.4E-02 & 6.2E+01 & 47.9 & 87.0 & 1.0 & 7.0 & 4.0E-02 & 1.4E+02 \\
\multicolumn{1}{c|}{Entanglement} & 44.5 & 86.1 & 1.0 & 7.0 & 9.4E-02 & 1.4E+02 & 58.3 & 89.6 & 1.0 & 7.0 & 3.9E-02 & 2.0E+02 \\
\multicolumn{1}{c|}{Eruption} & 36.9 & 84.2 & 1.0 & 7.0 & 6.3E-02 & 1.2E+02 & 43.7 & 85.9 & 1.0 & 7.0 & 3.8E-02 & 1.3E+02 \\
\multicolumn{1}{c|}{HotZone} & 49.4 & 87.4 & 1.0 & 7.0 & 9.0E-02 & 5.9E+01 & 48.8 & 87.2 & 1.0 & 7.0 & 3.9E-02 & 1.2E+02 \\
\multicolumn{1}{c|}{Isolation} & 53.3 & 88.3 & 1.0 & 7.0 & 9.7E-02 & 2.0E+02 & 61.0 & 90.3 & 1.0 & 7.0 & 4.0E-02 & 1.8E+02 \\
\multicolumn{1}{c|}{Legacy} & 36.1 & 84.0 & 1.0 & 7.0 & 8.9E-02 & 9.8E+01 & 35.1 & 83.8 & 1.0 & 7.0 & 3.8E-02 & 1.6E+02 \\
\multicolumn{1}{c|}{OrbitalGully} & 43.2 & 85.7 & 1.0 & 7.0 & 9.6E-02 & 6.3E+01 & 40.5 & 85.1 & 1.0 & 7.0 & 3.9E-02 & 1.3E+02 \\
\multicolumn{1}{c|}{Predators} & 42.2 & 85.6 & 1.0 & 7.0 & 9.5E-02 & 1.4E+02 & 56.3 & 89.1 & 1.0 & 7.0 & 3.8E-02 & 2.1E+02 \\
\multicolumn{1}{c|}{Ramparts} & 31.3 & 82.8 & 1.0 & 7.0 & 9.0E-02 & 6.9E+01 & 34.2 & 83.6 & 1.0 & 7.0 & 3.8E-02 & 1.2E+02 \\
\multicolumn{1}{c|}{RedCanyons} & 41.1 & 85.2 & 1.0 & 7.0 & 9.4E-02 & 1.4E+02 & 51.2 & 87.8 & 1.0 & 7.0 & 3.9E-02 & 2.0E+02 \\
\multicolumn{1}{c|}{Rosewood} & 41.6 & 85.4 & 1.0 & 7.0 & 9.3E-02 & 1.1E+02 & 44.0 & 86.0 & 1.0 & 7.0 & 3.8E-02 & 1.9E+02 \\
\multicolumn{1}{c|}{Sanctuary} & 32.4 & 83.1 & 1.0 & 7.0 & 9.2E-02 & 1.6E+02 & 36.3 & 84.1 & 1.0 & 7.0 & 3.9E-02 & 2.3E+02 \\
\multicolumn{1}{c|}{ShroudPlatform} & 40.5 & 85.1 & 1.0 & 7.0 & 9.3E-02 & 1.6E+02 & 44.3 & 86.1 & 1.0 & 7.0 & 3.9E-02 & 2.3E+02 \\
\multicolumn{1}{c|}{SpaceAtoll} & 34.4 & 83.6 & 1.0 & 7.0 & 9.2E-02 & 8.0E+01 & 39.0 & 84.7 & 1.0 & 7.0 & 3.7E-02 & 1.6E+02 \\
\multicolumn{1}{c|}{SpaceDebris} & 40.2 & 85.0 & 1.0 & 7.0 & 9.1E-02 & 5.7E+01 & 38.4 & 84.6 & 1.0 & 7.0 & 3.8E-02 & 1.1E+02 \\
\multicolumn{1}{c|}{SteppingStones} & 55.1 & 88.8 & 1.0 & 7.0 & 9.1E-02 & 1.2E+02 & 63.4 & 90.8 & 1.0 & 7.0 & 4.0E-02 & 1.9E+02 \\
\multicolumn{1}{c|}{Triskelion} & 31.2 & 82.8 & 1.0 & 7.0 & 9.4E-02 & 8.4E+01 & 48.2 & 87.1 & 1.0 & 7.0 & 3.9E-02 & 1.5E+02 \\
\multicolumn{1}{c|}{WarpGates} & 34.8 & 83.7 & 1.0 & 7.0 & 8.8E-02 & 8.8E+01 & 42.3 & 85.6 & 1.0 & 7.0 & 3.8E-02 & 1.6E+02 \\
\multicolumn{1}{c|}{WatersEdge} & 31.2 & 82.8 & 1.0 & 7.0 & 9.1E-02 & 1.5E+02 & 39.0 & 84.7 & 1.0 & 7.0 & 3.8E-02 & 2.2E+02 \\
\multicolumn{1}{c|}{WaypointJunction} & 32.7 & 83.2 & 1.0 & 7.0 & 9.2E-02 & 1.3E+02 & 50.0 & 87.5 & 1.0 & 7.0 & 4.0E-02 & 1.7E+02 \\
\multicolumn{1}{c|}{WinterConquest} & 47.3 & 86.8 & 1.0 & 7.0 & 9.1E-02 & 1.0E+02 & 46.1 & 86.5 & 1.0 & 7.0 & 3.9E-02 & 1.8E+02 \\ \hline
\multicolumn{1}{c|}{Average} & 41.6 & 85.4 & 1.0 & 7.0 & 8.9E-02 & 1.0E+02 & 47.7 & 86.9 & 1.0 & 7.0 & \multicolumn{1}{c}{3.8E-02} & \multicolumn{1}{c}{1.7E+02} \\ \hline
\end{tabular}
\caption{Comparison among continuous online goal recognition methods for each scenario. Vector methods results for $k=1$ and $k=5$.}
\end{table*}

\begin{table*}[]
\centering
\setlength{\tabcolsep}{5pt}
\begin{tabular}{ccccccc|cccccc}
\cline{2-13}
 & \multicolumn{6}{c|}{Vector k=10} & \multicolumn{6}{c}{Vector k=15} \\ \hline
\multicolumn{1}{c|}{Scenario} & PPV(\%) & ACC(\%) & SPR & PC & \begin{tabular}[c]{@{}c@{}}Online\\ Time(s)\end{tabular} & \begin{tabular}[c]{@{}c@{}}Offline\\ Time(s)\end{tabular} & PPV(\%) & ACC(\%) & SPR & PC & \begin{tabular}[c]{@{}c@{}}Online\\ Time(s)\end{tabular} & \begin{tabular}[c]{@{}c@{}}Offline\\ Time(s)\end{tabular} \\ \hline
\multicolumn{1}{c|}{Aftershock} & 51.5 & 87.9 & 1.0 & 7.0 & 4.2E-02 & 2.0E+02 & 49.1 & 87.3 & 1.0 & 7.0 & 3.7E-02 & 2.1E+02 \\
\multicolumn{1}{c|}{Archipelago} & 41.7 & 85.4 & 1.0 & 7.0 & 4.0E-02 & 1.7E+02 & 47.3 & 86.8 & 1.0 & 7.0 & 3.7E-02 & 1.9E+02 \\
\multicolumn{1}{c|}{BigGameHunters} & 42.0 & 85.5 & 1.0 & 7.0 & 4.2E-02 & 1.9E+02 & 47 & 86.8 & 1.0 & 7.0 & 3.7E-02 & 1.9E+02 \\
\multicolumn{1}{c|}{Brushfire} & 49.7 & 87.4 & 1.0 & 7.0 & 4.1E-02 & 3.1E+02 & 51.2 & 87.8 & 1.0 & 7.0 & 3.7E-02 & 3.2E+02 \\
\multicolumn{1}{c|}{Caldera} & 53.9 & 88.5 & 1.0 & 7.0 & 4.1E-02 & 1.8E+02 & 57.1 & 89.3 & 1.0 & 7.0 & 3.6E-02 & 1.9E+02 \\
\multicolumn{1}{c|}{Circle} & 82.7 & 95.6 & 1.0 & 7.0 & 3.8E-02 & 1.7E+02 & 84.8 & 96.2 & 1.0 & 7.0 & 3.5E-02 & 1.7E+02 \\
\multicolumn{1}{c|}{CrescentMoon} & 41.7 & 85.4 & 1.0 & 7.0 & 4.0E-02 & 1.8E+02 & 44.3 & 86.1 & 1.0 & 7.0 & 3.7E-02 & 1.9E+02 \\
\multicolumn{1}{c|}{Desolation} & 43.5 & 85.9 & 1.0 & 7.0 & 4.2E-02 & 2.1E+02 & 44.3 & 86.1 & 1.0 & 7.0 & 3.9E-02 & 2.2E+02 \\
\multicolumn{1}{c|}{EbonLakes} & 46.7 & 86.7 & 1.0 & 7.0 & 3.9E-02 & 1.9E+02 & 60.1 & 90 & 1.0 & 7.0 & 3.7E-02 & 2.0E+02 \\
\multicolumn{1}{c|}{Entanglement} & 55.4 & 88.9 & 1.0 & 7.0 & 4.2E-02 & 2.6E+02 & 44.6 & 86.2 & 1.0 & 7.0 & 3.7E-02 & 2.6E+02 \\
\multicolumn{1}{c|}{Eruption} & 44.0 & 86.0 & 1.0 & 7.0 & 4.3E-02 & 1.8E+02 & 53 & 88.2 & 1.0 & 7.0 & 3.6E-02 & 2.0E+02 \\
\multicolumn{1}{c|}{HotZone} & 48.5 & 87.1 & 1.0 & 7.0 & 4.2E-02 & 1.8E+02 & 51.5 & 87.9 & 1.0 & 7.0 & 3.8E-02 & 1.9E+02 \\
\multicolumn{1}{c|}{Isolation} & 55.7 & 88.9 & 1.0 & 7.0 & 4.2E-02 & 2.9E+02 & 41.4 & 85.3 & 1.0 & 7.0 & 3.7E-02 & 3.0E+02 \\
\multicolumn{1}{c|}{Legacy} & 38.4 & 84.6 & 1.0 & 7.0 & 4.0E-02 & 2.0E+02 & 55.1 & 88.8 & 1.0 & 7.0 & 3.6E-02 & 2.2E+02 \\
\multicolumn{1}{c|}{OrbitalGully} & 55.7 & 88.9 & 1.0 & 7.0 & 4.1E-02 & 1.8E+02 & 51.2 & 87.8 & 1.0 & 7.0 & 3.6E-02 & 1.9E+02 \\
\multicolumn{1}{c|}{Predators} & 53.9 & 88.5 & 1.0 & 7.0 & 4.2E-02 & 2.5E+02 & 41.7 & 85.4 & 1.0 & 7.0 & 3.7E-02 & 2.7E+02 \\
\multicolumn{1}{c|}{Ramparts} & 44.0 & 86.0 & 1.0 & 7.0 & 4.2E-02 & 1.7E+02 & 54 & 88.5 & 1.0 & 7.0 & 3.7E-02 & 1.9E+02 \\
\multicolumn{1}{c|}{RedCanyons} & 49.5 & 87.4 & 1.0 & 7.0 & 4.3E-02 & 2.6E+02 & 50.6 & 87.6 & 1.0 & 7.0 & 3.7E-02 & 2.7E+02 \\
\multicolumn{1}{c|}{Rosewood} & 46.4 & 86.6 & 1.0 & 7.0 & 4.0E-02 & 2.0E+02 & 41.4 & 85.3 & 1.0 & 7.0 & 3.7E-02 & 2.3E+02 \\
\multicolumn{1}{c|}{Sanctuary} & 40.5 & 85.1 & 1.0 & 7.0 & 4.2E-02 & 2.5E+02 & 48.5 & 87.1 & 1.0 & 7.0 & 3.7E-02 & 2.8E+02 \\
\multicolumn{1}{c|}{ShroudPlatform} & 47.0 & 86.8 & 1.0 & 7.0 & 4.2E-02 & 2.6E+02 & 38.7 & 84.7 & 1.0 & 7.0 & 3.7E-02 & 2.8E+02 \\
\multicolumn{1}{c|}{SpaceAtoll} & 43.2 & 85.8 & 1.0 & 7.0 & 4.0E-02 & 1.8E+02 & 43.5 & 85.9 & 1.0 & 7.0 & 3.9E-02 & 2.0E+02 \\
\multicolumn{1}{c|}{SpaceDebris} & 47.3 & 86.8 & 1.0 & 7.0 & 4.5E-02 & 1.8E+02 & 60.1 & 90 & 1.0 & 7.0 & 3.6E-02 & 1.8E+02 \\
\multicolumn{1}{c|}{SteppingStones} & 66.4 & 91.6 & 1.0 & 7.0 & 4.0E-02 & 2.3E+02 & 51.5 & 87.9 & 1.0 & 7.0 & 3.7E-02 & 2.4E+02 \\
\multicolumn{1}{c|}{Triskelion} & 50.9 & 87.7 & 1.0 & 7.0 & 4.2E-02 & 1.8E+02 & 46.7 & 86.7 & 1.0 & 7.0 & 3.7E-02 & 2.0E+02 \\
\multicolumn{1}{c|}{WarpGates} & 42.9 & 85.7 & 1.0 & 7.0 & 4.2E-02 & 1.8E+02 & 45.5 & 86.4 & 1.0 & 7.0 & 3.8E-02 & 2.0E+02 \\
\multicolumn{1}{c|}{WatersEdge} & 33.0 & 83.3 & 1.0 & 7.0 & 4.2E-02 & 2.6E+02 & 46.1 & 86.5 & 1.0 & 7.0 & 3.7E-02 & 2.8E+02 \\
\multicolumn{1}{c|}{WaypointJunction} & 41.4 & 85.3 & 1.0 & 7.0 & 4.3E-02 & 2.5E+02 & 47 & 86.8 & 1.0 & 7.0 & 3.7E-02 & 2.6E+02 \\
\multicolumn{1}{c|}{WinterConquest} & 51.8 & 87.9 & 1.0 & 7.0 & 4.1E-02 & 2.1E+02 & 48.5 & 87.1 & 1.0 & 7.0 & 3.7E-02 & 2.3E+02 \\ \hline
\multicolumn{1}{c|}{Average} & 48.5 & 87.1 & 1.0 & 7.0 & 4.1E-02 & 2.1E+02 & 49.7 & 87.4 & 1.0 & 7.0 & 3.7E-02 & 2.2E+02 \\ \hline
\end{tabular}
\caption{Comparison among continuous online goal recognition methods for each scenario. Vector methods results for $k=10$ and $k=15$.}
\end{table*}

\begin{table*}[]
\centering
\setlength{\tabcolsep}{5pt}
\begin{tabular}{ccccccc}
\cline{2-7}
 & \multicolumn{6}{c}{Vector k=20} \\ \hline
\multicolumn{1}{c|}{Scenario} & PPV(\%) & ACC(\%) & SPR & PC & \begin{tabular}[c]{@{}c@{}}Online\\ Time(s)\end{tabular} & \begin{tabular}[c]{@{}c@{}}Offline\\ Time(s)\end{tabular} \\ \hline
\multicolumn{1}{c|}{Aftershock} & 51.8 & 87.9 & 1.0 & 7.0 & 3.7E-02 & 2.3E+02 \\
\multicolumn{1}{c|}{Archipelago} & 47.3 & 86.8 & 1.0 & 7.0 & 3.8E-02 & 2.1E+02 \\
\multicolumn{1}{c|}{BigGameHunters} & 44.3 & 86.1 & 1.0 & 7.0 & 3.9E-02 & 2.0E+02 \\
\multicolumn{1}{c|}{Brushfire} & 48.5 & 87.1 & 1.0 & 7.0 & 3.9E-02 & 3.6E+02 \\
\multicolumn{1}{c|}{Caldera} & 57.4 & 89.4 & 1.0 & 7.0 & 3.9E-02 & 2.0E+02 \\
\multicolumn{1}{c|}{Circle} & 85.7 & 96.4 & 1.0 & 7.0 & 3.7E-02 & 1.8E+02 \\
\multicolumn{1}{c|}{CrescentMoon} & 43.8 & 85.9 & 1.0 & 7.0 & 3.7E-02 & 2.0E+02 \\
\multicolumn{1}{c|}{Desolation} & 45.5 & 86.4 & 1.0 & 7.0 & 3.7E-02 & 2.4E+02 \\
\multicolumn{1}{c|}{EbonLakes} & 51.8 & 87.9 & 1.0 & 7.0 & 3.8E-02 & 2.1E+02 \\
\multicolumn{1}{c|}{Entanglement} & 53.0 & 88.2 & 1.0 & 7.0 & 3.8E-02 & 2.8E+02 \\
\multicolumn{1}{c|}{Eruption} & 48.8 & 87.2 & 1.0 & 7.0 & 3.7E-02 & 2.1E+02 \\
\multicolumn{1}{c|}{HotZone} & 54.5 & 88.6 & 1.0 & 7.0 & 3.8E-02 & 2.0E+02 \\
\multicolumn{1}{c|}{Isolation} & 58.3 & 89.6 & 1.0 & 7.0 & 3.8E-02 & 3.2E+02 \\
\multicolumn{1}{c|}{Legacy} & 38.1 & 84.5 & 1.0 & 7.0 & 3.7E-02 & 2.3E+02 \\
\multicolumn{1}{c|}{OrbitalGully} & 48.5 & 87.1 & 1.0 & 7.0 & 3.8E-02 & 2.0E+02 \\
\multicolumn{1}{c|}{Predators} & 58.0 & 89.5 & 1.0 & 7.0 & 3.8E-02 & 2.9E+02 \\
\multicolumn{1}{c|}{Ramparts} & 48.5 & 87.1 & 1.0 & 7.0 & 3.8E-02 & 2.0E+02 \\
\multicolumn{1}{c|}{RedCanyons} & 55.7 & 89.0 & 1.0 & 7.0 & 3.8E-02 & 2.9E+02 \\
\multicolumn{1}{c|}{Rosewood} & 49.7 & 87.4 & 1.0 & 7.0 & 3.8E-02 & 2.5E+02 \\
\multicolumn{1}{c|}{Sanctuary} & 42.6 & 85.6 & 1.0 & 7.0 & 3.8E-02 & 3.0E+02 \\
\multicolumn{1}{c|}{ShroudPlatform} & 50.9 & 87.7 & 1.0 & 7.0 & 3.8E-02 & 3.0E+02 \\
\multicolumn{1}{c|}{SpaceAtoll} & 33.3 & 83.3 & 1.0 & 7.0 & 3.8E-02 & 2.2E+02 \\
\multicolumn{1}{c|}{SpaceDebris} & 46.1 & 86.5 & 1.0 & 7.0 & 3.7E-02 & 1.9E+02 \\
\multicolumn{1}{c|}{SteppingStones} & 59.8 & 90.0 & 1.0 & 7.0 & 3.8E-02 & 2.6E+02 \\
\multicolumn{1}{c|}{Triskelion} & 51.2 & 87.8 & 1.0 & 7.0 & 3.8E-02 & 2.1E+02 \\
\multicolumn{1}{c|}{WarpGates} & 44.3 & 86.1 & 1.0 & 7.0 & 4.0E-02 & 2.2E+02 \\
\multicolumn{1}{c|}{WatersEdge} & 39.9 & 85.0 & 1.0 & 7.0 & 3.8E-02 & 2.9E+02 \\
\multicolumn{1}{c|}{WaypointJunction} & 42.0 & 85.5 & 1.0 & 7.0 & 4.0E-02 & 2.7E+02 \\
\multicolumn{1}{c|}{WinterConquest} & 47.3 & 86.8 & 1.0 & 7.0 & 3.8E-02 & 2.4E+02 \\ \hline
\multicolumn{1}{c|}{Average} & 49.8 & 87.4 & 1.0 & 7.0 & 3.80E-02 & 2.4E+02 \\ \hline
\end{tabular}
\caption{Comparison among continuous online goal recognition methods for each scenario. Vector methods results for $k=20$.}
\label{table:vector_k20}
\end{table*}

\begin{table*}[]
\centering
\setlength{\tabcolsep}{5pt}
\begin{tabular}{ccccccccc|cccccc}
\cline{4-15}
 &  &  & \multicolumn{6}{c|}{Mirroring} & \multicolumn{6}{c}{Landmark} \\ \hline
\multicolumn{1}{c|}{Scenario} & \multicolumn{1}{c|}{O} & \multicolumn{1}{c|}{G} & PPV(\%) & ACC(\%) & SPR & PC & \begin{tabular}[c]{@{}c@{}}Online\\ Time(s)\end{tabular} & \begin{tabular}[c]{@{}c@{}}Offline\\ Time(s)\end{tabular} & PPV(\%) & ACC(\%) & SPR & PC & \begin{tabular}[c]{@{}c@{}}Online\\ Time(s)\end{tabular} & \begin{tabular}[c]{@{}c@{}}Offline\\ Time(s)\end{tabular} \\ \hline
\multicolumn{1}{c|}{blocks-world} & \multicolumn{1}{c|}{8.8} & \multicolumn{1}{c|}{20.0} & 18.6 & 75.6 & 5.9 & 195.7 & 116.7 & 14.3 & 40.3 & 93.7 & 1.2 & 0.0 & 5.12E-01 & 5.72E-02 \\
\multicolumn{1}{c|}{depots} & \multicolumn{1}{c|}{9.5} & \multicolumn{1}{c|}{8.0} & 33.8 & 70.4 & 3.3 & 84.0 & 28.3 & 3.0 & 28.6 & 79.6 & 1.5 & 0.0 & 1.71E-01 & 6.35E-02 \\
\multicolumn{1}{c|}{driverlog} & \multicolumn{1}{c|}{12.2} & \multicolumn{1}{c|}{6.7} & 48.1 & 81.8 & 2.3 & 86.2 & 26.9 & 2.3 & 41.8 & 81.8 & 1.1 & 0.0 & 1.51E-01 & 3.96E-02 \\
\multicolumn{1}{c|}{dwr} & \multicolumn{1}{c|}{22.0} & \multicolumn{1}{c|}{6.7} & 39.1 & 74.6 & 2.7 & 147.3 & 95.3 & 3.6 & 38.0 & 79.2 & 1.4 & 0.0 & 1.46E+00 & 1.04E-01 \\
\multicolumn{1}{c|}{ipc-grid} & \multicolumn{1}{c|}{11.9} & \multicolumn{1}{c|}{7.5} & 55.8 & 85.3 & 2.1 & 100.0 & 28.0 & 2.4 & 45.7 & 77.5 & 2.4 & 0.0 & 3.90E-01 & 9.05E-02 \\
\multicolumn{1}{c|}{ferry} & \multicolumn{1}{c|}{18.8} & \multicolumn{1}{c|}{6.3} & 62.1 & 88.8 & 1.7 & 126.1 & 41.3 & 2.0 & 59.9 & 84.6 & 1.5 & 0.0 & 6.69E-01 & 1.48E-02 \\
\multicolumn{1}{c|}{logistics} & \multicolumn{1}{c|}{18.2} & \multicolumn{1}{c|}{10.0} & 52.3 & 89.5 & 2.1 & 191.7 & 50.7 & 2.8 & 39.5 & 85.2 & 1.8 & 0.0 & 6.49E-01 & 9.41E-02 \\
\multicolumn{1}{c|}{miconic} & \multicolumn{1}{c|}{16.3} & \multicolumn{1}{c|}{6.0} & 62.5 & 88.5 & 1.6 & 104.0 & 29.8 & 2.2 & 67.5 & 89.9 & 1.4 & 0.0 & 3.01E-01 & 4.03E-02 \\
\multicolumn{1}{c|}{rovers} & \multicolumn{1}{c|}{10.8} & \multicolumn{1}{c|}{6.0} & 51.0 & 82.6 & 1.9 & 71.0 & 55.8 & 2.1 & 57.4 & 85.4 & 1.3 & 0.0 & 1.66E-01 & 3.32E-02 \\
\multicolumn{1}{c|}{satellite} & \multicolumn{1}{c|}{10.8} & \multicolumn{1}{c|}{6.0} & 33.8 & 66.6 & 3.0 & 70.5 & 17.7 & 1.7 & 43.6 & 77.1 & 1.8 & 0.0 & 1.57E-01 & 2.53E-02 \\
\multicolumn{1}{c|}{small-sokoban} & \multicolumn{1}{c|}{26.0} & \multicolumn{1}{c|}{9.0} & 49.1 & 88.5 & 2.0 & 243.0 & 441.7 & 17.6 & 35.1 & 79.5 & 2.8 & 0.0 & 2.12E+00 & 9.59E-01 \\
\multicolumn{1}{c|}{zeno-travel} & \multicolumn{1}{c|}{12.0} & \multicolumn{1}{c|}{6.0} & 62.1 & 89.3 & 1.6 & 78.0 & 32.4 & 1.9 & 40.5 & 78.9 & 1.3 & 0.0 & 1.54E-01 & 6.38E-02 \\ \hline
\multicolumn{1}{c|}{Average} & \multicolumn{1}{c|}{14.8} & \multicolumn{1}{c|}{8.2} & 47.3 & 81.8 & 2.5 & 124.8 & 80.4 & 4.6 & 44.8 & 82.7 & 1.6 & 0.0 & 5.75E-01 & 1.32E-01 \\ \hline
\end{tabular}
\caption{Discrete domain experiments results of goal recognition using Mirroring and Landmarks methods.}
\label{table:discrete_mirroring_landmarks}
\end{table*}

\begin{table*}[]
\centering
\setlength{\tabcolsep}{5pt}
\begin{tabular}{ccccccccc|cccccc}
\cline{4-15}
 &  &  & \multicolumn{6}{c|}{GM+L} & \multicolumn{6}{c}{Vector k=1} \\ \hline
\multicolumn{1}{c|}{Scenario} & \multicolumn{1}{c|}{O} & \multicolumn{1}{c|}{G} & PPV(\%) & ACC(\%) & SPR & PC & \begin{tabular}[c]{@{}c@{}}Online\\ Time(s)\end{tabular} & \begin{tabular}[c]{@{}c@{}}Offline\\ Time(s)\end{tabular} & PPV(\%) & ACC(\%) & SPR & PC & \begin{tabular}[c]{@{}c@{}}Online\\ Time(s)\end{tabular} & \begin{tabular}[c]{@{}c@{}}Offline\\ Time(s)\end{tabular} \\ \hline
\multicolumn{1}{c|}{blocks-world} & \multicolumn{1}{c|}{8.8} & \multicolumn{1}{c|}{20.0} & 40.6 & 93.9 & 1.1 & 30.5 & 7.5 & 14.7 & 24.9 & 85.5 & 3.5 & 20.0 & 1.80E-03 & 1.31E+01 \\
\multicolumn{1}{c|}{depots} & \multicolumn{1}{c|}{9.5} & \multicolumn{1}{c|}{8.0} & 43.8 & 86.0 & 1.0 & 22.3 & 5.4 & 2.9 & 32.5 & 73.8 & 2.8 & 8.0 & 1.20E-03 & 2.75E+00 \\
\multicolumn{1}{c|}{driverlog} & \multicolumn{1}{c|}{12.2} & \multicolumn{1}{c|}{6.7} & 41.6 & 82.1 & 1.0 & 19.6 & 4.3 & 2.2 & 35.4 & 75.4 & 2.0 & 6.7 & 1.80E-03 & 2.08E+00 \\
\multicolumn{1}{c|}{dwr} & \multicolumn{1}{c|}{22.0} & \multicolumn{1}{c|}{6.7} & 39.0 & 80.1 & 1.4 & 34.3 & 20.9 & 3.8 & 37.4 & 74.1 & 2.6 & 6.7 & 2.40E-03 & 3.72E+00 \\
\multicolumn{1}{c|}{ipc-grid} & \multicolumn{1}{c|}{11.9} & \multicolumn{1}{c|}{7.5} & 52.1 & 84.9 & 1.2 & 39.9 & 12.8 & 2.8 & 49.6 & 80.7 & 2.3 & 7.5 & 6.80E-03 & 2.53E+00 \\
\multicolumn{1}{c|}{ferry} & \multicolumn{1}{c|}{18.8} & \multicolumn{1}{c|}{6.3} & 59.5 & 85.8 & 1.0 & 35.5 & 8.5 & 1.9 & 39.0 & 75.5 & 1.9 & 6.3 & 2.10E-03 & 1.73E+00 \\
\multicolumn{1}{c|}{logistics} & \multicolumn{1}{c|}{18.2} & \multicolumn{1}{c|}{10.0} & 46.1 & 88.6 & 1.3 & 41.8 & 9.2 & 2.8 & 39.7 & 83.1 & 2.4 & 10.0 & 2.70E-03 & 2.65E+00 \\
\multicolumn{1}{c|}{miconic} & \multicolumn{1}{c|}{16.3} & \multicolumn{1}{c|}{6.0} & 70.3 & 90.3 & 1.1 & 28.3 & 6.7 & 1.8 & 36.4 & 69.8 & 2.3 & 6.0 & 4.50E-03 & 1.70E+00 \\
\multicolumn{1}{c|}{rovers} & \multicolumn{1}{c|}{10.8} & \multicolumn{1}{c|}{6.0} & 65.5 & 88.0 & 1.1 & 20.1 & 11.6 & 2.2 & 36.7 & 68.0 & 2.8 & 6.0 & 2.30E-03 & 1.88E+01 \\
\multicolumn{1}{c|}{satellite} & \multicolumn{1}{c|}{10.8} & \multicolumn{1}{c|}{6.0} & 43.6 & 77.6 & 1.7 & 26.8 & 6.1 & 1.6 & 35.9 & 63.4 & 3.1 & 6.0 & 1.00E-03 & 1.70E+00 \\
\multicolumn{1}{c|}{small-sokoban} & \multicolumn{1}{c|}{26.0} & \multicolumn{1}{c|}{9.0} & 31.3 & 78.2 & 2.6 & 38.0 & 49.6 & 19.4 & 53.1 & 90.2 & 1.9 & 9.0 & 3.12E-02 & 1.96E+01 \\
\multicolumn{1}{c|}{zeno-travel} & \multicolumn{1}{c|}{12.0} & \multicolumn{1}{c|}{6.0} & 44.7 & 81.4 & 1.1 & 21.8 & 7.1 & 2.0 & 42.9 & 76.8 & 1.9 & 6.0 & 1.30E-03 & 1.12E+01 \\ \hline
\multicolumn{1}{c|}{Average} & \multicolumn{1}{c|}{14.8} & \multicolumn{1}{c|}{8.2} & 48.2 & 84.7 & 1.3 & 29.9 & 12.5 & 4.8 & 38.6 & 76.4 & 2.5 & 8.2 & 4.93E-03 & 6.79E+00 \\ \hline
\end{tabular}
\caption{Discrete domain experiments results of goal recognition using GM+L and Vector inference with $k=1$.}
\end{table*}

\begin{table*}[]
\centering
\setlength{\tabcolsep}{5pt}
\begin{tabular}{ccccccccc|cccccc}
\cline{4-15}
 &  &  & \multicolumn{6}{c|}{Vector k=5} & \multicolumn{6}{c}{Vector k=10} \\ \hline
\multicolumn{1}{c|}{Scenario} & \multicolumn{1}{c|}{O} & \multicolumn{1}{c|}{G} & PPV(\%) & ACC(\%) & SPR & PC & \begin{tabular}[c]{@{}c@{}}Online\\ Time(s)\end{tabular} & \begin{tabular}[c]{@{}c@{}}Offline\\ Time(s)\end{tabular} & PPV(\%) & ACC(\%) & SPR & PC & \begin{tabular}[c]{@{}c@{}}Online\\ Time(s)\end{tabular} & \begin{tabular}[c]{@{}c@{}}Offline\\ Time(s)\end{tabular} \\ \hline
\multicolumn{1}{c|}{blocks-world} & \multicolumn{1}{c|}{8.8} & \multicolumn{1}{c|}{20.0} & 36.1 & 90.1 & 2.4 & 20.0 & 4.8E-03 & 1.5E+01 & 39.7 & 90.7 & 2.3 & 20.0 & 9.2E-03 & 1.6E+01 \\
\multicolumn{1}{c|}{depots} & \multicolumn{1}{c|}{9.5} & \multicolumn{1}{c|}{8.0} & 43.9 & 81.8 & 1.9 & 8.0 & 4.0E-03 & 2.9E+00 & 44.3 & 82.2 & 1.9 & 8.0 & 7.0E-03 & 3.0E+00 \\
\multicolumn{1}{c|}{driverlog} & \multicolumn{1}{c|}{12.2} & \multicolumn{1}{c|}{6.7} & 39.2 & 77.7 & 1.8 & 6.7 & 4.5E-03 & 2.6E+00 & 48 & 84 & 1.2 & 6.7 & 9.1E-03 & 3.2E+00 \\
\multicolumn{1}{c|}{dwr} & \multicolumn{1}{c|}{22.0} & \multicolumn{1}{c|}{6.7} & 39.5 & 77.1 & 2.3 & 6.7 & 6.9E-03 & 4.0E+00 & 39.8 & 77.3 & 2.3 & 6.7 & 1.5E-02 & 4.3E+00 \\
\multicolumn{1}{c|}{ipc-grid} & \multicolumn{1}{c|}{11.9} & \multicolumn{1}{c|}{7.5} & 55.5 & 86.5 & 1.6 & 7.5 & 3.2E-02 & 2.7E+00 & 60.1 & 89 & 1.4 & 7.5 & 6.2E-02 & 3.0E+00 \\
\multicolumn{1}{c|}{ferry} & \multicolumn{1}{c|}{18.8} & \multicolumn{1}{c|}{6.3} & 58.3 & 85.9 & 1.1 & 6.3 & 7.8E-03 & 1.9E+00 & 65 & 88.7 & 1.1 & 6.3 & 1.4E-02 & 2.1E+00 \\
\multicolumn{1}{c|}{logistics} & \multicolumn{1}{c|}{18.2} & \multicolumn{1}{c|}{10.0} & 47.6 & 87.7 & 1.9 & 10.0 & 8.3E-03 & 3.0E+00 & 47.4 & 88.1 & 1.8 & 10.0 & 1.2E-02 & 2.9E+00 \\
\multicolumn{1}{c|}{miconic} & \multicolumn{1}{c|}{16.3} & \multicolumn{1}{c|}{6.0} & 40.9 & 77.3 & 1.7 & 6.0 & 2.0E-02 & 1.8E+00 & 45.9 & 80.3 & 1.5 & 6.0 & 4.6E-02 & 2.1E+00 \\
\multicolumn{1}{c|}{rovers} & \multicolumn{1}{c|}{10.8} & \multicolumn{1}{c|}{6.0} & 44.3 & 74.0 & 2.3 & 6.0 & 8.1E-03 & 7.1E+01 & 46.6 & 77.4 & 1.9 & 6.0 & 1.5E-02 & 1.3E+02 \\
\multicolumn{1}{c|}{satellite} & \multicolumn{1}{c|}{10.8} & \multicolumn{1}{c|}{6.0} & 39.0 & 69.6 & 2.7 & 6.0 & 3.3E-03 & 2.4E+00 & 40.7 & 72.4 & 2.5 & 6.0 & 5.8E-03 & 2.8E+00 \\
\multicolumn{1}{c|}{small-sokoban} & \multicolumn{1}{c|}{26.0} & \multicolumn{1}{c|}{9.0} & 56.5 & 91.5 & 1.8 & 9.0 & 1.5E-01 & 2.1E+01 & 53.7 & 90.2 & 1.6 & 9.0 & 3.3E-01 & 2.5E+01 \\
\multicolumn{1}{c|}{zeno-travel} & \multicolumn{1}{c|}{12.0} & \multicolumn{1}{c|}{6.0} & 50.4 & 82.3 & 1.5 & 6.0 & 3.9E-03 & 4.5E+01 & 53 & 83.7 & 1.4 & 6.0 & 7.3E-03 & 8.5E+01 \\ \hline
\multicolumn{1}{c|}{Average} & \multicolumn{1}{c|}{14.8} & \multicolumn{1}{c|}{8.2} & 45.9 & 81.8 & 1.9 & 8.2 & 2.1E-02 & 1.4E+01 & 48.7 & 83.7 & 1.7 & 8.2 & 4.4E-02 & 2.4E+01 \\ \hline
\end{tabular}
\caption{Discrete domain experiments results of goal recognition using Vector inference method with $k=5$ and $k=10$.}
\end{table*}

\begin{table*}[]
\centering
\setlength{\tabcolsep}{5pt}
\begin{tabular}{ccccccccc}
\cline{4-9}
 & \multicolumn{1}{l}{} & \multicolumn{1}{l}{} & \multicolumn{6}{c}{Vector k=15} \\ \hline
\multicolumn{1}{c|}{Scenario} & \multicolumn{1}{c|}{O} & \multicolumn{1}{c|}{G} & PPV(\%) & ACC(\%) & SPR & PC & \begin{tabular}[c]{@{}c@{}}Online\\ Time(s)\end{tabular} & \begin{tabular}[c]{@{}c@{}}Offline\\ Time(s)\end{tabular} \\ \hline
\multicolumn{1}{c|}{blocks-world} & \multicolumn{1}{c|}{8.8} & \multicolumn{1}{c|}{20.0} & 37.3 & 91.1 & 2.0 & 20.0 & 1.1E-02 & 1.6E+01 \\
\multicolumn{1}{c|}{depots} & \multicolumn{1}{c|}{9.5} & \multicolumn{1}{c|}{8.0} & 41.4 & 81.4 & 2.0 & 8.0 & 8.1E-03 & 3.2E+00 \\
\multicolumn{1}{c|}{driverlog} & \multicolumn{1}{c|}{12.2} & \multicolumn{1}{c|}{6.7} & 47.6 & 83.9 & 1.2 & 6.7 & 1.6E-02 & 3.9E+00 \\
\multicolumn{1}{c|}{dwr} & \multicolumn{1}{c|}{22.0} & \multicolumn{1}{c|}{6.7} & 39.5 & 77.1 & 2.2 & 6.7 & 2.0E-02 & 4.4E+00 \\
\multicolumn{1}{c|}{ipc-grid} & \multicolumn{1}{c|}{11.9} & \multicolumn{1}{c|}{7.5} & 57 & 87.4 & 1.4 & 7.5 & 9.6E-02 & 3.2E+00 \\
\multicolumn{1}{c|}{ferry} & \multicolumn{1}{c|}{18.8} & \multicolumn{1}{c|}{6.3} & 68.4 & 89.7 & 1.0 & 6.3 & 1.9E-02 & 2.2E+00 \\
\multicolumn{1}{c|}{logistics} & \multicolumn{1}{c|}{18.2} & \multicolumn{1}{c|}{10.0} & 48.9 & 88.6 & 1.7 & 10.0 & 1.6E-02 & 2.9E+00 \\
\multicolumn{1}{c|}{miconic} & \multicolumn{1}{c|}{16.3} & \multicolumn{1}{c|}{6.0} & 49.6 & 82 & 1.4 & 6.0 & 6.1E-02 & 2.2E+00 \\
\multicolumn{1}{c|}{rovers} & \multicolumn{1}{c|}{10.8} & \multicolumn{1}{c|}{6.0} & 49.1 & 80.7 & 1.6 & 6.0 & 2.4E-02 & 2.0E+02 \\
\multicolumn{1}{c|}{satellite} & \multicolumn{1}{c|}{10.8} & \multicolumn{1}{c|}{6.0} & 40.5 & 72.4 & 2.5 & 6.0 & 1.0E-02 & 3.9E+00 \\
\multicolumn{1}{c|}{small-sokoban} & \multicolumn{1}{c|}{26.0} & \multicolumn{1}{c|}{9.0} & 53.7 & 90.2 & 1.6 & 9.0 & 4.7E-01 & 3.3E+01 \\
\multicolumn{1}{c|}{zeno-travel} & \multicolumn{1}{c|}{12.0} & \multicolumn{1}{c|}{6.0} & 57.3 & 85.7 & 1.3 & 6.0 & 1.2E-02 & 1.3E+02 \\ \hline
\multicolumn{1}{c|}{Average} & \multicolumn{1}{c|}{14.8} & \multicolumn{1}{c|}{8.2} & 49.2 & 84.2 & 1.7 & 8.2 & 6.4E-02 & 3.4E+01 \\ \hline
\end{tabular}
\caption{Discrete domain experiments results of goal recognition using Vector inference method with $k=15$}
\label{table:discrete_vector_k15}
\end{table*}

\begin{table*}[h]
\centering
\setlength{\tabcolsep}{5pt}
\begin{tabular}{cccccccccc}
\cline{4-9}
 & \multicolumn{1}{l}{} & \multicolumn{1}{l}{} & \multicolumn{6}{c}{Vector k=50} \\ \hline
\multicolumn{1}{c|}{Scenario} & \multicolumn{1}{c|}{O} & \multicolumn{1}{c|}{G} & PPV(\%) & ACC(\%) & SPR & PC & \begin{tabular}[c]{@{}c@{}}Online\\ Time(s)\end{tabular} & \begin{tabular}[c]{@{}c@{}}Offline\\ Time(s)\end{tabular} &
\multicolumn{1}{c}{OPS} \\
\hline
\multicolumn{1}{c|}{blocks-world} & \multicolumn{1}{c|}{8.8} & \multicolumn{1}{c|}{20.0}  & 36.0\% & 91.5\% & 1.9 & 20.0 & 1.6E-02 & 1.4E+01 & 13.8 \\
\multicolumn{1}{c|}{depots} & \multicolumn{1}{c|}{9.5} & \multicolumn{1}{c|}{8.0} & 49.0\% & 84.5\% & 1.7 & 8.0 & 8.0E-03 & 2.7E+00 & 10.2  \\
\multicolumn{1}{c|}{driverlog} & \multicolumn{1}{c|}{12.2} & \multicolumn{1}{c|}{6.7} & 41.3\% & 80.7\% & 1.4 & 6.7 & 4.3E-02 & 7.2E+00 & 36.7 \\
\multicolumn{1}{c|}{dwr} & \multicolumn{1}{c|}{22.0} & \multicolumn{1}{c|}{6.7} & 39.3\% & 77.4\% & 2.2 & 6.7 & 7.0E-02 & 5.6E+00 & 40.2 \\
\multicolumn{1}{c|}{ipc-grid} & \multicolumn{1}{c|}{11.9} & \multicolumn{1}{c|}{7.5}  & 58.4\% & 86.6\% & 1.9 & 7.5 & 2.0E-02 & 2.8E+00 & 2.1 \\
\multicolumn{1}{c|}{ferry} & \multicolumn{1}{c|}{18.8} & \multicolumn{1}{c|}{6.3} & 85.2\% & 95.2\% & 1.0 & 6.3 & 5.4E-02 & 2.9E+00 & 35.0 \\
\multicolumn{1}{c|}{logistics} & \multicolumn{1}{c|}{18.2} & \multicolumn{1}{c|}{10.0} & 56.8\% & 90.6\% & 1.7 & 10.0 & 8.7E-02 & 4.0E+00 & 50.0 \\
\multicolumn{1}{c|}{miconic} & \multicolumn{1}{c|}{16.3} & \multicolumn{1}{c|}{6.0} & 55.3\% & 84.3\% & 1.3 & 6.0 & 2.4E-01 & 3.2E+00 & 50.0  \\
\multicolumn{1}{c|}{rovers} & \multicolumn{1}{c|}{10.8} & \multicolumn{1}{c|}{6.0} & 59.1\% & 86.5\% & 1.6 & 6.0 & 9.3E-02 & 6.7E+02 & 50.0 \\
\multicolumn{1}{c|}{satellite} & \multicolumn{1}{c|}{10.8} & \multicolumn{1}{c|}{6.0} & 42.5\% & 73.7\% & 2.3 & 6.0 & 2.8E-02 & 6.6E+00 & 33.4 \\
\multicolumn{1}{c|}{small-sokoban} & \multicolumn{1}{c|}{26.0} & \multicolumn{1}{c|}{9.0} & 57.8\% & 91.9\% & 1.7 & 9.0 & 8.5E-01 & 3.1E+01 & 25.1 \\
\multicolumn{1}{c|}{zeno-travel} & \multicolumn{1}{c|}{12.0} & \multicolumn{1}{c|}{6.0} & 64.0\% & 88.1\% & 1.2 & 6.0 & 4.2E-02 & 3.2E+02 & 48.1 \\ \hline
\multicolumn{1}{c|}{Average} & \multicolumn{1}{c|}{14.8} & \multicolumn{1}{c|}{8.2} & 53.7\% & 85.9\% & 1.7 & 8.2 & 1.3E-01 & 8.9E+01 & 32.9 \\ \hline
\end{tabular}
\caption{Discrete domain experiments results of goal recognition using Vector inference method with $k=50$ and filtering for optimal plan solutions (OPS).}
\label{table:discrete_vector_k50}
\end{table*}

\begin{table}[]
\centering
\begin{tabular}{c|c}
\hline
Scenario & Vacancy \\ \hline
Aftershock & 63.4\% \\
Archipelago & 50.3\% \\
BigGameHunters & 68.3\% \\
Brushfire & 40.0\% \\
Caldera & 62.9\% \\
CatwalkAlley & 86.2\% \\
Circle & 97.0\% \\
CrescentMoon & 46.7\% \\
Desolation & 56.8\% \\
EbonLakes & 66.3\% \\
Entanglement & 55.3\% \\
Eruption & 71.3\% \\
HotZone & 69.1\% \\
IceMountain & 70.4\% \\
Isolation & 78.5\% \\
Legacy & 70.9\% \\
NovaStation & 35.4\% \\
OrbitalGully & 75.3\% \\
Predators & 52.4\% \\
Ramparts & 58.0\% \\
RedCanyons & 66.7\% \\
Rosewood & 52.8\% \\
Sanctuary & 61.7\% \\
ShroudPlatform & 66.5\% \\
SpaceAtoll & 62.6\% \\
SpaceDebris & 54.5\% \\
SteppingStones & 58.9\% \\
Triskelion & 69.2\% \\
WarpGates & 53.9\% \\
WatersEdge & 51.6\% \\
WaypointJunction & 58.2\% \\
WinterConquest & 60.0\% \\ \hline
\end{tabular}
\caption{Vacancy percentage of each scenario.}
\label{table:vacancy}
\end{table}

\clearpage

\begin{figure*}[h]
    \centering
    
    \begin{subfigure}{0.45\textwidth}
        \includegraphics[width=\linewidth]{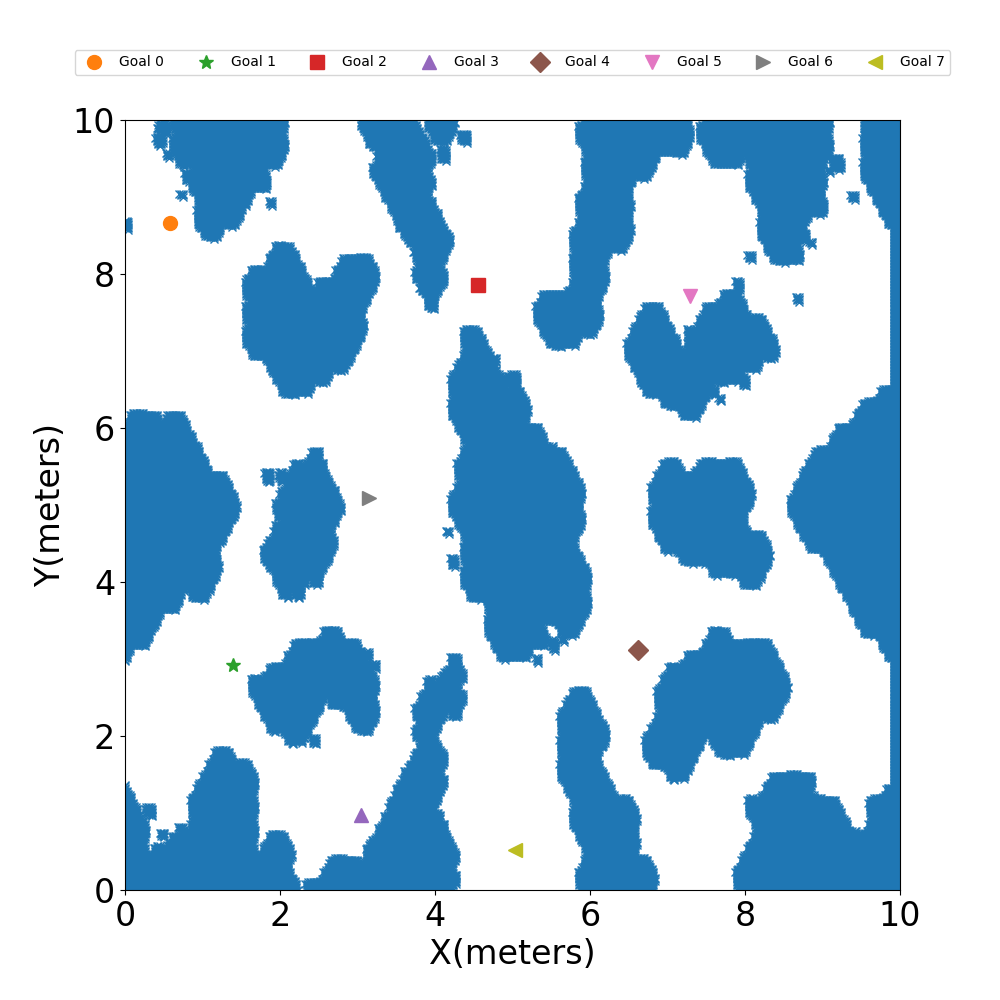}
        \caption{Caldera scenario.}
    \end{subfigure}
    \hfill
    \begin{subfigure}{0.45\textwidth}
        \includegraphics[width=\linewidth]{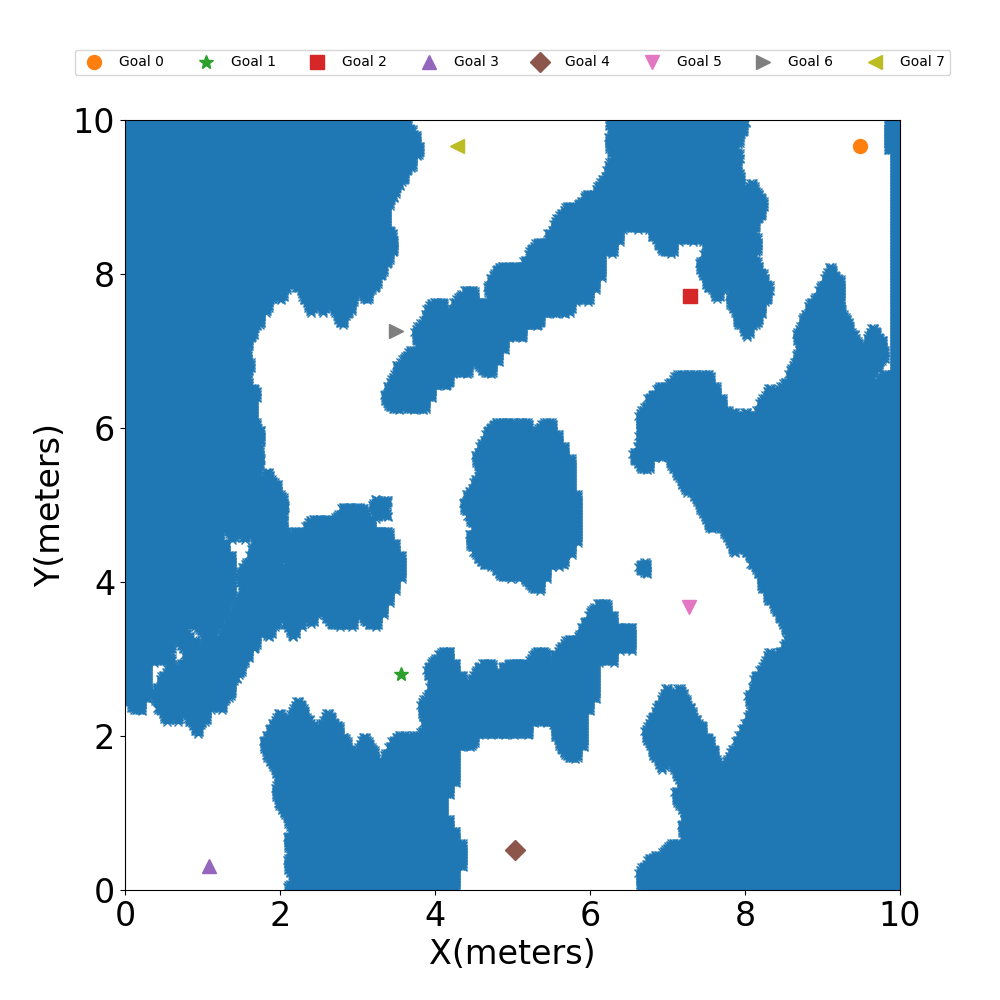}
        \caption{CrescentMoon scenario.}
    \end{subfigure}
    
    \vspace{\baselineskip}
    
    \begin{subfigure}{0.45\textwidth}
        \includegraphics[width=\linewidth]{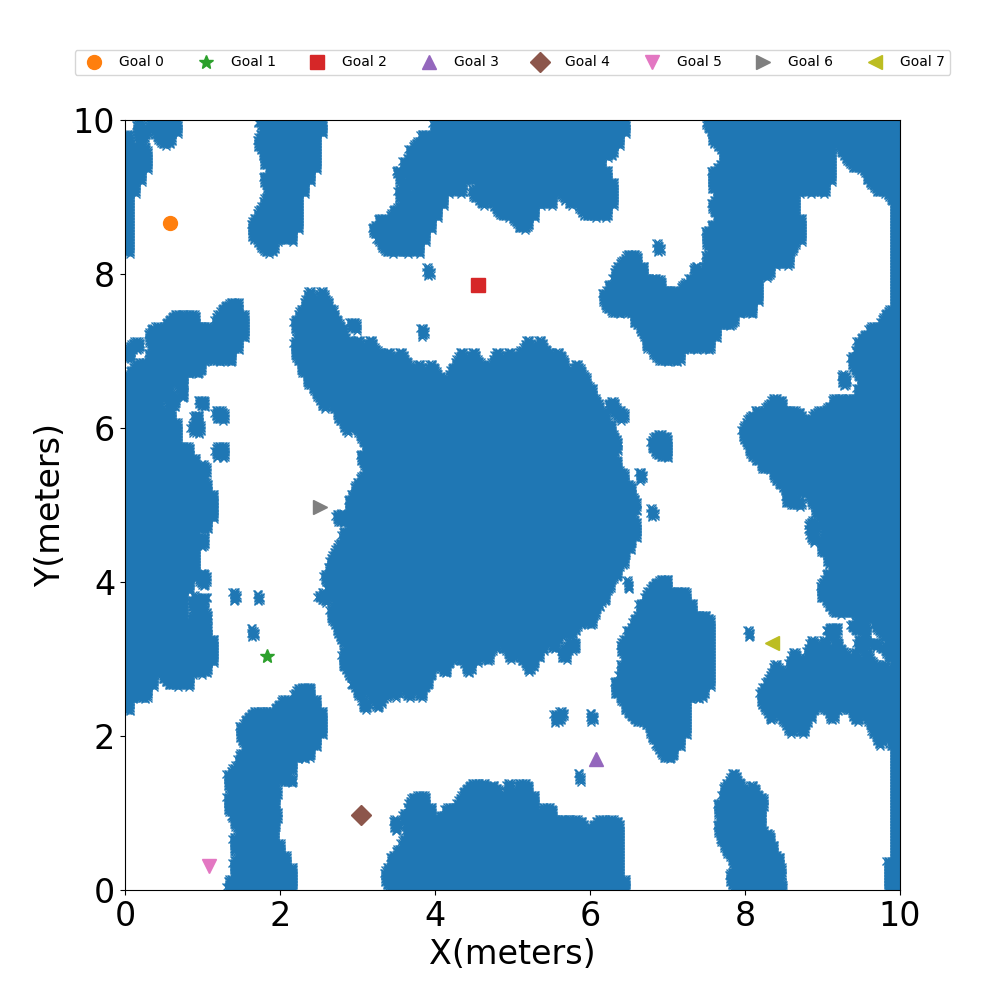}
        \caption{Desolation scenario.}
    \end{subfigure}
    \hfill
    \begin{subfigure}{0.45\textwidth}
        \includegraphics[width=\linewidth]{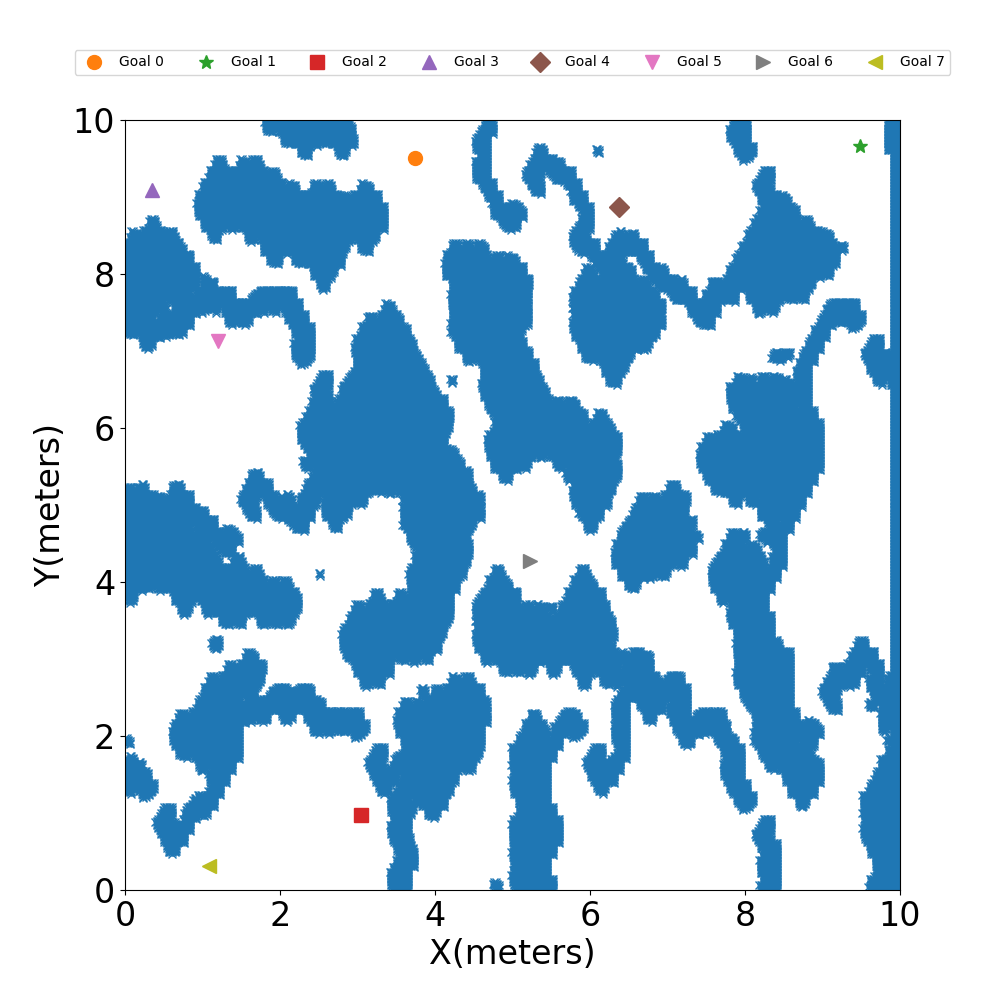}
        \caption{EbonLakes scenario.}
    \end{subfigure}
    \vspace*{1em}
    \caption{Caldera, CrescentMoon, Desolation, and EbonLakes scenarios used in the experiments.}
\end{figure*}

\begin{figure*}[h]
    \centering
    
    \begin{subfigure}{0.45\textwidth}
        \includegraphics[width=\linewidth]{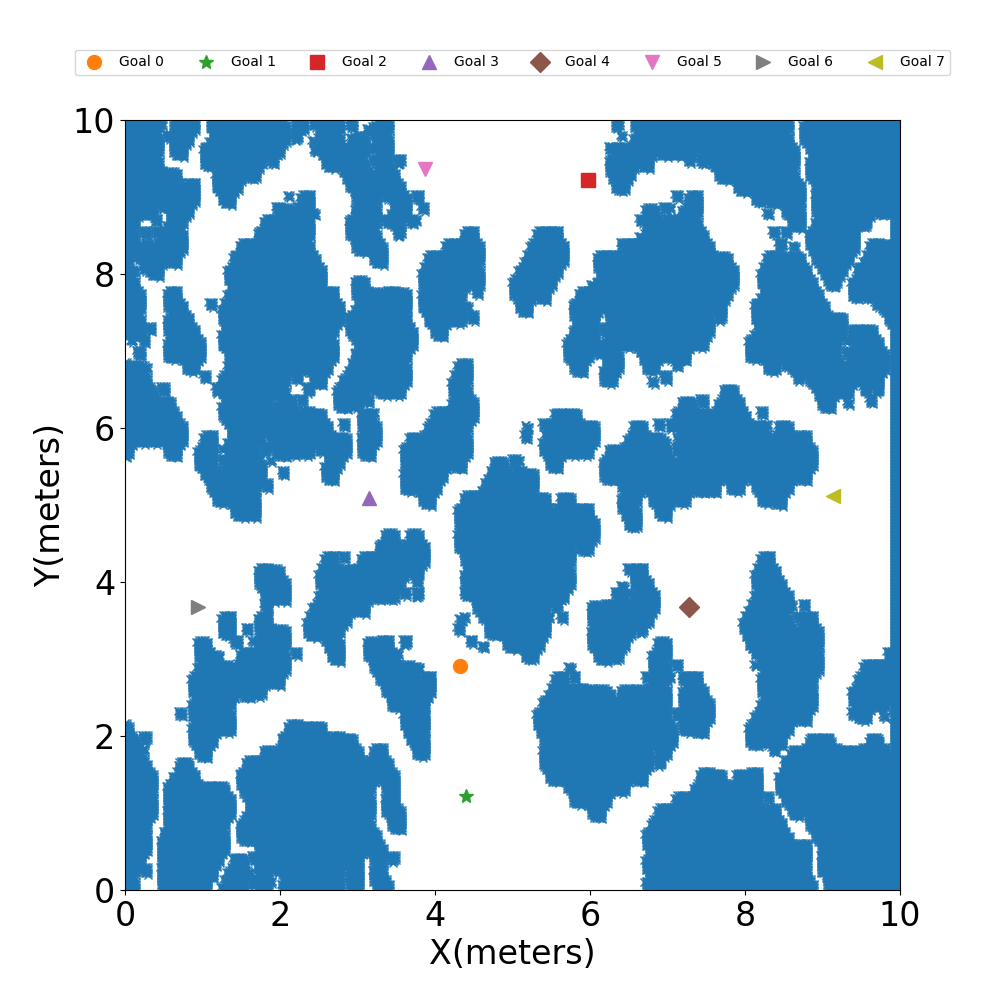}
        \caption{Entanglement scenario.}
    \end{subfigure}
    \hfill
    \begin{subfigure}{0.45\textwidth}
        \includegraphics[width=\linewidth]{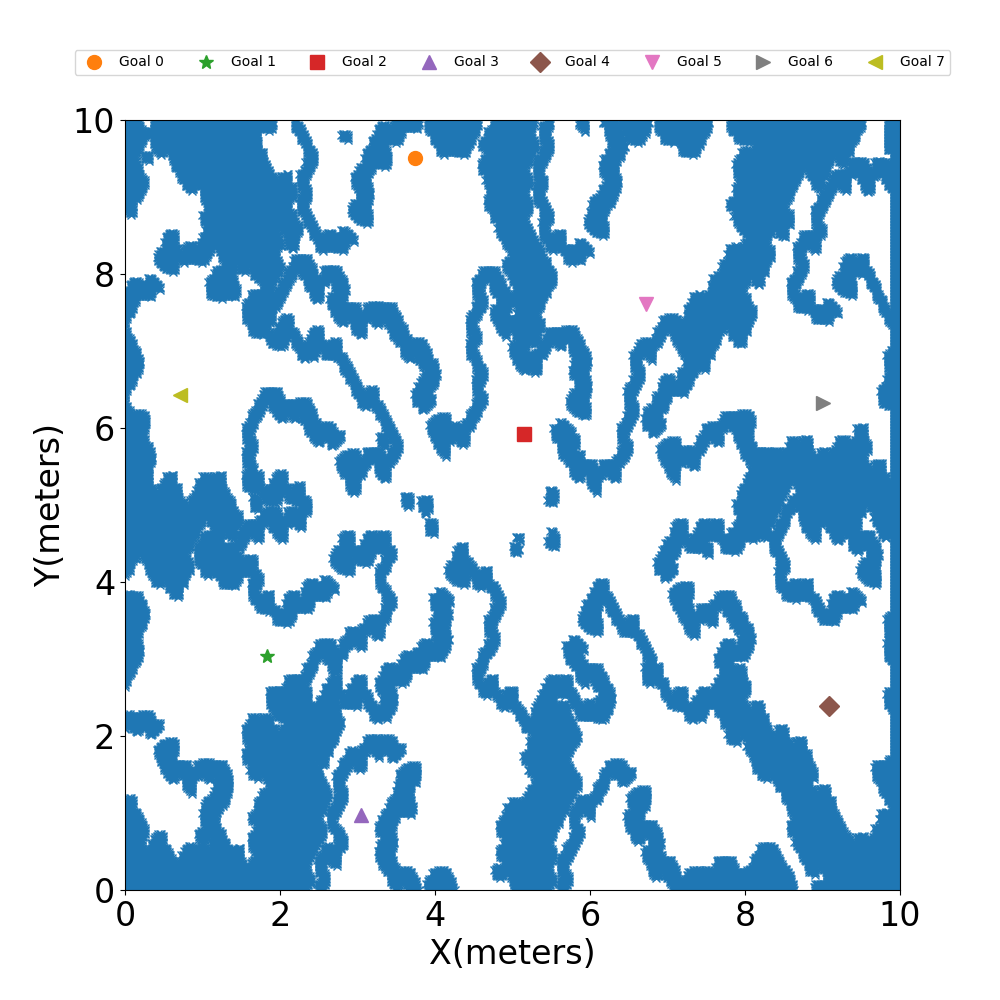}
        \caption{Eruption scenario.}
    \end{subfigure}
    
    \vspace{\baselineskip}
    
    \begin{subfigure}{0.45\textwidth}
        \includegraphics[width=\linewidth]{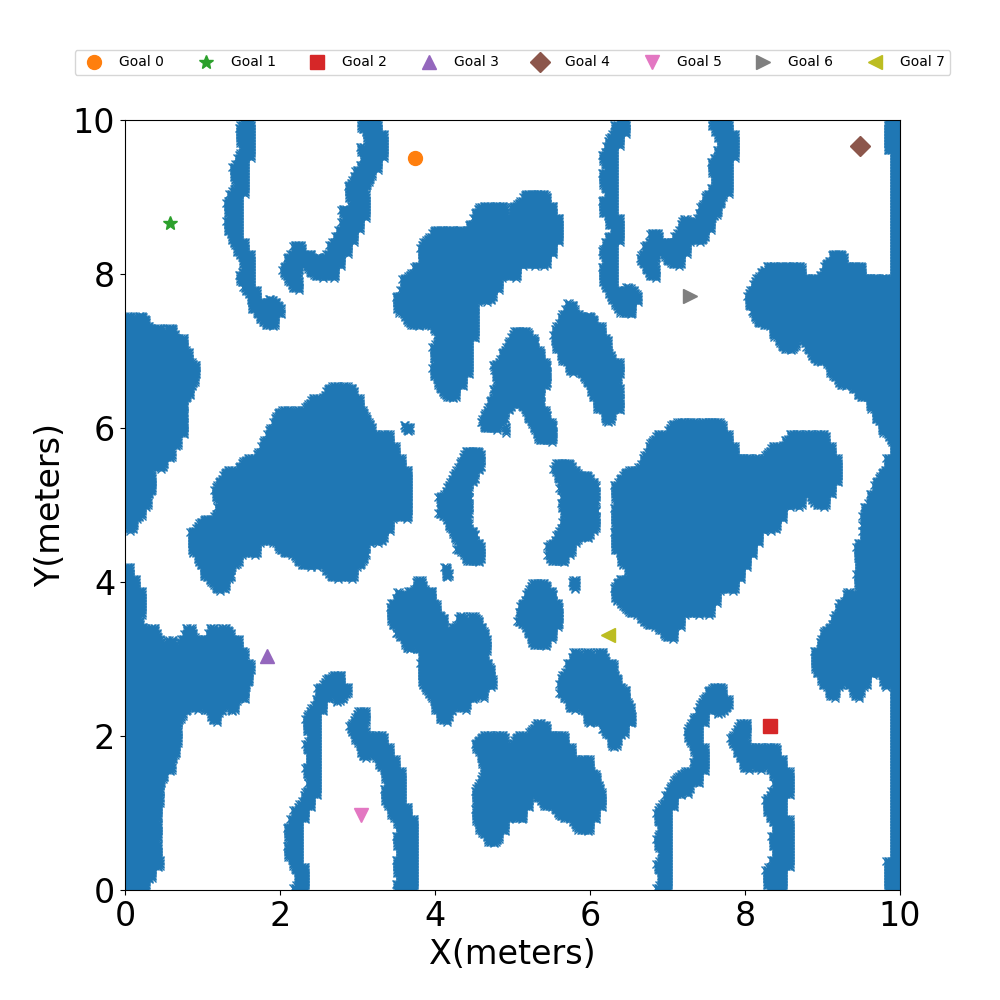}
        \caption{HotZone scenario.}
    \end{subfigure}
    \hfill
    \begin{subfigure}{0.45\textwidth}
        \includegraphics[width=\linewidth]{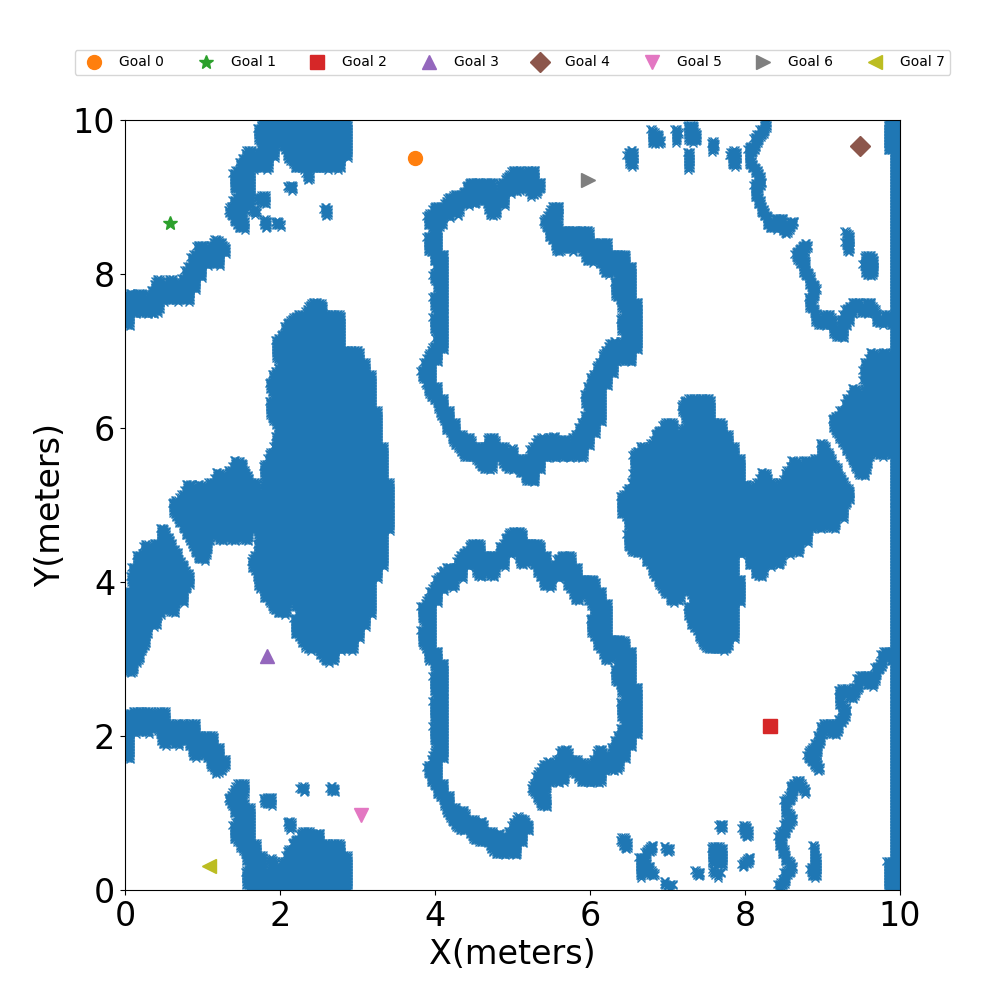}
        \caption{Isolation scenario.}
    \end{subfigure}
    \vspace*{1em}
    \caption{Entanglement, Eruption, HotZone, and Isolation scenarios used in the experiments.}
\end{figure*}

\begin{figure*}[h]
    \centering
    
    \begin{subfigure}{0.45\textwidth}
        \includegraphics[width=\linewidth]{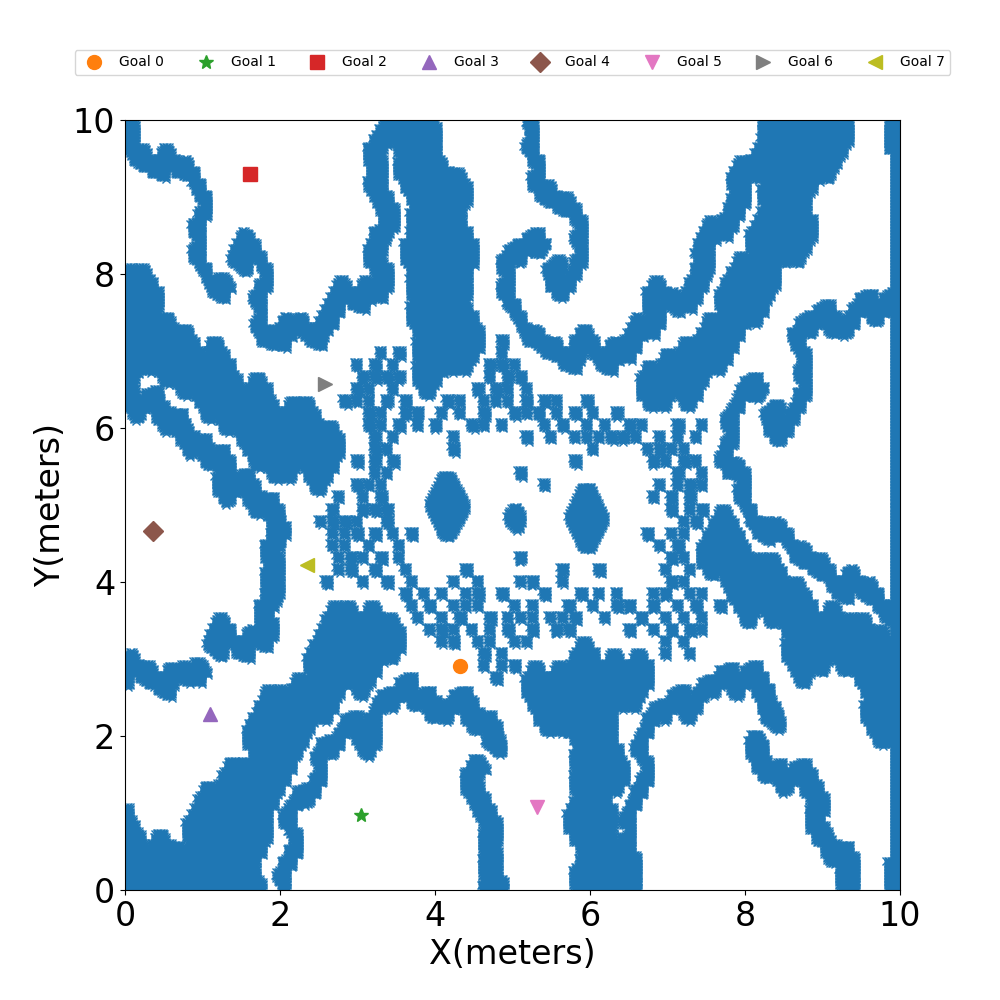}
        \caption{Legacy scenario.}
    \end{subfigure}
    \hfill
    \begin{subfigure}{0.45\textwidth}
        \includegraphics[width=\linewidth]{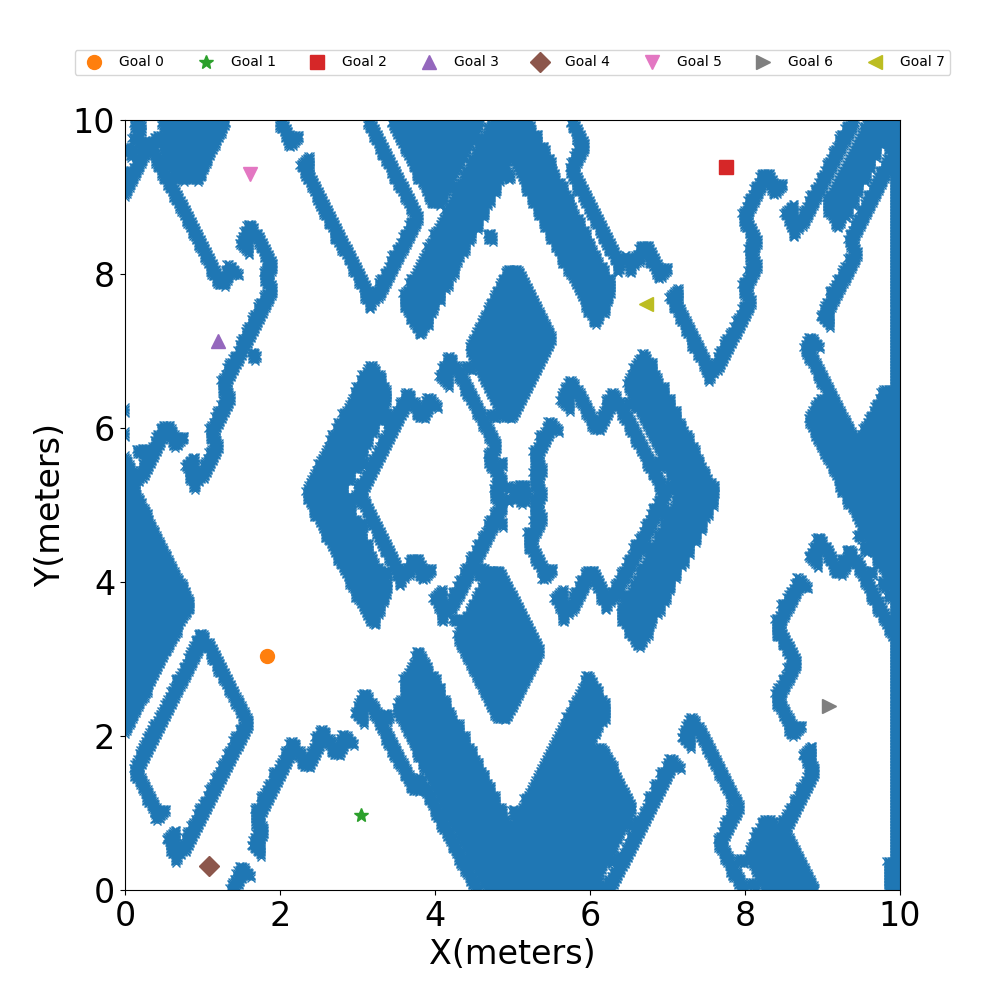}
        \caption{OrbitalGully scenario.}
    \end{subfigure}
    
    \vspace{\baselineskip}
    
    \begin{subfigure}{0.45\textwidth}
        \includegraphics[width=\linewidth]{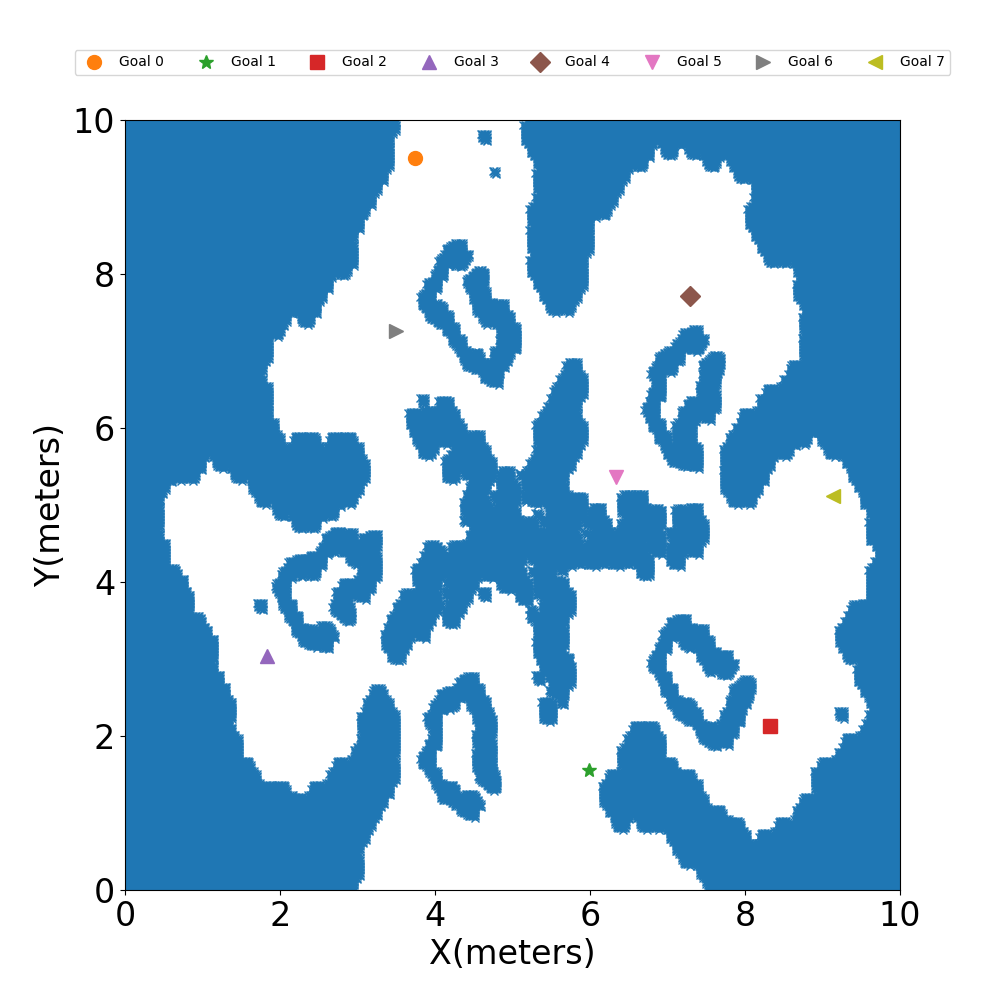}
        \caption{Predators scenario.}
    \end{subfigure}
    \hfill
    \begin{subfigure}{0.45\textwidth}
        \includegraphics[width=\linewidth]{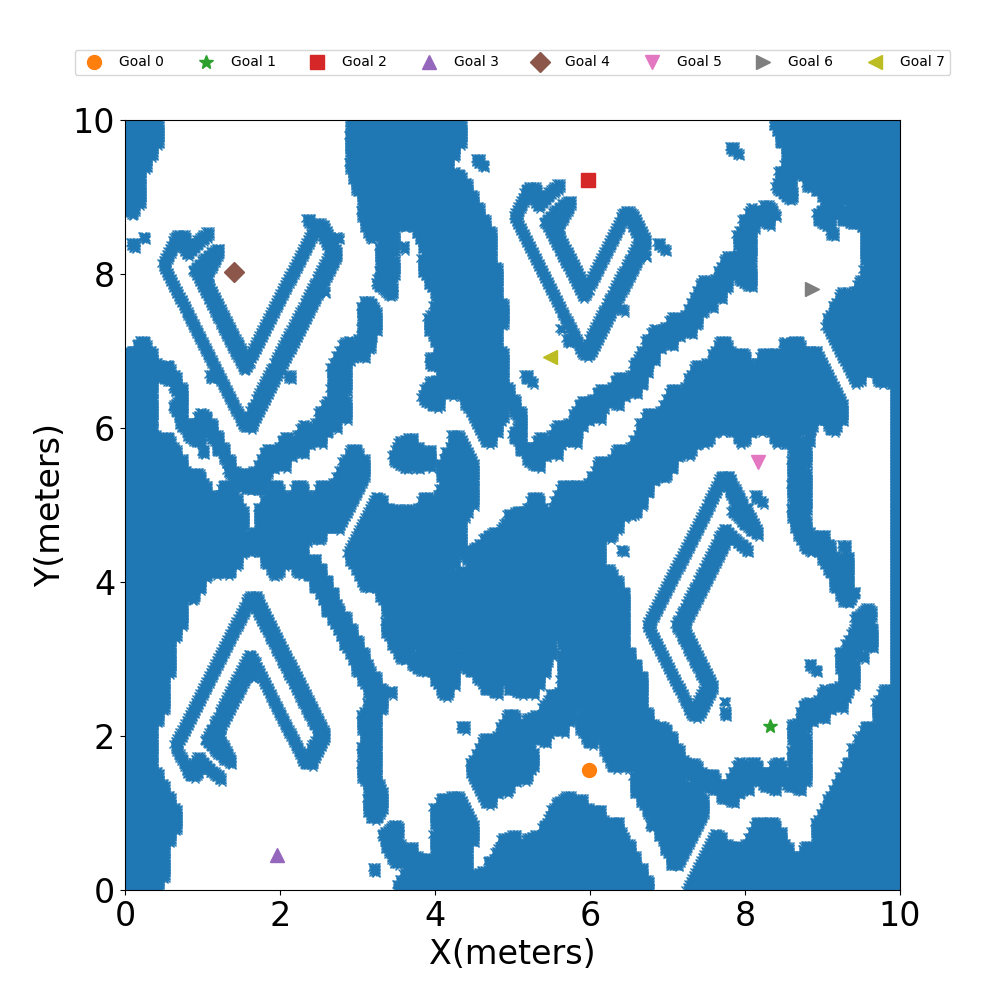}
        \caption{Ramparts scenario.}
    \end{subfigure}
    \vspace*{1em}
    \caption{Legacy, OrbitalGully, Predators, and Ramparts scenarios used in the experiments.}
\end{figure*}

\begin{figure*}[h]
    \centering
    
    \begin{subfigure}{0.45\textwidth}
        \includegraphics[width=\linewidth]{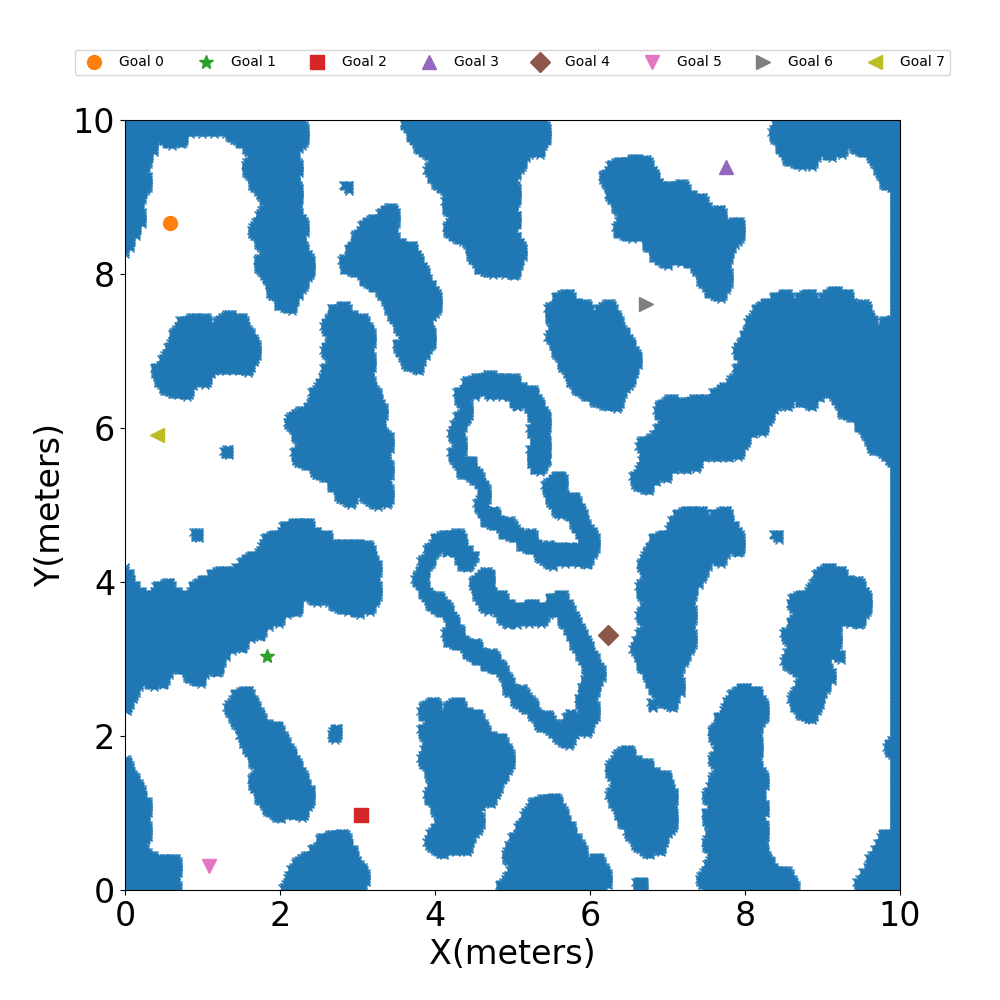}
        \caption{RedCanyons scenario.}
    \end{subfigure}
    \hfill
    \begin{subfigure}{0.45\textwidth}
        \includegraphics[width=\linewidth]{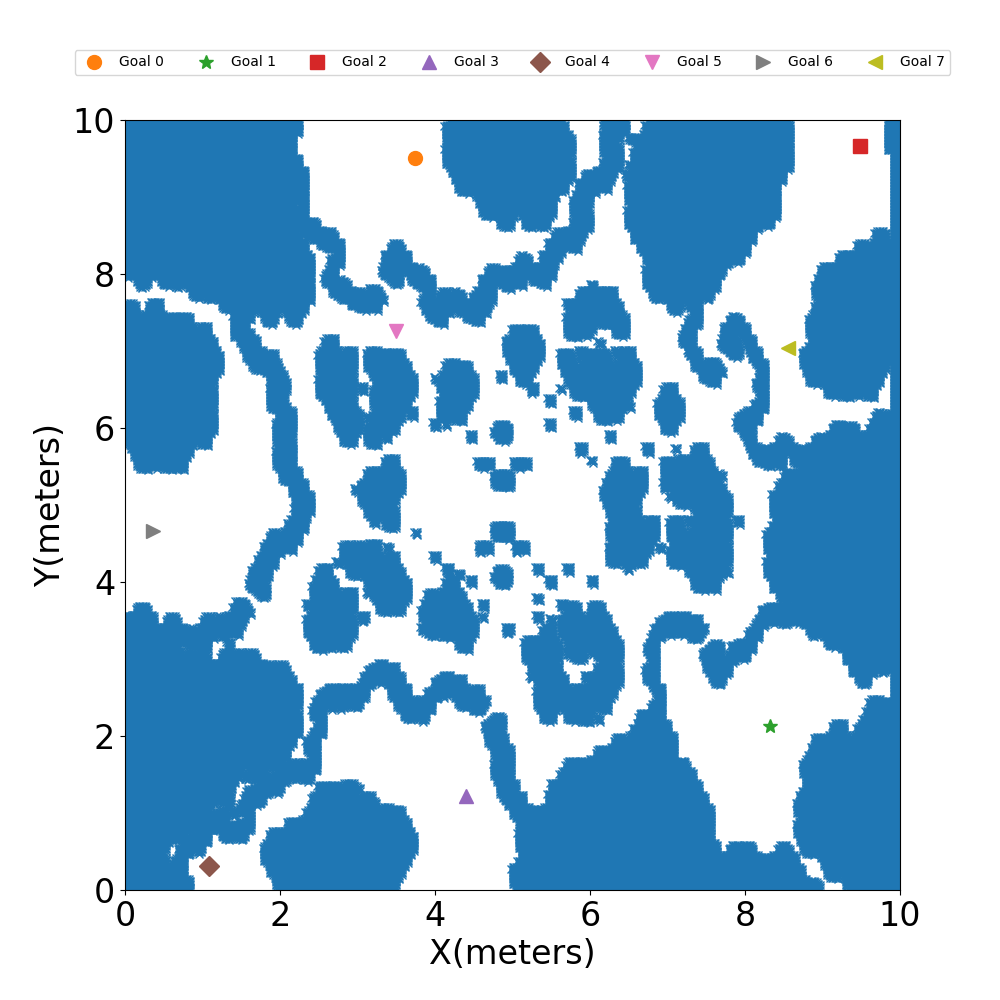}
        \caption{Rosewood scenario.}
    \end{subfigure}
    
    \vspace{\baselineskip}
    
    \begin{subfigure}{0.45\textwidth}
        \includegraphics[width=\linewidth]{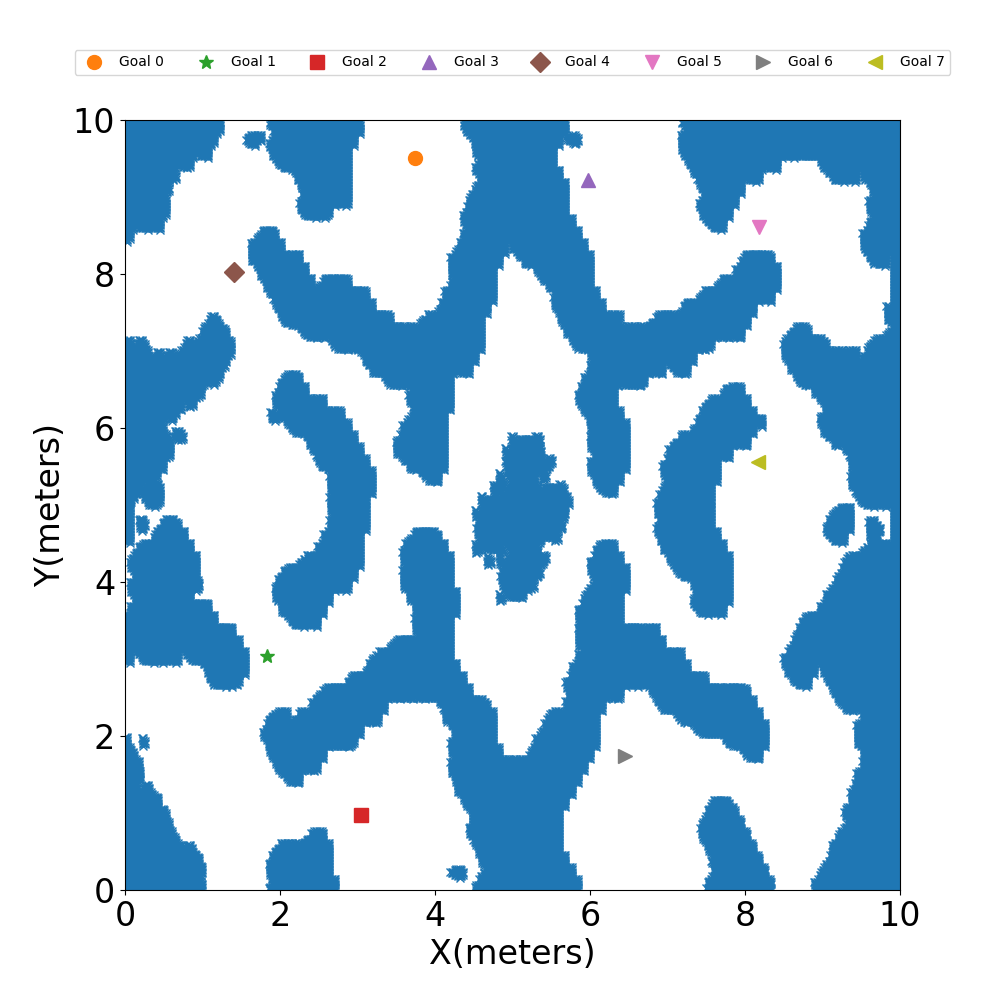}
        \caption{Sanctuary scenario.}
    \end{subfigure}
    \hfill
    \begin{subfigure}{0.45\textwidth}
        \includegraphics[width=\linewidth]{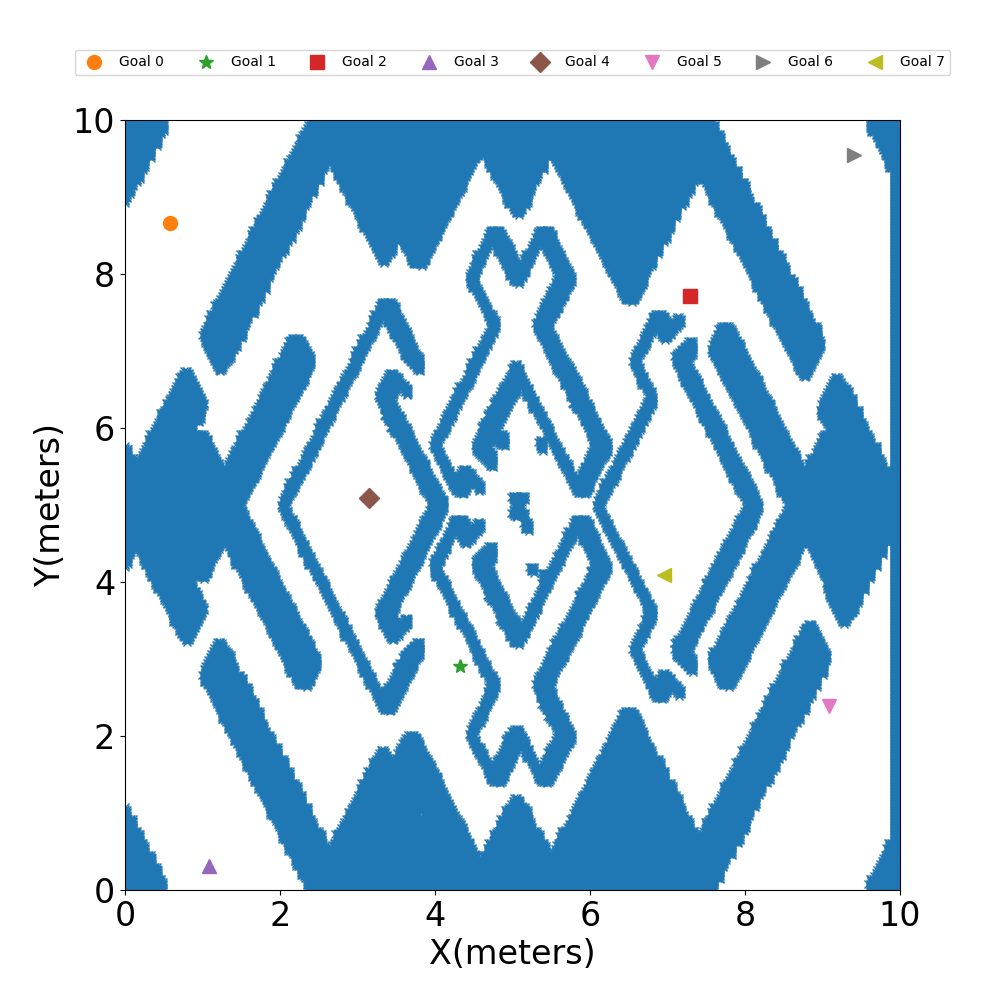}
        \caption{ShroudPlatform scenario.}
    \end{subfigure}
    \vspace*{1em}
    \caption{RedCanyons, Rosewood, Sanctuary, and ShroudPlatform scenarios used in the experiments.}
\end{figure*}

\begin{figure*}[h]
    \centering
    
    \begin{subfigure}{0.45\textwidth}
        \includegraphics[width=\linewidth]{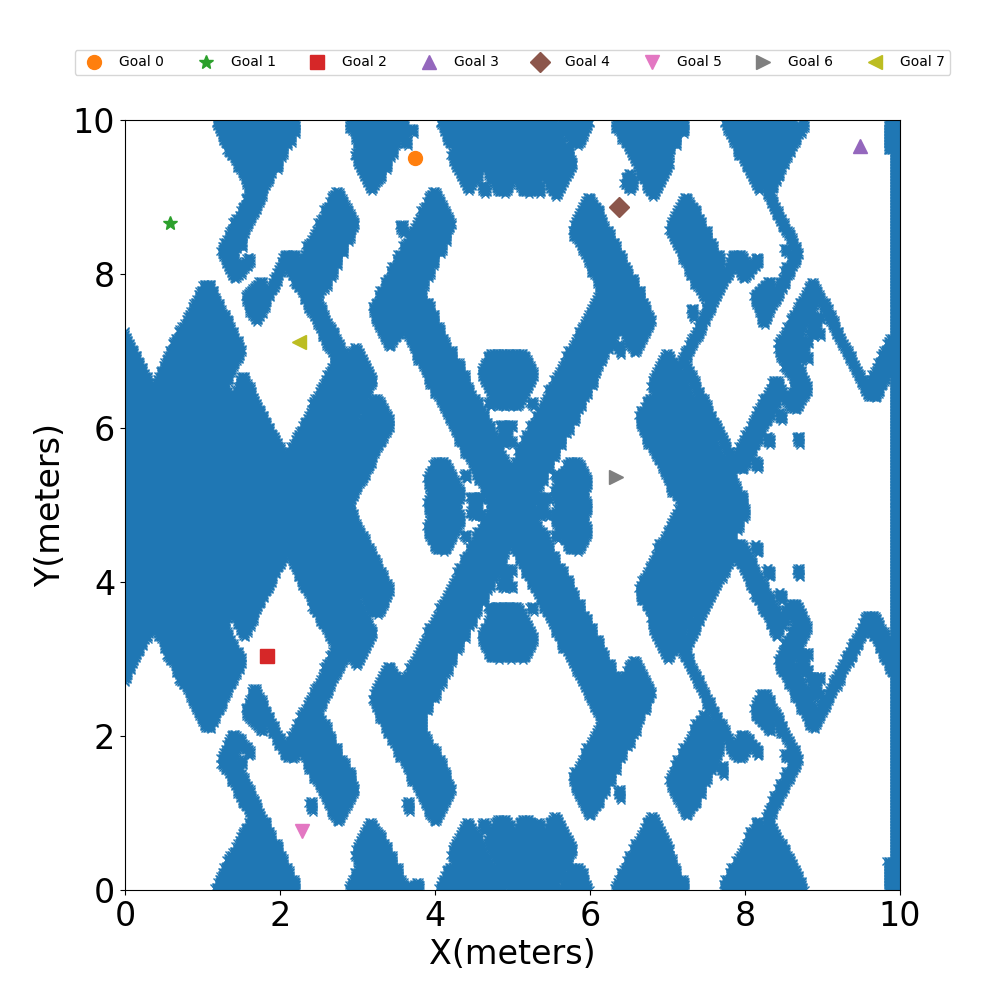}
        \caption{SpaceAtoll scenario.}
    \end{subfigure}
    \hfill
    \begin{subfigure}{0.45\textwidth}
        \includegraphics[width=\linewidth]{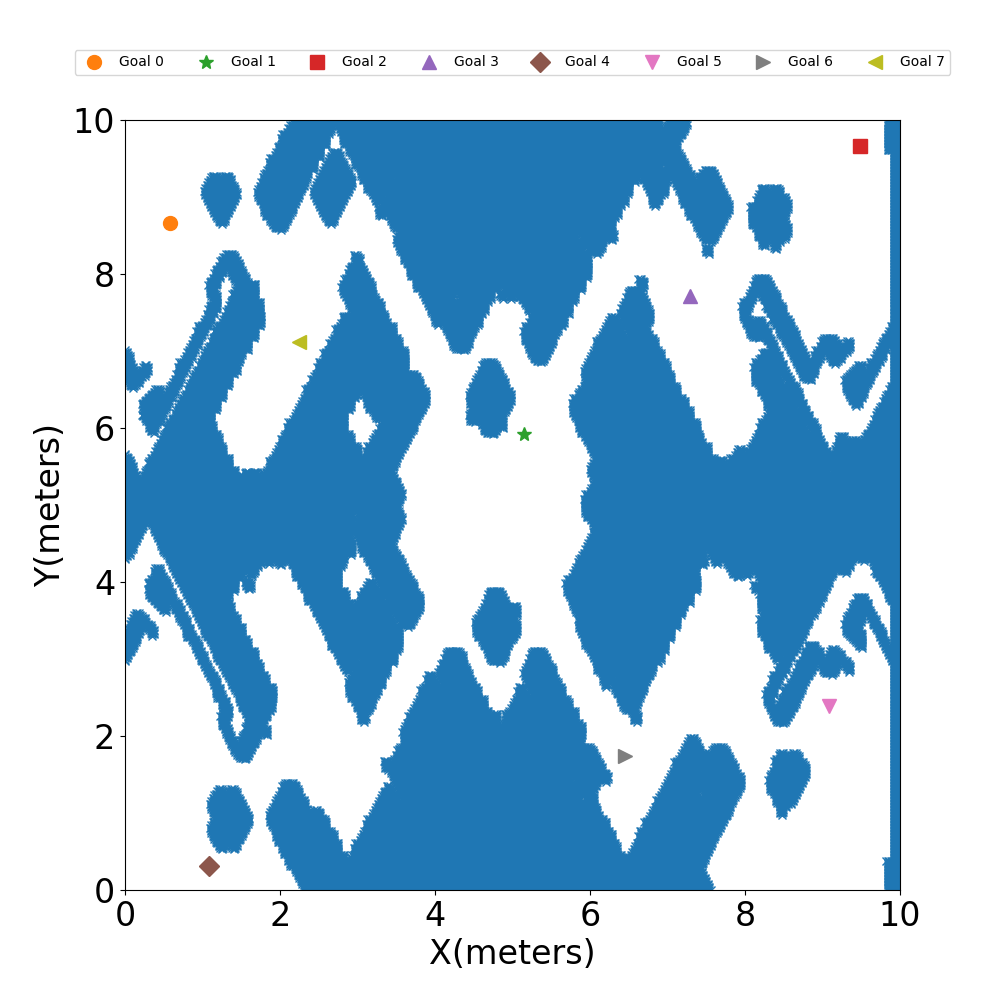}
        \caption{SpaceDebris scenario.}
    \end{subfigure}
    
    \vspace{\baselineskip}
    
    \begin{subfigure}{0.45\textwidth}
        \includegraphics[width=\linewidth]{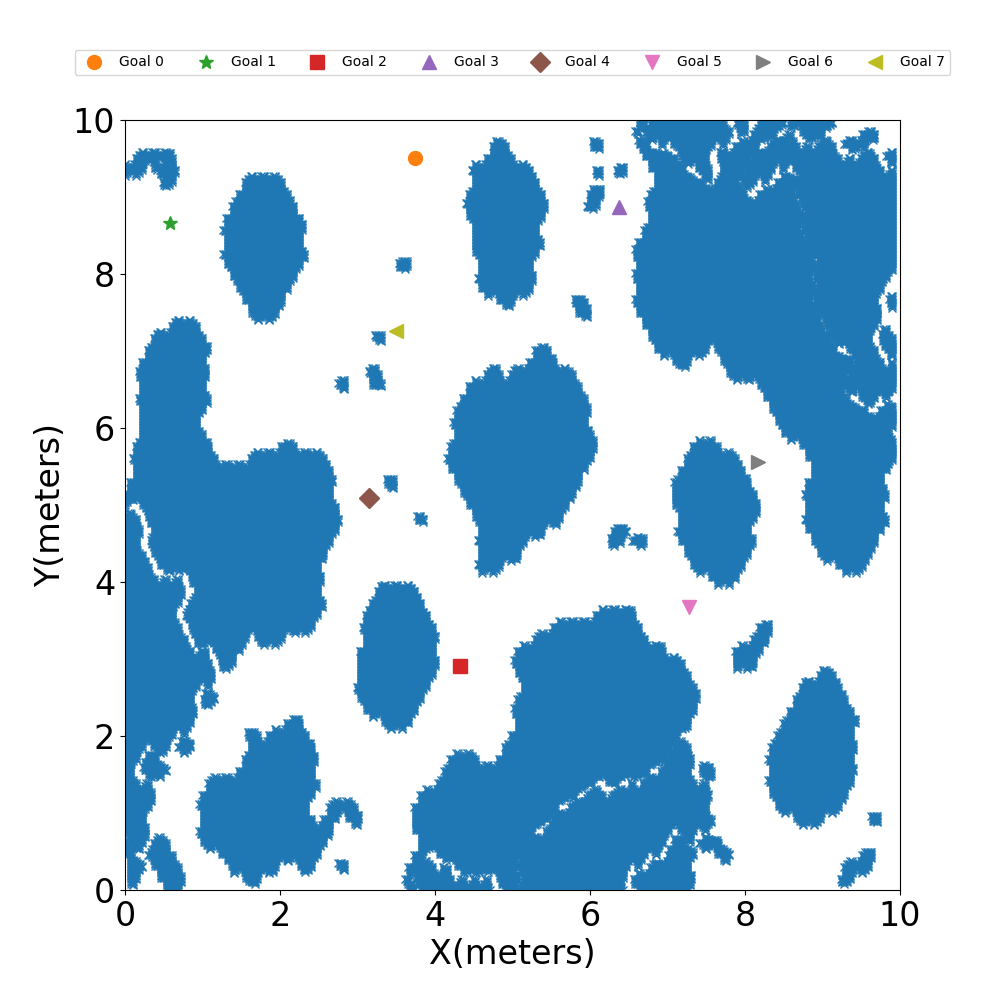}
        \caption{SteppingStones scenario.}
    \end{subfigure}
    \hfill
    \begin{subfigure}{0.45\textwidth}
        \includegraphics[width=\linewidth]{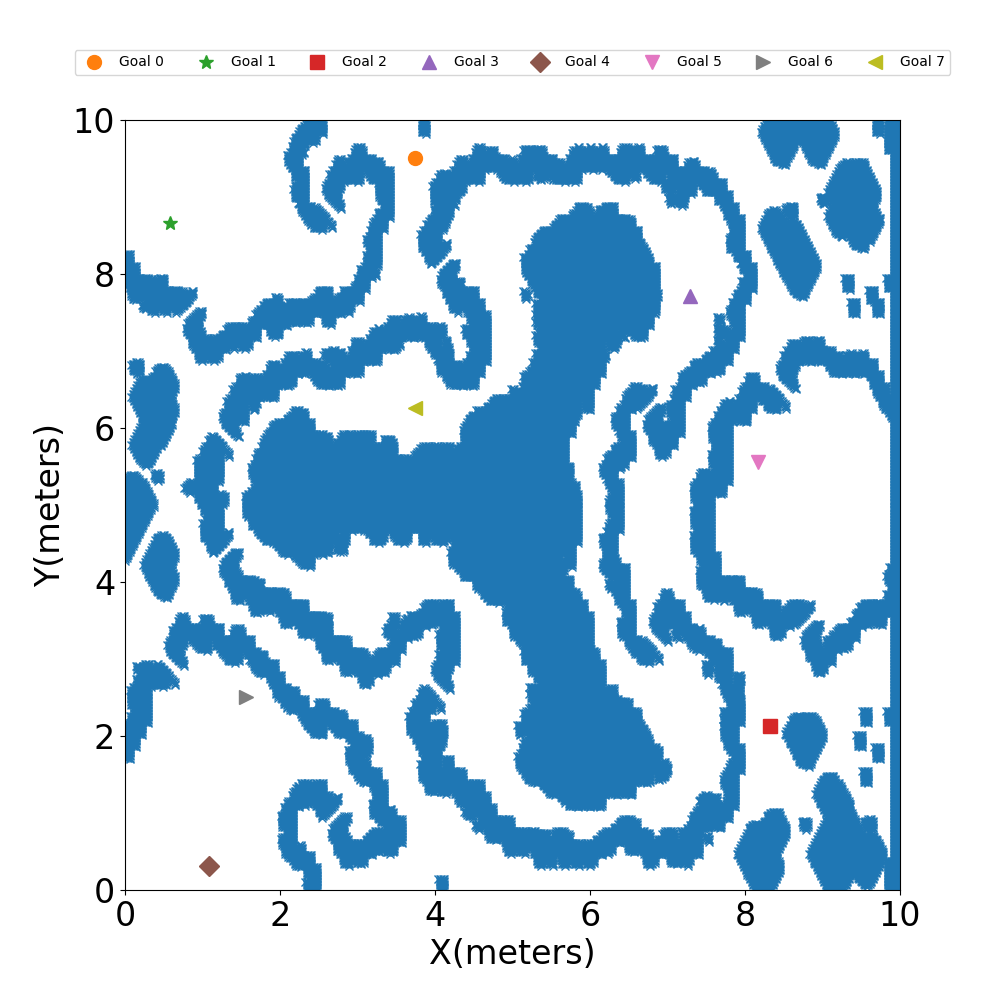}
        \caption{Triskelion scenario.}
    \end{subfigure}
    \vspace*{1em}
    \caption{SpaceAtoll, SpaceDebris, SteppingStones, and Triskelion scenarios used in the experiments.}
\end{figure*}

\end{document}